\documentclass[10pt]{article}

\usepackage[margin=1in]{geometry}
\usepackage{amsmath,amssymb,amsthm,bm,mathtools}
\usepackage{enumitem}
\usepackage{graphicx}
\usepackage{subcaption}
\usepackage{booktabs}
\usepackage{longtable}
\usepackage{adjustbox}
\usepackage{tabularx}
\usepackage{makecell}
\usepackage{float}
\usepackage{placeins}
\usepackage{xcolor}
\usepackage{hyperref}
\usepackage[style=numeric,sorting=nyt]{biblatex}
\hypersetup{hidelinks}


\newtheorem{proposition}{Proposition}

\newtheorem{corollary}{Corollary}
\theoremstyle{definition}

\newtheorem{remark}{Remark}

\newcommand{\R}{\mathbb{R}}
\newcommand{\E}{\mathbb{E}}
\newcommand{\Cov}{\operatorname{Cov}}
\newcommand{\Corr}{\operatorname{Corr}}
\newcommand{\Var}{\operatorname{Var}}
\newcommand{\diag}{\operatorname{diag}}
\newcommand{\1}{\mathbf{1}}

\newcommand{\Hcal}{\mathcal{H}}
\newcommand{\Vcal}{\mathcal{V}}
\newcommand{\argminop}{\operatorname*{arg\,min}}

\newcommand{\safeincludegraphics}[2][]{\includegraphics[#1]{#2}}

\title{Fair Property Tax: Reducing Regressivity in Machine-Learning Valuation}

\author{
Nicol\'as Acevedo Villena\thanks{Operations Research Center, MIT, Cambridge, MA 02139. \texttt{nacevedo@mit.edu}}
\qquad
Haihao Lu\thanks{Sloan School of Management, MIT, Cambridge, MA 02139. \texttt{haihao@mit.edu}}
\\[1ex]
Michael Wagner\thanks{Cook County Assessor's Office, Chicago, IL 60602. \texttt{Michael.Wagner2@cookcountyil.gov}}
\qquad
Timothy Sparer\thanks{Cook County Assessor's Office, Chicago, IL 60602. \texttt{Timothy.Sparer@cookcountyil.gov}}
}
\date{}

\begin{document}
\maketitle

\begin{abstract}

Advanced machine-learning models are increasingly used for property valuation in tax assessment. Although these models can achieve high predictive accuracy, they may systematically assign higher valuation-to-sale ratios to lower-priced properties, a pattern known as regressivity. In collaboration with the Cook County Assessor's Office (CCAO), we develop and evaluate two training methods that substantially reduce this regressive pattern while maintaining price-scale predictive accuracy. The Direct method has been incorporated into CCAO's public residential valuation software. Retaining the LightGBM model class used at CCAO, we propose a Direct penalty on squared covariance between log valuation errors and log sale prices, and an observation-additive Surrogate that yields a weighted squared-error loss. We evaluate both methods in our implementation of the CCAO workflow, using seven chronological validation folds, a held-out test set, and a 2025-forward test set. Measured by the magnitude of price-related bias (PRB), regressivity falls by approximately 49\% on held-out sales and 39\% on 2025 sales at a Surrogate setting selected by cross-validation, relative to the same LightGBM model fitted without the penalty. Price-scale $R^2$ is essentially unchanged. Its small rise, from 0.893 to 0.897 on held-out sales and from 0.904 to 0.908 on 2025 sales, reflects overall valuation level. Log-price RMSE rises by 3.1\% and 1.7\%, respectively. Paired sale-bootstrap 90\% intervals for the PRB reductions lie within about two percentage points of these estimates. In an additional eight-county benchmark, both methods reduce regressivity in every county, with higher log-price prediction error as the expected tradeoff. These findings demonstrate a practical way to reduce measured regressivity within an existing assessor workflow while preserving price-scale predictive performance.

\end{abstract}

\section{Introduction}
\label{sec:introduction}

Property taxation is a cornerstone of local public finance in the United States, supporting public schools, infrastructure, and other essential services \cite{IAAO2020PropertyTaxPolicy,Berry2021}. Because market values are not directly observable, assessment offices must estimate them using available information about properties and market transactions. The quality of these estimates therefore has consequences beyond predictive performance: systematic valuation errors can shift relative tax burdens across property owners. Accurate and equitable property valuation is thus fundamental to the administration of the property tax system.

Producing these valuations at scale is a non-trivial task. Assessors must value large and heterogeneous property inventories using sales observed for only a subset of properties. Automated valuation models (AVMs) support this task by relating observed sale prices to property characteristics, location, and market conditions \cite{IAAO2018AVM,IAAO2025Mass}. Recent advances in machine learning have expanded the methods available for automated valuation, and U.S. assessment offices, including Cook County, have begun incorporating these methods into their valuation workflows \cite{IAAOAITaskForce2022,CCAOModelResAVM2026}. Advanced models such as gradient-boosted trees can capture nonlinear relationships and interactions that conventional specifications may miss, offering opportunities to improve predictive accuracy in mass appraisal \cite{Rootetal2023,Sevgen2024100111,Zilli2024}.

These predictive gains, however, do not by themselves ensure equitable valuation. A model can achieve low aggregate prediction error while retaining systematic differences in relative valuation across the property-value distribution. A particular concern is \emph{regressivity}, in which lower-valued properties are valued more highly relative to their market values than higher-valued properties. Such a pattern can place a disproportionate relative assessment burden on owners of less valuable homes \cite{Berry2021,AvenancioLeonHoward2022}. In sales-based evaluations, it is reflected in \emph{valuation-to-sale ratios}, namely predicted market values divided by observed sale prices, that tend to decline as property value increases. Conventional prediction accuracy does not explicitly control this price-related pattern. The challenge is therefore to retain the predictive advantages of flexible valuation models while reducing systematic disparities across the value distribution.

This paper addresses that challenge through a collaboration with the Cook County Assessor's Office (CCAO). We develop training objectives for the LightGBM model class used in the office's residential valuation workflow, and the Direct objective has already been incorporated as a configurable option in CCAO's public residential modeling software, making it available for future assessment-model runs \cite{CCAOModelResAVM2026,CCAOWagnerResCovPR2026}. Our central empirical finding is that targeted changes to training can substantially reduce measured regressivity while preserving price-scale predictive accuracy. The intervention uses the same sales, predictor information, and model class as the previous CCAO workflow, making it relatively straightforward to adopt for assessors seeking to improve valuation equity within their existing modeling systems.

The assessment literature provides measures of vertical equity and models designed to address regressivity \cite{Cheng1974,IAAO2013Ratio,McMillenSingh2023,CandoganHanLu2023}. Research on fair regression and residual--outcome correction provides related approaches for controlling statistical dependence during training or through post-hoc adjustment \cite{KomiyamaEtAl2018,TrederEtAl2021}. We build on these foundations and propose new approaches to study a question within an assessor's existing model class: 
\begin{center}
    Can we reduce measured regressivity while maintaining price-scale accuracy?
\end{center}

We introduce two methods for incorporating the concern about regressivity into model training. The \emph{Direct} method adds a penalty on the overall price-related trend in valuation ratios, encouraging more consistent relative valuations across the price distribution. The \emph{Surrogate} method uses a weighted prediction loss motivated by this penalty: it assigns greater weight to prediction errors for sales toward the lower and upper ends of the price distribution. This directs the model to place more emphasis on accurately valuing properties at both ends and allows implementation through a standard weighted loss. Both methods retain the existing sales, predictors, and model class, and each introduces a single parameter controlling the strength of the training intervention.

We evaluate the methods using 382,897 eligible Cook County sales from 2016--2024 and a separate sample of 26,641 sales from 2025. The design includes seven chronological validation folds, a held-out test set, and a 2025-forward test set evaluated after refitting on eligible 2016--2024 sales. Within each penalty path, we hold sales, predictors, learner hyperparameters, and evaluation procedures fixed, following the CCAO workflow. We assess price-scale and log-price accuracy alongside valuation level, ratio dispersion, vertical-equity measures, and nonlinear residual patterns. Penalty effects are measured relative to the shared unpenalized custom-objective fit, with ordinary LightGBM retained as the workflow benchmark.

The Cook County results demonstrate substantial reductions in measured regressivity at essentially unchanged price-scale accuracy. At a Surrogate setting selected by cross-validation, held-out price-related bias (PRB) moves from $-0.088$ to $-0.045$, approximately a 49\% reduction in magnitude, while price-scale $R^2$ moves from 0.893 to 0.897 under direct exponentiation. On 2025 sales, PRB moves from $-0.103$ to $-0.062$, approximately a 39\% reduction, while price-scale $R^2$ moves from 0.904 to 0.908. Log-price RMSE rises by 3.1\% on held-out sales and 1.7\% on 2025 sales. Across the four configurations selected by cross-validation, two per method, $|\mathrm{PRB}|$ falls by 22--49\% on the held-out and 2025-forward test sets, relative to the unpenalized fit. Log-price error and nonlinear residual structure must still be checked.

We characterize two ways to incorporate a price-related assessment diagnostic into model training, and demonstrate their effects in the CCAO workflow, whose public software now includes the Direct objective. For assessment practice, changing the training objective is another way to reduce measured regressivity without losing price-scale accuracy. Tracing the full penalty path shows how the correction and its costs change with penalty strength. At matched cross-validation slopes, the Surrogate leaves less nonlinear residual structure on held-out sales than Direct, usually at a higher log-price error. At moderate Direct penalties, a simple post-hoc rescaling of the unpenalized fit gives a similar first-order tradeoff. Paired sale-bootstrap intervals for the regressivity reductions are narrow on both test sets. A separate benchmark using data from ATTOM, a commercial provider of U.S. property data, checks the two correction families in eight other counties under a different data and validation protocol.

The remainder of this section reviews related work. Section~\ref{sec:setting} introduces the valuation setting and evaluation measures. Section~\ref{sec:method} develops the objectives, Section~\ref{sec:empirical_design} describes the empirical design, and Section~\ref{sec:results} presents the results. Section~\ref{sec:discussion} discusses their implications and limitations.

\subsection{Related Work}
\label{subsec:related_work}

\paragraph{Machine learning in mass appraisal.}
Research on automated valuation models (AVMs) compares multiple regression with tree ensembles, including random forests and gradient boosting, and with neural-network methods \cite{WangLi2019,Rootetal2023,Zilli2024,purdueAssessingTechniques}. Reviews emphasize predictive performance across these methods \cite{WangLi2019,Rootetal2023}. In a systematic mapping of 22 machine-learning mass-appraisal studies, \textcite{Zilli2024} find that 82\% report RMSE but only 14\% report the price-related differential, a measure of vertical equity. Tree ensembles are among the competitive methods these studies document \cite{Rootetal2023,Sevgen2024100111,Zilli2024}, making an equity intervention within an existing learner relevant to appraisal practice.

Adoption of ML by assessment offices remains uneven. \textcite{BidansetRakow2022} identify communication, technical capacity, and budget as perceived barriers. An assessor's model must also fit a wider process of sales verification, out-of-time evaluation, ratio studies, interpretation, and administrative review \cite{IAAO2020SalesVerification,IAAOAITaskForce2022}. These requirements motivate testing a training change within a familiar valuation workflow.

\paragraph{Vertical equity and regressivity.}
Assessor research distinguishes \emph{horizontal equity}, the uniform treatment of similar properties, from \emph{vertical equity}, how assessment ratios vary across property values \cite{IAAO2013Ratio,cornia2006property,McMillenSingh2023}. Regressivity is a pattern in which assessment-to-sale ratios tend to decline as sale prices increase: lower-valued properties tend to be relatively overassessed and higher-valued properties relatively underassessed. Ratio studies examine several aspects of uniformity. The Coefficient of Dispersion (COD) summarizes the dispersion of ratios; the Price-Related Differential (PRD) and Price-Related Bias (PRB) address price-related patterns \cite{IAAO2013Ratio,IAAO2023VerticalReview}.

The assessment literature also supplies the specific diagnostic used for training here. \textcite{PaglinFogarty1972} studied the distribution of assessment-to-value ratios, and \textcite{Cheng1974} related log assessed values to log sale prices. Subtracting log sale price from that relationship gives the log-ratio/log-price slope used in this paper. Later studies examined regressivity and alternative statistical tests \cite{Edelstein1979,SundermanEtAl1990}. The slope is thus an established diagnostic, not a new definition of vertical equity. No single statistic captures every relevant pattern: vertical-equity measures can disagree, and some regression-based diagnostics can be biased when applied to values generated by regression-based valuation models \cite{IAAO2023VerticalReview,McMillenSingh2023}. A flatter overall slope can still leave curvature or differences within property and geographic groups. We therefore report complementary diagnostics and examine patterns across the price range in addition to the slope targeted during training.

Evidence from assessment systems shows why those patterns matter \cite{Berry2021,AvenancioLeonHoward2022}. In Cook County, \textcite{Liu2026} reports a flatter log assessed-to-sale-price gradient after a 2021 reform that introduced a more flexible county-wide valuation model and richer data. That study examines assessed values after a broader reform, not the training objectives studied here. More generally, \textcite{SmithEtAl2026} find that adding information to valuation models can improve both predictive accuracy and assessment equity. These studies motivate attention to measured regressivity without implying an unavoidable accuracy--equity tradeoff.

Researchers have pursued correction through spatial and geographically weighted methods, stratification, ratio-study-based validation, and fairness-aware valuation models \cite{Quintos2014,BidansetEtAl2019,BidansetEtAl2025,HermansEtAl2022,HermansEtAl2025,CandoganHanLu2023}. Most closely, \textcite{CandoganHanLu2023} develop a \emph{K-segment} valuation model to address regressivity and study its relationship with prediction accuracy. Our experiments instead change the training objective within LightGBM. Within each penalty comparison, the sales, predictors, and base-learner configuration are held fixed.

\paragraph{Fair regression and dependence-based constrained learning.}
Fair regression studies how to add application-specific requirements when predictions are continuous. Methods can constrain group losses or the distributions of predictions and can act during training or through post-hoc adjustment \cite{AgarwalDudikWu2019,FitzsimonsEtAl2019,ChzhenEtAl2020Wasserstein,ChzhenEtAl2020Recalibration,WeiEtAl2023}. They show how predictive error can be considered alongside another criterion.

A relevant strand controls statistical dependence using correlation or related measures \cite{KomiyamaEtAl2018}, kernel measures such as the Hilbert--Schmidt Independence Criterion \cite{PerezSuayEtAl2017,LiEtAl2022Fairness}, and maximal correlation \cite{MaryEtAl2019,lee2022maximal}. In this application, sale price is both an observed proxy for market value and the prediction target; it is not a protected attribute. Our objectives target the signed first-order association between log prediction errors and log sale prices, not demographic parity. Assessment research supplies the price-related concern, while fair regression supplies a way to express an association as part of training.

\paragraph{Residual--outcome correction outside mass appraisal.}
Closely related corrections exist outside property valuation. In brain-age prediction, the difference between predicted and true age often varies systematically with true age. \textcite{SmithEtAl2019} analyze this problem, and \textcite{BeheshtiEtAl2019} propose a post-hoc correction. \textcite{TrederEtAl2021} control the residual--outcome association during training through a correlation constraint. Later work also changes training losses to address outcome-dependent prediction errors \cite{WangEtAl2023,RenEtAl2026}. These studies establish precedents for both training-time and post-hoc correction. Their goals and model settings differ from ratio-based assessment equity: a correlation constraint in a linear or kernel model is not the same objective as either of our LightGBM penalties. We therefore also report a simple post-hoc rescaling as a supplementary reference.

\paragraph{Position of this study.}
The log-ratio/log-price diagnostic and the general idea of controlling residual--outcome dependence precede this paper \cite{Cheng1974,TrederEtAl2021}. Our contribution is to study two specific objectives derived from that diagnostic within the CCAO's LightGBM workflow. 
We explain how the Direct and observation-additive Surrogate objectives differ, trace their full penalty paths, and evaluate prediction and equity measures on the held-out and 2025-forward test sets; we do not select a penalty strength for deployment.

\section{Preliminaries: The CCAO Baseline and Empirical Motivation}
\label{sec:setting}

In this section we describe the CCAO valuation setting, define the measures used to evaluate the models, and show the baseline result that motivates the correction.

\subsection{The CCAO Residential Valuation Workflow}
\label{subsec:ccao_workflow}

The CCAO values a large and heterogeneous residential inventory on a triennial reassessment cycle \cite{CCAOAssessmentCalendar2026}. Its residential modeling evolved from township-level linear-regression models, with linear-regression and gradient-boosting alternatives used in 2019--2020, to a county-wide LightGBM model beginning in 2021 \cite{CCAOModelResAVM2026}. LightGBM is an implementation of gradient-boosted decision trees \cite{lightgbm}. Studying the historical change in assessed-to-sale ratios, \textcite{Liu2026} reports a flatter price gradient after the transition.

The CCAO's current published residential automated valuation model (AVM) uses verified sales and a broad set of property, spatial, neighborhood, market, and contextual predictors to estimate residential market values \cite{CCAOModelResAVM2026}. The prediction model is embedded in a broader operational workflow comprising data preparation, model training, valuation, evaluation, interpretation, and final review. Candidate valuations are evaluated using both predictive-performance measures and assessor-specific ratio-study statistics across multiple geographic areas and property classes.

The office has merged a configurable Direct covariance-penalty objective into its public residential AVM code. The \texttt{model-res-avm} training configuration can select \texttt{mse\_cov} when log sale prices are used, with supporting custom-objective code in \texttt{lightsnip} \cite{CCAOWagnerResCovPR2026,CCAOWagnerLightsnipCovPR2026}. This establishes integration into the office's modeling software. The numerical results in this paper come from the separate Python research translation; they do not measure the performance of this R implementation or the effect of the objective on issued assessments.

Our main analysis compares ordinary LightGBM with corrections to the same model class, holding its predictors and evaluation design fixed. For context, we also fit a linear research benchmark on the same sales and temporal samples, with different feature processing. This benchmark does not re-estimate the office's earlier township-level models or reconstruct official historical assessments. Section~\ref{subsec:ccao_design} gives the sample construction and validation details.

\subsection{Prediction Setting and the Baseline Model}
\label{subsec:baseline_models}

Let $\mathcal{I}$ denote the set of residential sales retained after the CCAO sales-validation and transaction-eligibility procedures described in Section~\ref{subsec:ccao_design}. For each $i\in\mathcal{I}$, we observe
\[
(x_i,P_i,t_i),
\]
where $x_i$ contains property and parcel characteristics; tract-level socioeconomic and demographic variables from the U.S. Census Bureau's American Community Survey (ACS); and geographic, neighborhood, environmental, accessibility, market, and temporal features.

Here $P_i>0$ is the verified sale price, our observed proxy for the property's unobserved market value, and $t_i$ is the sale date used to construct the chronological samples \cite{CCAOModelResAVM2026,IAAO2026ExposureRatio}. Any time-derived predictors are included in $x_i$.

Since residential sale prices are positive and highly right-skewed, the baseline model is trained on the log-price scale. Define
\begin{equation}
\label{eq:prediction_scale}
y_i=\log P_i,
\qquad
\widehat{y}_i=f(x_i),
\qquad
e_i(f)=\widehat{y}_i-y_i,
\end{equation}
{where $f\in\mathcal{H}$ is a candidate prediction function and $e_i(f)$ is the \emph{log valuation error}, defined with prediction minus observed log price so that it coincides with the log valuation ratio below.}
The corresponding prediction on the original price scale is
\begin{equation}
\label{eq:price_prediction}
\widehat{P}_i=\exp\bigl(f(x_i)\bigr).
\end{equation}

We evaluate predictions on the price scale using $\widehat{P}_i$ from~\eqref{eq:price_prediction}. Direct exponentiation is an operational retransformation, not a correction for conditional-mean bias. Define the \emph{valuation-to-sale ratio} as
\begin{equation}
\label{eq:valuation_ratio}
r_i(f)
=
\frac{\widehat{P}_i}{P_i}
=
\exp\bigl(e_i(f)\bigr).
\end{equation}

We call $r_i(f)$ the \emph{valuation ratio}. Its numerator is the model-estimated market value, before any downstream statutory assessment adjustments. By construction, $e_i(f)=\log r_i(f)$: positive values mean the predicted value exceeds the sale price, and negative values mean it falls below it. For a fixed model we write $r_i$ and $e_i$.

The predictive baseline is ordinary LightGBM, a gradient-boosted tree model representing CCAO's current model class. For a training sample of size $n$, it minimizes mean squared log-price error:

\begin{equation}
\label{eq:baseline_learning}
L(f)
=
\frac{1}{n}\sum_{i=1}^n e_i(f)^2.
\end{equation}
This loss is the predictive component of the objective modified in Section~\ref{sec:method}. LightGBM fits regression trees sequentially and can represent nonlinearities and interactions \cite{CCAOModelResAVM2026,lightgbm}. The ordinary loss has no term that directly controls value-related trends in valuation ratios. For comparison, we also fit ordinary least squares to log prices after training-only feature processing. Appendix~\ref{app:implementation} describes that research benchmark. Section~\ref{subsec:baseline_tension} shows why changing LightGBM's objective is worth studying.

\subsection{
\texorpdfstring{
Predictive, Assessor-Facing, and Residual-Structure Diagnostics}{Predictive, Assessor-Facing, and Residual-Structure Diagnostics}}
\label{subsec:metrics}

For an evaluation sample $S\subseteq\mathcal{I}$, write $n_S=|S|$. We assess prediction on both the log-price training scale and the original price scale. Ratio-study measures use original-scale valuations and answer a different question from predictive accuracy.

\paragraph{Predictive performance.}
We report mean absolute error (MAE) in dollars, mean absolute percentage error (MAPE) relative to sale price, and $R^2_P$ on the original price scale. Root mean squared error on log price, $\operatorname{RMSE}_{\log P}$, measures error on the training scale. MAPE averages the distance of valuation ratios from one, whereas COD (defined below) averages their relative distance from the median ratio. Exact formulas are in Appendix~\ref{app:metrics}.

\paragraph{Valuation level.}
Recall from~\eqref{eq:valuation_ratio} that $r_i=\widehat P_i/P_i$. A ratio above one indicates that the predicted value exceeds the observed sale price, whereas a ratio below one indicates that it falls below the sale price. Define the median, arithmetic mean, and price-weighted mean ratios as
\begin{equation}
\label{eq:level_ratios}
m_S
=
\operatorname{median}_{i\in S}\{r_i\},
\qquad
\bar r_S
=
\frac{1}{n_S}\sum_{i\in S}r_i,
\qquad
\bar r_{W,S}
=
\frac{\sum_{i\in S}\widehat P_i}{\sum_{i\in S}P_i}
=
\frac{\sum_{i\in S}r_iP_i}{\sum_{i\in S}P_i}.
\end{equation}

These ratios describe aggregate valuation level, with one as the neutral value. The median is less sensitive to extremes; the arithmetic mean weights sales equally; and the weighted mean gives more weight to higher-price sales. Agreement on overall level does not establish uniformity or vertical equity.

\paragraph{Ratio dispersion (horizontal uniformity).}
Ratio dispersion is an important assessor-facing dimension of horizontal uniformity. We summarize it using the coefficient of dispersion (COD) and coefficient of variation (COV). The COD is
\begin{equation}
\label{eq:cod}
\operatorname{COD}(S)
=
100\times
\frac{1}{n_S}
\sum_{i\in S}
\frac{|r_i-m_S|}{m_S}.
\end{equation}
The coefficient of variation (COV), defined in Appendix~\ref{app:metrics}, is the standard deviation of ratios as a percentage of their mean. Lower values indicate less aggregate ratio dispersion, but reference ranges depend on property class and market conditions, and unusually low dispersion can itself warrant review \cite{IAAO2013Ratio,IAAO2026ExposureRatio}.
To avoid ambiguity with the covariance penalty introduced later, $\operatorname{COV}$ always denotes the coefficient of variation, whereas statistical covariance is written as $\operatorname{Cov}(\cdot,\cdot)$.

\paragraph{Regressivity (vertical equity).}
Vertical-equity measures ask whether valuation ratios change across the value range. True market value is unobserved, so each measure relies on a sale-based value proxy and can give a different answer \cite{IAAO2023VerticalReview,IAAO2026ExposureRatio}. We report PRD, PRB, MKI, and VEI together. Table~\ref{tab:assessment_metrics_summary} distinguishes adopted reference ranges from the proposed May 2026 exposure-draft ranges; none is treated as a finding of legal compliance.

The \emph{price-related differential} (PRD) is
\begin{equation}
\label{eq:prd}
\operatorname{PRD}(S)
=
\frac{\bar r_S}{\bar r_{W,S}}.
\end{equation}
Because the weighted mean gives more weight to higher-price sales, PRD above one suggests a regressive direction. The adopted reference range is $[0.98,1.03]$ \cite{IAAO2013Ratio}. High-value sales and extreme ratios can affect PRD, so we read it alongside the other measures \cite{IAAO2026ExposureRatio}.

{The following regression- and group-based vertical-equity diagnostics require a proxy for the property's value level. Ranking properties only by sale price can mechanically bias ratio-based vertical-equity tests toward finding regressivity because sale price also appears in the denominator of $r_i$, while using valuation alone can induce bias in the opposite direction. Following IAAO guidance, we therefore use the equal-weight market-value proxy}
\begin{equation}
\label{eq:market_value_proxy}
V_i^{\mathrm{proxy}}
=
\frac{1}{2}
\left(
P_i+\frac{\widehat P_i}{m_S}
\right),
\end{equation}
{which combines the verified sale price with the median-normalized predicted value to reduce this source of proxy-induced bias \cite{IAAO2013Ratio,IAAO2026ExposureRatio}.}

Both PRB and VEI use the equal-weight proxy in Eq.~\eqref{eq:market_value_proxy}. Appendix~\ref{app:metrics} records the exposure draft's notation discrepancy and the check against the executed metric code.

{We first report the \textit{price-related bias} (PRB).}
Let $\widetilde V_S^{\mathrm{proxy}}$ denote the median of $V_i^{\mathrm{proxy}}$ over $S$. The PRB is the slope in the simple linear regression
\begin{equation}
\label{eq:prb}
\frac{r_i-m_S}{m_S}
=
a_S
+
\operatorname{PRB}(S)\cdot
\log_2\left(
\frac{V_i^{\mathrm{proxy}}}
{\widetilde V_S^{\mathrm{proxy}}}
\right)
+
\varepsilon_i.
\end{equation}

Negative PRB values indicate that valuation ratios tend to decline as the market-value proxy increases and are therefore associated with regressivity. Values between $-0.05$ and $0.05$ are conventionally preferred \cite{IAAO2013Ratio}.

The \emph{Modified Kakwani Index} (MKI) \cite{Quintos2020} compares the concentration of predicted values, ordered by sale price, with the distribution of sale prices. When the index is defined, a value of one indicates neutrality on this measure; values below one point in the regressive direction \cite{IAAO2026ExposureRatio}. Its sale-price ordering can affect the result when sale price imperfectly measures market value. We therefore read MKI alongside the other measures; its construction is in Appendix~\ref{app:metrics}.

Finally, we report the \emph{Vertical Equity Indicator} (VEI), the quantile-grouped vertical-equity measure emphasized in the May 2026 IAAO exposure draft \cite{IAAO2026ExposureRatio}.
The draft varies the number of percentile groups with sample size: for the large evaluation samples used in the main countywide analysis ($n_S\geq 501$), observations are divided into deciles according to $V_i^{\mathrm{proxy}}$. Let $S_{q_1}$ and $S_{q_{10}}$ denote the lowest and highest deciles, and let $m_{S_{q_1}}$ and $m_{S_{q_{10}}}$ denote their median valuation ratios. The VEI is
\begin{equation}
\label{eq:vei}
\operatorname{VEI}(S)
=
100\times
\frac{m_{S_{q_{10}}}-m_{S_{q_1}}}{m_S}.
\end{equation}

A negative VEI indicates that higher-value properties have lower median valuation ratios than lower-value properties. The exposure draft proposes a reference band of $[-10\%,10\%]$, together with confidence-interval-based testing for point estimates outside that band.

\begin{table}[!ht]
\centering
\footnotesize
\setlength{\tabcolsep}{4pt}
\renewcommand{\arraystretch}{1.10}

\newcommand{\metriclabel}[1]{\hspace*{0.9em}#1}

\caption{Predictive and assessor-facing evaluation measures.}
\label{tab:assessment_metrics_summary}

\begin{tabularx}{\textwidth}{
@{}
>{\raggedright\arraybackslash}p{3.8cm}
>{\centering\arraybackslash}p{1.8cm}
>{\centering\arraybackslash}p{2.9cm}
>{\centering\arraybackslash}p{2.3cm}
>{\raggedright\arraybackslash}X
@{}
}
\toprule
\textbf{Measure}
& \textbf{\shortstack{Ideal value}}
&
\textbf{\shortstack{Reference range}}
& \textbf{Guidance status}
& \textbf{Interpretation} \\
\midrule

\textbf{Predictive performance}
& & & & \\
\cmidrule(r){1-1}

\metriclabel{$R^2_P$}
& $1$
& --
& --
& Original price-scale goodness of fit. \\

\metriclabel{$\operatorname{MAE}_P$}
& $0$
& --
& --
& Mean absolute prediction error in dollars. \\

\metriclabel{$\operatorname{MAPE}_P$}
& $0$
& --
& --
& Mean absolute prediction error relative to sale price. \\

\metriclabel{$\operatorname{RMSE}_{\log P}$}
& $0$
& --
& --
& RMSE on the log-price estimation scale. \\

\midrule
\addlinespace[5pt]
\textbf{{Valuation level}
}
& & & & \\
\cmidrule(r){1-1}

\metriclabel{Median ratio $m_S$}
& $1$
& $[0.90,1.10]$
& IAAO 2013 (adopted)
& Robust aggregate valuation level. \\

\metriclabel{Mean ratio $\bar r_S$}
& $1$
& $[0.90,1.10]$
& IAAO 2013 (adopted)
& Unweighted aggregate valuation level. \\

\metriclabel{Weighted mean $\bar r_{W,S}$}
& $1$
& $[0.90,1.10]$
& IAAO 2013 (adopted)
& Value-weighted valuation level. \\

\midrule
\addlinespace[6pt]
\textbf{{Horizontal uniformity}
}
& & & & \\
\cmidrule(r){1-1}

\metriclabel{COD}
& --
& $[5.0,15.0]$
& IAAO 2013 (adopted)
& Absolute dispersion relative to the median ratio. \\

\metriclabel{COV}
& --
& --
& --
& Relative dispersion around the mean ratio. \\

\midrule
\addlinespace[5pt]
\textbf{Vertical equity}
& & & & \\
\cmidrule(r){1-1}

\metriclabel{PRD}
& $1$
& $[0.98,1.03]$
& IAAO 2013 (adopted)
& Mean-to-weighted-mean ratio comparison. \\

\metriclabel{PRB}
& $0$
& $[-0.05,0.05]$
& IAAO 2013 (adopted)
& Regression-based value-related ratio trend. \\

\metriclabel{MKI}
& $1$
& $[0.95,1.05]$
& May 2026 (exposure draft)
& Distributional concentration relative to sale prices. \\

\metriclabel{VEI}
& $0$
& $[-10\%,10\%]$
& May 2026 (exposure draft)
& High-versus-low value-group median-ratio difference. \\

\bottomrule
\end{tabularx}

\vspace{1mm}
\begin{minipage}{\textwidth}
\scriptsize
\emph{Notes.}
``--'' means that no universal ideal or range is imposed. The level-ratio, COD,
PRD, and PRB ranges follow adopted IAAO guidance
\cite{IAAO2013Ratio,IAAO2018AVM,IAAO2025Mass}; CCAO's ratio-study software
uses the same ranges and the MKI band \cite{CCAOassessr}. Appendix F of the
May 2026 exposure draft notes that some consider the MKI band optimal for
lower-variability strata, and its Section 11.2.6 and Appendix E propose the
VEI band. Neither band is an adopted standard \cite{IAAO2026ExposureRatio}.
The COD range is the 2013 IAAO range for older or heterogeneous single-family areas; newer or homogeneous areas have $[5.0,10.0]$.
Appendix~\ref{app:metrics} gives further guidance details.
\end{minipage}

\end{table}

\paragraph{
Mechanism and residual-structure diagnostics.}
{These two diagnostics have no assessor-standard reference ranges, and each has a
distinct role: $\beta_{\log}$ measures the targeted first-order trend, and
$\Delta_{\mathrm{NL}}$ measures nonlinear conditional-mean structure beyond it.
A broader dependence measure, distance correlation, is defined in
Appendix~\ref{app:metrics}.

\emph{First-order mechanism.}
We first report the slope from regressing log valuation ratios on log sale prices:
\begin{equation}
\label{eq:log_ratio_slope}
\beta_{\log}(S)
=
\frac{
\widehat{\operatorname{Cov}}_S(e,y)
}{
\widehat{\operatorname{Var}}_S(y)
}
=
\frac{
\widehat{\operatorname{Cov}}_S(\log r,\log P)
}{
\widehat{\operatorname{Var}}_S(\log P)
}.
\end{equation}
A value of zero indicates no first-order log-ratio--log-sale-price trend. Negative
values indicate the regressive direction and positive values the progressive
direction. For a fixed evaluation sample, $\widehat{\operatorname{Var}}_S(y)$ is
common across models, so $\beta_{\log}(S)$ is the signed first-order slope
corresponding directly to the covariance targeted during training.
The slope predates this paper \cite{Cheng1974}. It is a training-mechanism
diagnostic rather than a standalone measure of vertical equity: sale price is
an imperfect value proxy and appears in the denominator of $r_i$.

\emph{Nonlinear conditional-mean structure.}
Let $Z$ be standardized log sale price. The correlation ratio $\eta^2(e\mid Z)$ measures how much log-error variation is explained by its conditional mean given $Z$; $\Corr(e,Z)^2$ measures the part explained by the best straight-line relationship \cite{DoksumSamarov1995}. Their difference is the \emph{correlation-ratio nonlinearity gap}:
\begin{equation}
\label{eq:nonlinearity_gap}
\Delta_{\mathrm{NL}}
:=
\eta^2(e\mid Z)-\Corr(e,Z)^2.
\end{equation}
At the population level the gap is nonnegative and is zero when average log error has a straight-line relationship with log sale price. A smaller $\lvert\beta_{\log}\rvert$ therefore need not mean that curved patterns have disappeared. The empirical estimate uses cross-fitted spline regressions and can miss patterns outside that family, so a small value does not establish a flat ratio profile. Appendix~\ref{app:metrics} gives the definitions and estimator.
}

The measures are descriptive diagnostics, not findings of standards compliance. We compute them on chronological validation folds before aggregating the paths; countywide summaries can hide local differences. The training objectives do not optimize the full suite of evaluation measures.

\subsection{Baseline Comparison and Motivation}
\label{subsec:baseline_tension}

The two unpenalized models show a clear prediction--ratio-pattern contrast. On held-out sales, ordinary LightGBM has a higher price-scale $R^2_P$ than the linear benchmark (0.894 versus 0.799), but a more negative PRB ($-0.091$ versus $-0.016$). The same ordering appears in the separate 2025 evaluation. Linear regression also has greater ratio dispersion and does not eliminate residual nonlinearity (Appendix Table~\ref{tab:ccao_baseline_complementary}).

Table~\ref{tab:ccao_baseline_results} reports both models on the held-out and 2025-forward test sets. The 2025 sales are not used for model selection; Section~\ref{subsec:ccao_design} gives the temporal design. The question pursued in the rest of the paper is whether changing LightGBM's training objective reduces its value-related ratio pattern while retaining its predictive strength.

{\begin{table}[H]
\centering
\scriptsize
\setlength{\tabcolsep}{2.4pt}
\renewcommand{\arraystretch}{1.06}
\newcommand{\baselineprimarymetric}[1]{\hspace*{0.6em}#1}
\caption{Prediction and vertical-equity measures for the unpenalized linear benchmark and ordinary LightGBM on held-out and 2025 sales.}
\label{tab:ccao_baseline_results}
\begin{tabularx}{\textwidth}{@{} >{\raggedright\arraybackslash}p{2.80cm} >{\centering\arraybackslash}X >{\centering\arraybackslash}X >{\centering\arraybackslash}X >{\centering\arraybackslash}X @{}}
\toprule
\textbf{Measure} & \multicolumn{2}{c}{\textbf{Held-out}} & \multicolumn{2}{c}{\textbf{2025-forward}} \\
\cmidrule(lr){2-3}\cmidrule(l){4-5}
& \textbf{Linear} & \textbf{LightGBM} & \textbf{Linear} & \textbf{LightGBM} \\
\midrule
\multicolumn{5}{@{}l}{\textbf{Prediction}} \\
\cmidrule(r){1-1}
\baselineprimarymetric{$R^2_P$} & 0.799 & \textbf{0.894} & 0.799 & \textbf{0.904} \\
\baselineprimarymetric{$\operatorname{MAE}_P$} & \$90,092 & \textbf{\$75,655} & \$99,371 & \textbf{\$78,484} \\
\baselineprimarymetric{$\operatorname{MAPE}_P$} & 24.1\% & \textbf{21.2\%} & 24.9\% & \textbf{20.8\%} \\
\baselineprimarymetric{$\operatorname{RMSE}_{\log P}$} & 0.322 & \textbf{0.289} & 0.313 & \textbf{0.278} \\
\addlinespace[5pt]
\multicolumn{5}{@{}l}{\textbf{Vertical equity}} \\
\cmidrule(r){1-1}
\baselineprimarymetric{PRD} & \textbf{1.042} & 1.069 & \textbf{1.052} & 1.079 \\
\baselineprimarymetric{PRB} & \textbf{-0.016} & -0.091 & \textbf{-0.029} & -0.106 \\
\baselineprimarymetric{MKI} & \textbf{0.975} & 0.923 & \textbf{0.954} & 0.907 \\
\baselineprimarymetric{VEI} & \textbf{-11.6\%} & -26.5\% & \textbf{-17.3\%} & -28.6\% \\
\bottomrule
\end{tabularx}
\vspace{1mm}
\begin{minipage}{\textwidth}
\scriptsize
\emph{Notes.}
Both models use the same sales, log-price target, temporal samples, metric code, and direct exponentiation of predictions. Held-out results use models fitted on 344,607 development sales ($n=38{,}290$ evaluated); 2025 results use models refitted on 382,897 eligible 2016--2024 sales ($n=26{,}641$ evaluated). The linear fit uses different feature processing (Appendix~\ref{app:implementation}); the accuracy difference does not isolate model class. Bold marks the better displayed value within each sample: higher $R^2_P$, lower errors, and closer to neutrality for equity measures. MKI and VEI have proposed, not adopted, guidance (Table~\ref{tab:assessment_metrics_summary}). These point estimates do not establish standards compliance or a significant difference. MAPE and VEI are percentages; MAE is in dollars.
\end{minipage}
\end{table}

}

\begin{figure}[H]
\centering
\safeincludegraphics[width=0.98\textwidth]{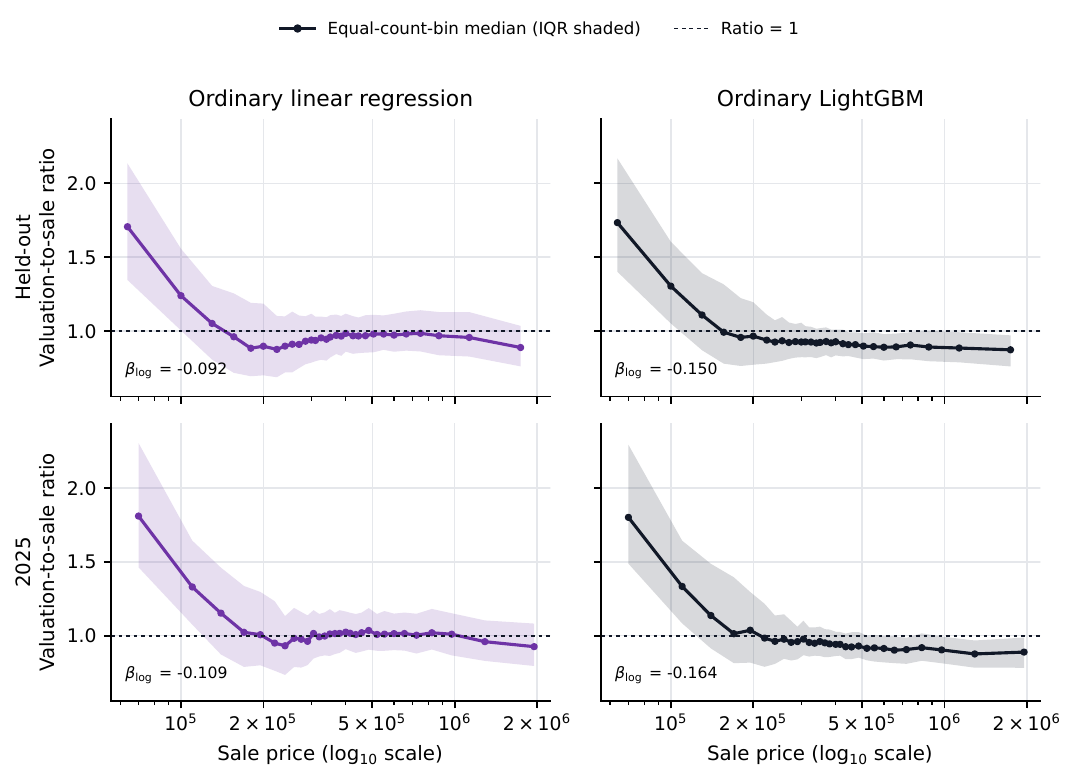}
\caption{Valuation-to-sale ratios by sale price for the linear benchmark and ordinary LightGBM on held-out and 2025 sales. Curves show medians in 30 equal-count price bins; shading shows within-bin interquartile ranges (IQR), not confidence intervals. Panels report the global log-ratio/log-price slope $\beta_{\log}$; horizontal lines mark a ratio of one.}
\label{fig:baseline_motivation}
\end{figure}

Figure~\ref{fig:baseline_motivation} shows that both models have elevated ratios among the lowest-priced sales. Valuation ratios also vary more within the lowest-priced bins. The linear model has a less negative global slope, yet its binned ratio profile bends across prices. Grouping sales by the value proxy (Eq.~\eqref{eq:market_value_proxy}) reveals a similar low-end elevation for ordinary LightGBM: the lowest decile's median valuation ratio is 1.15 on held-out and 1.19 on 2025 sales, while deciles 2--10 range from 0.91 to 0.96. The negative VEI thus mainly reflects the elevated lowest decile.

The penalty developed next targets the square of the covariance in Eq.~\eqref{eq:log_ratio_slope}. On a fixed training sample, log-sale-price variance is common to all fitted models, so reducing the magnitude of covariance reduces the magnitude of the corresponding first-order slope. PRB, VEI, and the other assessor-facing measures remain separate out-of-sample checks.

\section{Covariance-Guided Training Objectives}
\label{sec:method}

We target the overall slope of log valuation ratios against log sale prices among
training sales. The Direct objective penalizes the squared covariance behind this
slope. The Surrogate uses an observation-additive upper bound that fits a standard
boosting custom objective. Within either formulation, $\rho$ controls penalty
strength; the prediction target, predictors, and base model class remain fixed.

Let $J_0(f)$ denote the baseline training objective, whose data-fit component
is the squared log-price loss $L(f)$ defined in
Section~\ref{subsec:baseline_models}. We compute the centering quantities and
penalties from the training sample only, separately for each temporal fold.

\subsection{{Log-Scale Regressivity Target}}
\label{subsec:training_regressivity}

Let
\begin{equation}
\label{eq:centered_log_price}
\bar y=\frac{1}{n}\sum_{i=1}^n y_i,
\qquad
c_i=y_i-\bar y,
\qquad
\bar e(f)=\frac{1}{n}\sum_{i=1}^n e_i(f).
\end{equation}
The empirical covariance between log valuation errors and log sale prices is
\begin{equation}
\label{eq:training_covariance}
C(f)
=
\widehat{\operatorname{Cov}}\bigl(e(f),y\bigr)
=
\frac{1}{n}\sum_{i=1}^n
\bigl(e_i(f)-\bar e(f)\bigr)(y_i-\bar y)
=
\frac{1}{n}\sum_{i=1}^n e_i(f)c_i.
\end{equation}
The last equality follows from $\sum_{i=1}^n c_i=0$, so centering the residuals does not change the covariance.

This covariance has a direct slope interpretation. Let
\[
V_y=\frac{1}{n}\sum_{i=1}^n c_i^2.
\]
Provided $V_y>0$, the least-squares slope from regressing log valuation ratios on log prices is
\begin{equation}
\label{eq:covariance_slope}
\operatorname{slope}\bigl(e(f)\sim y\bigr)
=
\frac{C(f)}{V_y}.
\end{equation}
Because $V_y$ is fixed by the training sample, reducing $|C(f)|$ is equivalent to reducing the magnitude of this global linear trend. In particular, $C(f)<0$ indicates that log valuation ratios tend to decline with log sale price, which is the first-order pattern associated with regressivity.

Because sale price enters both the valuation error and the value benchmark,
even the population predictor that minimizes squared error has negative
error--price covariance when the features cannot explain all price variation
(Appendix~\ref{app:population_identity}; Appendix~\ref{app:theory} gives a
finite-sample analogue for least squares). Negative covariance alone therefore
does not establish inequity relative to latent market values, and zero
covariance does not establish fairness.

{This slope is a training-scale mechanism measure, not the assessor-facing PRB statistic defined in Section~\ref{subsec:metrics}. We use covariance rather than correlation because $V_y$ is fixed within the training sample, whereas correlation additionally normalizes by the model-dependent residual variance. Covariance therefore provides a direct training target for the global value-related slope.}

{Sale price enters this construction as the observed market-value benchmark for verified training sales; it is not introduced as a protected attribute. The broader connection to dependence-based fair regression is discussed in Section~\ref{subsec:related_work}.}

\subsection{{Direct Squared-Covariance Penalty}}
\label{subsec:direct_penalty}

{We first penalize the chosen first-order target directly. The \emph{covariance-penalized learner} is}

{
\begin{equation}
\label{eq:direct_penalty}
f_\rho^{\mathrm{cov}}
\in
\argminop_{f\in\Hcal}
\left\{
J_0(f)+\frac{\rho}{2}C(f)^2
\right\},
\qquad
\rho\geq 0.
\end{equation}
}

At $\rho=0$, formulation~\eqref{eq:direct_penalty} reduces to the baseline
training objective. This equality of objectives does not imply that our
custom-objective LightGBM fit is numerically identical to ordinary native
LightGBM; Section~\ref{subsec:ccao_design} defines their distinct roles.
Increasing $\rho$ puts more weight on moving the global first-order trend
toward zero, without favoring a negative or a positive slope.

{For interpretation, it is enough to consider the contribution of the covariance penalty to the gradient of observation $i$:}
{
\begin{equation}
\label{eq:direct_penalty_gradient}
\frac{\partial}{\partial \widehat y_i}
\left[
\frac{\rho}{2}C(f)^2
\right]
=
\frac{\rho}{n}C(f)c_i.
\end{equation}
}

If $C(f)<0$, this gradient pushes predictions down for training sales below
the geometric-mean price and up for sales above it. This describes the local
penalty gradient, not a guaranteed change to each prediction after trees are
retrained. The direction reverses when $C(f)>0$.

The Direct penalty does not set the overall valuation level: multiplying
every predicted value by the same positive constant changes the ratios but
leaves $C(f)$ unchanged. We assess ratio level separately.

{For second-order models (e.g., LightGBM), the corresponding Hessian of the squared log-price loss plus direct penalty is}
{
\begin{equation}
\label{eq:direct_gradient_hessian}
\nabla_{\widehat y}^2
\left(
L(f)+\frac{\rho}{2}C(f)^2
\right)
=
\frac{2}{n}I+\frac{\rho}{n^2}cc^\top.
\end{equation}
}
The rank-one term couples observations. LightGBM's standard custom-objective
interface accepts one Hessian value per observation, so it cannot supply this
cross-observation curvature. Our Direct implementation uses the exact
gradient and the Hessian's diagonal; the omitted curvature and its
implementation audit are described in Appendix~\ref{app:implementation}.
The next objective avoids this coupling.

\subsection{{Observation-Additive Covariance Upper-Bound Surrogate}}
\label{subsec:surrogate}

For a penalty that separates across training sales, apply Jensen's inequality
to $C(f)=n^{-1}\sum_i e_i(f)c_i$:

{
\begin{equation}
\label{eq:surrogate_upper_bound}
C(f)^2
=
\left(
\frac{1}{n}\sum_{i=1}^n e_i(f)c_i
\right)^2
\leq
\frac{1}{n}\sum_{i=1}^n e_i(f)^2c_i^2.
\end{equation}
}
Define the observation-additive Surrogate penalty as
\begin{equation}
\label{eq:surrogate}
\Psi^{\mathrm{surr}}(f)
=
\frac{1}{n}\sum_{i=1}^n e_i(f)^2c_i^2
=
\frac{1}{n}\sum_{i=1}^n
e_i(f)^2(y_i-\bar y)^2.
\end{equation}
The resulting \emph{surrogate-penalized learner} is therefore
\begin{equation}
\label{eq:surrogate_learning}
f_\rho^{\mathrm{surr}}
\in
\argminop_{f\in\Hcal}
\left\{
{J_0(f)}
+
\rho\Psi^{\mathrm{surr}}(f)
\right\}.
\end{equation}
{
The Surrogate is an upper bound on squared empirical covariance. It also
penalizes variation in the individual contributions $e_i(f)c_i$:
\begin{equation}
\label{eq:surrogate_decomposition}
\Psi^{\mathrm{surr}}(f)
=
C(f)^2
+
\frac{1}{n}
\sum_{i=1}^n
\left(e_i(f)c_i-C(f)\right)^2.
\end{equation}
Opposite-signed contributions can cancel in $C(f)$ but not in the
Surrogate. The bound makes the loss separable, but it does not make the
Surrogate equivalent to Direct, guarantee a smaller covariance on later
sales, or control nonlinear residual--price patterns.
}

Combining the baseline loss with the surrogate gives
\begin{equation}
\label{eq:surrogate_weighted_loss}
L(f)+\rho\Psi^{\mathrm{surr}}(f)
=
\frac{1}{n}\sum_{i=1}^n
\left[1+\rho(y_i-\bar y)^2\right]e_i(f)^2.
\end{equation}
The squared-error weights, fixed once the training sales are known, are
\begin{equation}
\label{eq:surrogate_weights}
w_i(\rho)=1+\rho(y_i-\bar y)^2.
\end{equation}
Properties near the center of the log-price distribution receive weights close to the baseline weight, whereas properties in the lower and upper tails receive larger weights.

The center $\exp(\bar y)$ is the geometric-mean sale price. Prices equally far
from it on the log scale receive the same additional weight, although the
numbers of sales and fitted responses in the two tails need not match.
Weights grow quadratically with log-price distance, so extreme sales can
carry substantial weight.

Unlike Direct, the Surrogate's observation-level correction depends on each
error and its squared log-price deviation, rather than on the sign of the
aggregate covariance. Its effect on the signed slope and the assessor-facing
diagnostics must be evaluated out of sample.

Because its loss separates across sales, the Surrogate supplies one gradient
and one Hessian value per sale. Appendix~\ref{app:implementation} gives the
derivatives and the scaling used in the reported LightGBM implementation.
The same loss could also be supplied as fixed observation weights, but
whether this gives the same boosting path would need a separate check.

At large $\rho$, the weights concentrate on sales far from the log-price
center; Appendix~\ref{app:ccao_path_details} reports this concentration at
$\rho=100$.

{Unlike the direct covariance penalty, the surrogate is not invariant to a common shift in all log predictions because it penalizes the magnitude of $e_i(f)$ itself. It can therefore affect
overall valuation level in addition to the value-related trend.
}

\subsection{Interpretation and Scope}
\label{subsec:correction_scope}

Direct responds to the aggregate signed covariance; Surrogate gives more
weight to squared errors farther from the training price center. Its
additional dispersion term in~\eqref{eq:surrogate_decomposition} makes it a
distinct objective, not an approximation to Direct. The penalties also have
different scales, so equal numerical values of $\rho$ do not represent equal
correction strengths.

The implementation audit leaves open how much the omitted Direct curvature
affects the fitted path (Appendix~\ref{app:implementation}).

Neither objective directly optimizes the assessor-facing measures in
Section~\ref{subsec:metrics}. Direct's penalty is invariant to a common shift
in log predictions; Surrogate can affect valuation level, but neither penalty
imposes a target valuation ratio of one. The first-order training slope also
cannot rule out nonlinear, geographic, or
property-class patterns. We therefore evaluate prediction, ratio level,
vertical equity, and residual structure separately on the held-out and 2025-forward
test sets.
Sale price enters the penalty only for verified training sales; once fitted,
the model predicts for unsold properties from their features alone.

Raw $\rho$ is not intrinsically portable across training samples.
Appendix~\ref{app:implementation} explains a variance-normalized coordinate
for comparing the scales of the idealized objectives. In a fixed linear
prediction space with an intercept, Direct acts as a centered spread
rescaling, while Surrogate is a weighted projection
(Appendix~\ref{app:theory}). The Direct result motivates the supplementary post-hoc
rescaling reference in Section~\ref{subsec:comparators}. These fixed-space
results also show that the baseline squared-error loss increases only to
second order near $\rho=0$. They explain why the objectives can behave
differently, but do not guarantee the paths of retrained boosted trees.
Section~\ref{sec:empirical_design} describes how we evaluate those paths.

\section{Application and Evaluation Design}
\label{sec:empirical_design}

{This section specifies the empirical design used to evaluate the proposed correction. We first describe the CCAO application, data, and implementation; then define the chronological validation and the range of penalty strengths tested; and finally summarize the models used for comparison. All model comparisons reported in the next section use the evaluation criteria defined in Section~\ref{subsec:metrics}.}

\subsection{{The CCAO Residential Pipeline: Application, Data, and Implementation}}
\label{subsec:ccao_design}

We implemented in Python the parts of CCAO's published residential workflow needed for model development and evaluation: data preparation, features, model fitting, and ratio studies. The results come from this research implementation, not from official CCAO assessment runs.

In the LightGBM comparisons, we hold the sale filters, log-price target,
predictors, temporal design, price-scale transformation, evaluation
procedures, and model hyperparameters fixed. Within each custom objective,
only $\rho$ varies. Ordinary LightGBM and the custom-objective fit at
$\rho=0$ have very similar overall results, but their individual predictions
differ because LightGBM's built-in and custom objectives follow different
execution paths under feature subsampling. We use the custom-objective fit at
$\rho=0$ as the reference for measuring the effect of each penalty.
Appendix~\ref{app:implementation} gives details.

We use verified Cook County sales of eligible single-family and small multifamily residential properties, including townhomes. We exclude condominiums, which CCAO values using a separate model \cite{CCAOModelResAVM2026}, as well as multicard PINs and sales marked as outliers. All reported CCAO samples come from the final residential-modeling extract provided directly by CCAO.

We sort eligible 2016--2024 sales chronologically. The oldest 90\% (344,607 sales, through November 9, 2023) form the development set, used for training and cross-validation, and the newest 10\% (38,290 sales, November 9, 2023 through December 31, 2024) form the held-out test set; sales on the cutoff date fall on both sides, but no sale appears in both. Each configuration is refit on the full development set before evaluation on the held-out test set. We also evaluate each configuration on a 2025-forward test set of 26,641 sales, after refitting it on all 382,897 eligible 2016--2024 sales (Table~\ref{tab:ccao_samples}). This approximates how a model fitted to past sales is used to value properties in a later year.

\begin{table}[!ht]
\centering
\caption{Final CCAO temporal samples.}
\label{tab:ccao_samples}
\begin{tabular}{llr}
\toprule
Sample & Period & Observations \\
\midrule
Development (training and validation) & 2016-01-01--2023-11-09 & 344,607 \\
Held-out test set & 2023-11-09--2024-12-31 & 38,290 \\
Full 2016--2024 refit set & 2016-01-01--2024-12-31 & 382,897 \\
2025-forward test set & 2025-01-01--2025-12-29 & 26,641 \\
\bottomrule
\end{tabular}
\end{table}

The models use 95 predictors from the CCAO pipeline, including property characteristics, location, proximity, Census measures, time, parcel shape, and neighborhood identifiers. Of these, 23 are categorical. All LightGBM fits use the same predictors; the penalties add none. Table~\ref{tab:feature_groups} summarizes the predictor groups.

{
\begin{table}[!ht]
\centering
\caption{CCAO predictor groups used in the models.}
\label{tab:feature_groups}
\begin{tabular}{lrl}
\toprule
Feature group & Count & Examples \\
\midrule
Assessor/property variables & 30 & Class, building area, land area, age, rooms, neighborhood \\
Location variables & 9 & Coordinates, census tract, school district, municipality \\
Proximity variables & 26 & Transit, parks, roads, hospitals, water, airports \\
ACS 5-year estimates & 16 & Income, education, poverty, employment, tenure \\
Time variables & 8 & Sale year, month, quarter, day, post-COVID indicator \\
Parcel-shape variables & 6 & Edge/angle dispersion, bounding ratios, vertex count \\
\midrule
Total & 95 & 23 categorical predictors in LightGBM \\
\bottomrule
\end{tabular}
\end{table}
}

\subsection{Cross-Validation and Penalty Strengths}
\label{subsec:path_design}

We use seven expanding-window cross-validation folds within the 344,607-sale
development set. The first six windows end at 15-month intervals from
January 1, 2016; the seventh ends on November 9, 2023. In each window, the
most recent 10\% of sales form the validation set. For fold $v$, let
$\mathcal T_v$ and $\mathcal V_v$ denote the training and validation sets. We
calculate all quantities needed by the objectives, including the mean and
centered log prices, using only $\mathcal T_v$.
Appendix~\ref{app:ccao_path_details} reports the fold dates and sample sizes.

We summarize each metric by its equal-weight mean over the seven folds (D1).
The last two validation sets overlap (20,988 sales appear in both), so the
folds are not independent replications and their spread is descriptive.
Appendix~\ref{app:ccao_path_details} gives two sensitivity summaries (D2 and D3).

We evaluate Direct and Surrogate at 82 values of the penalty strength
$\rho>0$, from approximately 0.00110 to 100
(Appendix~\ref{app:ccao_path_details}). We measure each penalty's effect
relative to the custom-objective fit at $\rho=0$ and also compare results
with ordinary LightGBM. The two penalties have different scales, so equal
$\rho$ values do not mean equal correction strength. We use cross-validation
results to identify exploratory ranges and selected configurations.

We first tuned the LightGBM hyperparameters (994 trees) on the seven folds
using price-scale RMSE, then held them fixed while evaluating and selecting
penalty strengths by cross-validation. Neither test set was used for tuning,
defining the grid, or selection.

For each model and fold, we evaluate prediction, valuation level, ratio
dispersion, vertical equity, and residual structure. We report the mean and
standard deviation across folds for each penalty strength. The ranges and
selections derived from these results are exploratory and do not determine a
penalty strength for deployment.

For concise presentation, we label five tested penalty strengths as
approximately $\rho=0.01$, $0.1$, $1$, $10$, and $100$.
Appendix~\ref{app:ccao_path_details} reports the exact values.

Price-scale measures use directly exponentiated predictions. Small differences
in $R^2_P$ and MAE can therefore reflect changes in overall valuation level
rather than relative valuation accuracy (Appendix~\ref{app:smearing}). For the
ratio-profile figures, we divide each test set into 30 equal-count sale-price
bins and plot the median valuation ratio in each bin. The bins are used only
for visualization; the reported metrics use all sales in the relevant sample.

\paragraph{Uncertainty and stability checks.}
Unless stated otherwise, we measure sampling uncertainty with a paired sale
bootstrap: distinct sales are resampled with replacement, and each
configuration is compared with the $\rho=0$ fit on the same draw. The
bootstrap reuses stored predictions, so it does not capture variability from
refitting the models. To check sensitivity to individual folds, we omit each
of the seven folds in turn (leave-one-fold-out), recompute the equal-weight CV
means from the remaining six, and rerun the CV-based penalty ranges and
selections reported in Section~\ref{sec:results}, again without refitting.
Appendix~\ref{subsec:screening_interpretation} gives replicate counts and
details.

\subsection{Models and Comparisons}
\label{subsec:comparators}

As a supplementary post-hoc method, we rescale fitted log-price predictions:
$f_b(x)=\bar y_T+b\bigl(f_0(x)-\bar y_T\bigr)$, where $f_0$ is the fitted
predictor and $\bar y_T$ is the mean log sale price in the training sample.
At $b=1$, predictions are unchanged; other values of $b$ change their spread
without refitting the model. For the positive values of $b$ used here, the
affine map cannot remove nonlinear residual structure. We calibrate $b$ using out-of-fold predictions
and use the rescaling as both a reference and an add-on to the trained
objectives. Appendix~\ref{app:centered_spread_detail} provides details.

We also compare Direct, Surrogate, and post-hoc rescaling at six first-order
slopes within the range of cross-validation slopes that all three reach. The
targets are slopes achieved by tested Direct configurations; we use the
closest tested Surrogate configuration and calibrate the post-hoc rescaling to
each target. Section~\ref{subsec:matched_beta_results} reports the comparison,
and Appendix~\ref{app:matched_beta_detail} provides details.

We also evaluate Direct and Surrogate in eight other counties using sales and
property-history data from ATTOM. For each county, we tune LightGBM and select
penalty strengths using chronological cross-validation, then evaluate them on
2025 sales. We report penalty effects relative to the custom-objective
LightGBM fit at $\rho=0$. Appendix~\ref{app:external_benchmark} describes the
data and evaluation design.

\section{Results}
\label{sec:results}

We use cross-validation to examine the penalty paths and select exploratory configurations, then evaluate the results on the held-out and 2025-forward test sets. In Cook County, at a penalty strength near $\rho=1$, both methods move all four vertical-equity measures toward neutrality and slightly increase price-scale $R^2_P$ under direct exponentiation in cross-validation and on both test sets, relative to the custom-objective fit at $\rho=0$. At higher penalty strengths, prediction accuracy and ratio dispersion eventually worsen.

\subsection{Penalty-Path Results}
\label{subsec:path_results}

\begin{figure}[!htbp]
\centering
\safeincludegraphics[width=0.8\textwidth]{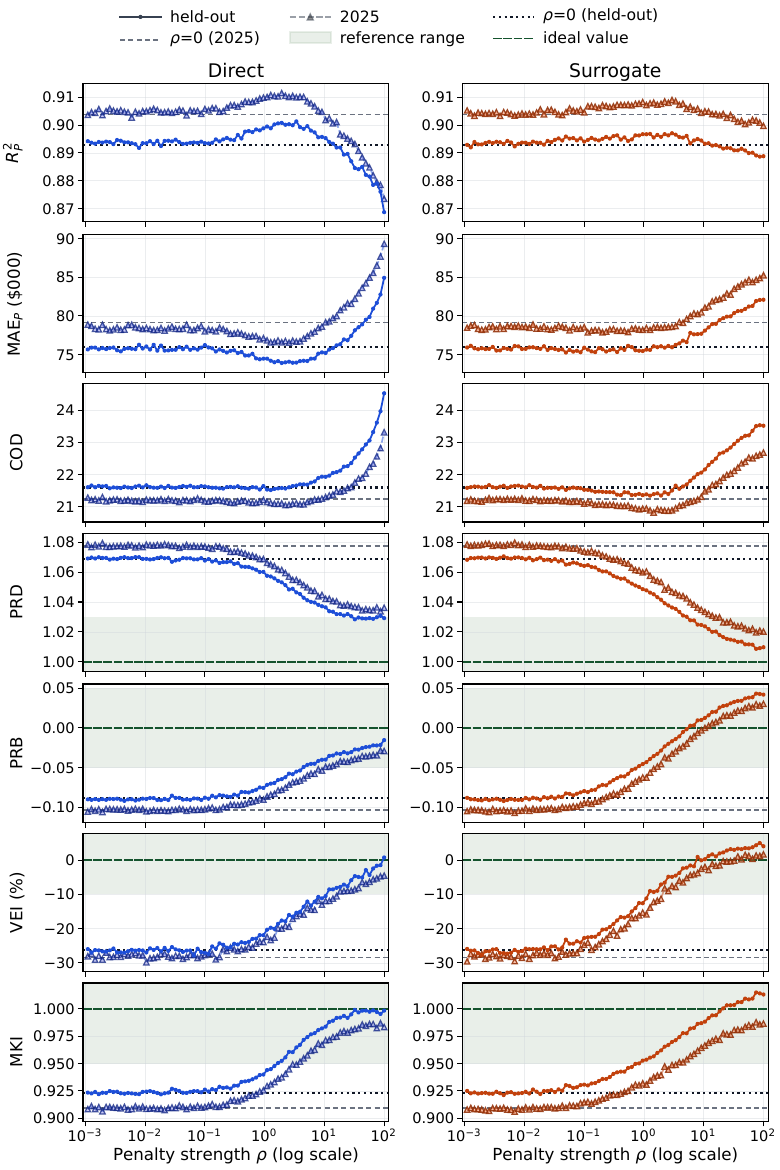}

\caption{Prediction, ratio dispersion, and vertical equity across penalty strengths (log scale). Solid circles show held-out results and dashed triangles show 2025-forward results; blue denotes Direct and orange Surrogate. Dotted and gray dashed horizontal lines show the respective $\rho=0$ values. Green dashed lines mark neutrality, and shading shows the reference ranges of Table~\ref{tab:assessment_metrics_summary} (the MKI and VEI ranges are proposed).}

\label{fig:main_metric_paths}
\end{figure}

Table~\ref{tab:path_anchor_frozen} reports prediction, ratio dispersion, and vertical-equity measures on the held-out and 2025-forward test sets at five penalty strengths. Figure~\ref{fig:main_metric_paths} shows the full range of tested strengths.

\begin{table}[!htbp]
\centering
\begin{minipage}{\textwidth}
\footnotesize

\caption{Prediction, ratio dispersion, and vertical-equity measures at five
penalty strengths on the held-out and 2025-forward test sets.}
\label{tab:path_anchor_frozen}

\setlength{\tabcolsep}{2.6pt}
\renewcommand{\arraystretch}{1.03}

\begin{adjustbox}{max width=\linewidth,center}
\begin{tabular}{@{}ll|rr|r|rrrr@{}}
\toprule
Method & $\rho$
& $R^2_P$ & MAE & COD & PRD & PRB & VEI & MKI \\
\midrule[0.8pt]
\multicolumn{9}{l}{\textit{Panel A: Held-out evaluation}} \\
\midrule[0.8pt]
\addlinespace[2pt]
\multicolumn{9}{l}{\textbf{Reference cells}} \\
\cmidrule(lr){1-9}
Linear regression & --
& 0.799 & \$90,092 & 24.7\% & 1.042 & -0.016 & -11.6\% & 0.975 \\
Ordinary LightGBM & --
& 0.894 & \$75,655 & 21.6\% & 1.069 & -0.091 & -26.5\% & 0.923 \\
Custom-objective (any) & $0$
& 0.893 & \$75,976 & 21.6\% & 1.069 & -0.088 & -26.2\% & 0.923 \\
\addlinespace[3pt]
\multicolumn{9}{l}{\textbf{Direct penalty}} \\
\cmidrule(lr){1-9}
Direct & $0.01$
& 0.893
& \$76,058
& 21.7\%
& 1.069
& -0.089
& -26.9\%
& 0.924 \\
Direct & $0.1$
& 0.894
& \$76,235
& 21.6\%
& 1.068
& -0.088
& -27.0\%
& 0.926 \\
Direct & $1$
& 0.899
& \$74,485
& 21.6\%
& 1.060
& -0.075
& -21.9\%
& 0.940 \\
Direct & $10$
& 0.895
& \$75,218
& 21.9\%
& 1.036
& -0.040
& -10.5\%
& 0.984 \\
Direct & $100$
& 0.869
& \$84,918
& 24.5\%
& 1.029
& -0.015
& 0.7\%
& 0.998 \\
\addlinespace[3pt]
\multicolumn{9}{l}{\textbf{Surrogate penalty}} \\
\cmidrule(lr){1-9}
Surrogate & $0.01$
& 0.894
& \$75,748
& 21.6\%
& 1.069
& -0.088
& -25.8\%
& 0.924 \\
Surrogate & $0.1$
& 0.894
& \$75,897
& 21.5\%
& 1.064
& -0.079
& -22.7\%
& 0.931 \\
Surrogate & $1$
& 0.897
& \$75,468
& 21.4\%
& 1.049
& -0.045
& -12.6\%
& 0.953 \\
Surrogate & $10$
& 0.893
& \$78,110
& 22.2\%
& 1.024
& 0.013
& 0.2\%
& 0.988 \\
Surrogate & $100$
& 0.889
& \$82,094
& 23.5\%
& 1.010
& 0.042
& 4.0\%
& 1.013 \\
\midrule[0.8pt]
\addlinespace[2pt]
\multicolumn{9}{l}{\textit{Panel B: 2025-forward evaluation}} \\
\midrule[0.8pt]
\addlinespace[2pt]
\multicolumn{9}{l}{\textbf{Reference cells}} \\
\cmidrule(lr){1-9}
Linear regression & --
& 0.799 & \$99,371 & 24.3\% & 1.052 & -0.029 & -17.3\% & 0.954 \\
Ordinary LightGBM & --
& 0.904 & \$78,484 & 21.3\% & 1.079 & -0.106 & -28.6\% & 0.907 \\
Custom-objective (any) & 0
& 0.904 & \$79,139 & 21.2\% & 1.078 & -0.103 & -28.4\% & 0.909 \\
\addlinespace[3pt]
\multicolumn{9}{l}{\textbf{Direct penalty}} \\
\cmidrule(lr){1-9}
Direct & $0.01$
& 0.905
& \$78,442
& 21.2\%
& 1.078
& -0.105
& -29.8\%
& 0.909 \\
Direct & $0.1$
& 0.906
& \$78,029
& 21.2\%
& 1.077
& -0.102
& -27.7\%
& 0.910 \\
Direct & $1$
& 0.910
& \$77,139
& 21.2\%
& 1.069
& -0.090
& -23.9\%
& 0.926 \\
Direct & $10$
& 0.902
& \$79,187
& 21.3\%
& 1.042
& -0.049
& -11.9\%
& 0.971 \\
Direct & $100$
& 0.873
& \$89,309
& 23.3\%
& 1.036
& -0.029
& -4.5\%
& 0.983 \\
\addlinespace[3pt]
\multicolumn{9}{l}{\textbf{Surrogate penalty}} \\
\cmidrule(lr){1-9}
Surrogate & $0.01$
& 0.904
& \$78,620
& 21.2\%
& 1.077
& -0.101
& -27.4\%
& 0.910 \\
Surrogate & $0.1$
& 0.905
& \$78,546
& 21.1\%
& 1.074
& -0.095
& -24.7\%
& 0.914 \\
Surrogate & $1$
& 0.908
& \$78,196
& 20.9\%
& 1.060
& -0.062
& -15.3\%
& 0.932 \\
Surrogate & $10$
& 0.905
& \$81,075
& 21.5\%
& 1.033
& 0.000
& -2.0\%
& 0.948 \\
Surrogate & $100$
& 0.900
& \$85,257
& 22.7\%
& 1.020
& 0.030
& 1.7\%
& 0.987 \\
\bottomrule
\end{tabular}
\end{adjustbox}

\vspace{1mm}

\emph{Notes.} Higher $R^2_P$ and lower MAE and COD are favorable; PRD and MKI are neutral at one, and PRB and VEI at zero. Equal $\rho$ values do not imply equal correction strength across methods.

\end{minipage}
\end{table}

Table~\ref{tab:path_anchor_frozen} shows that, on both test sets, the penalties generally move the vertical-equity measures toward neutrality, even though neither objective directly optimizes them. However, the changes are not uniform: Direct's VEI initially becomes more negative, and Surrogate's PRB passes zero and moves away from it at larger penalties. Stronger penalties can also worsen prediction accuracy and ratio dispersion, consistent with the expected tradeoff.

At $\rho\approx 1$, Direct increases held-out $R^2_P$ from $0.893$ to $0.899$ and reduces MAE from \$75,976 to \$74,485, relative to $\rho=0$. COD remains unchanged at the displayed precision, and all four vertical-equity measures move closer to neutrality. For example, PRB moves from $-0.088$ to $-0.075$. The 2025-forward results change in the same directions.

At $\rho\approx 1$, Surrogate increases held-out $R^2_P$ from $0.893$ to $0.897$, reduces MAE from \$75,976 to \$75,468, and reduces COD from $21.6\%$ to $21.4\%$. PRB moves from $-0.088$ to $-0.045$, a reduction of about 49\% in magnitude, and the other three vertical-equity measures also move closer to neutrality. The 2025-forward results change in the same directions.

At $\rho\approx 10$, Direct's PRB and MKI fall within their reference ranges on both test sets, while its PRD and VEI remain outside. Surrogate's PRB and VEI fall within their ranges on both sets; its PRD and MKI fall within them on the held-out set but just outside on the 2025-forward set. The PRD and PRB ranges are adopted guidance, whereas the MKI and VEI ranges are proposed. MAE and COD are already increasing for both methods.

At $\rho=100$, both methods show clear costs on the held-out test set. Direct and Surrogate have $R^2_P$ values of $0.869$ and $0.889$, compared with $0.893$ at $\rho=0$; MAE rises to \$84,918 and \$82,094, and COD rises to $24.5\%$ and $23.5\%$, respectively. Both PRB magnitudes remain below their $\rho=0$ values, but Surrogate's positive PRB is farther from zero than at $\rho\approx 10$. MAE and COD also increase on the 2025-forward test set.

\paragraph{Full penalty paths.}
Figure~\ref{fig:main_metric_paths} shows how the measures in Table~\ref{tab:path_anchor_frozen} change across all tested penalty strengths. Both methods generally move the vertical-equity measures toward neutrality. Changes are small below about $\rho=0.1$ for Direct and $\rho=0.01$ for Surrogate; intermediate penalties slightly raise $R^2_P$ and lower MAE, and stronger penalties reverse these changes and raise COD. The ratio profiles below show how valuation ratios vary across sale prices, including patterns that aggregate measures can miss.

Appendix Figure~\ref{fig:other_metric_paths_placeholder} shows the remaining prediction, valuation-level, and dispersion measures across the penalty strengths.

\subsection{Accuracy and Vertical-Equity Tradeoffs}
\label{subsec:tradeoff_results}

\begin{figure}[!htbp]
\centering
\safeincludegraphics[width=0.9\textwidth]{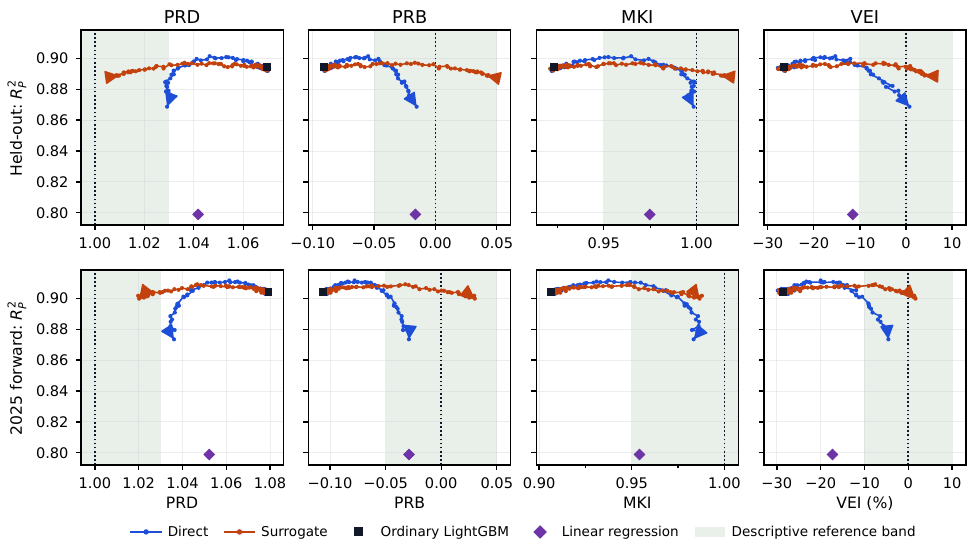}
\caption{Price-scale $R^2_P$ and vertical-equity measures across penalty strengths on the held-out (top) and 2025-forward (bottom) test sets. Blue shows Direct and orange Surrogate. Arrows mark $\rho=100$, the strongest tested penalty. Squares and diamonds mark ordinary LightGBM and the linear benchmark. Dotted lines mark neutrality, and shading shows the reference ranges of Table~\ref{tab:assessment_metrics_summary} (MKI and VEI proposed).}
\label{fig:accuracy_equity_placeholder}
\end{figure}

Figure~\ref{fig:accuracy_equity_placeholder} shows the tradeoffs between $R^2_P$ and the four vertical-equity measures. At some penalty strengths, both methods move the measures toward neutrality while keeping $R^2_P$ at or slightly above its $\rho=0$ value. At stronger penalties, Surrogate reaches all four reference ranges on both test sets; Direct reaches them on the held-out set, but its 2025-forward PRD remains outside the adopted range. The figure suggests that Direct offers more favorable combinations at some modest correction levels, while Surrogate's $R^2_P$ declines less sharply near the strongest tested penalties. Neither method is consistently better across measures and test sets.

\subsection{Valuation-Ratio Profiles}
\label{subsec:ratio_shape_results}

\begin{figure}[!htbp]
\centering
\safeincludegraphics[width=0.85\textwidth]{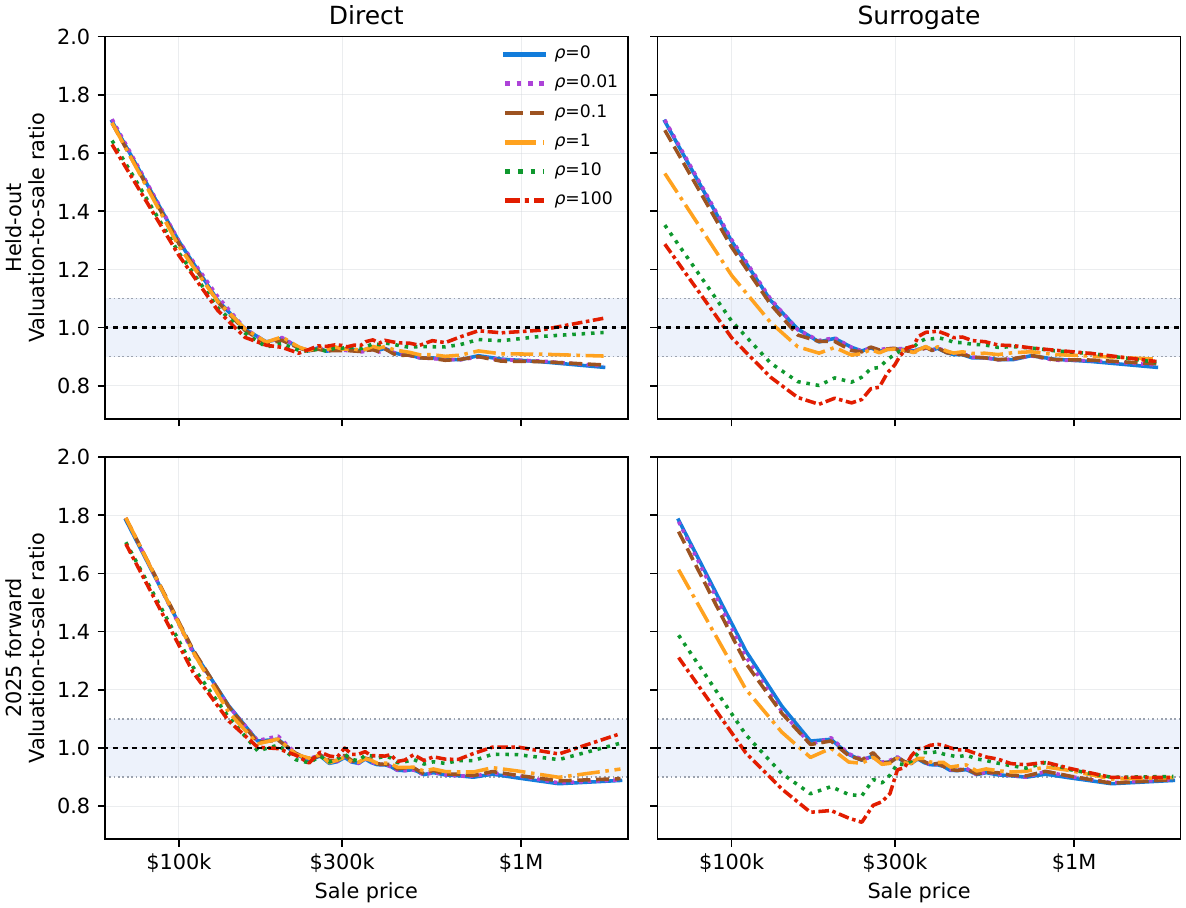}
\caption{Median valuation-to-sale ratios in 30 equal-count sale-price bins for the $\rho=0$ fit and five penalty strengths per method, on the held-out and 2025-forward test sets. The line at 1 indicates equal valuation and sale price; 0.9 and 1.1 are aggregate reference values.}
\label{fig:ratio_shape_anchor_profiles}
\end{figure}

Figure~\ref{fig:ratio_shape_anchor_profiles} shows median valuation ratios across 30 equal-count sale-price bins for the $\rho=0$ fit and five penalty strengths for each method. Both methods value the lowest-priced properties well above their sale prices, although Surrogate reduces this difference as the penalty increases. At stronger penalties, Direct's profile turns upward in the upper price range, while Surrogate's ratios fall substantially below one for some lower- and middle-priced bins before rising again, forming a trough in the profile. These patterns show that aggregate measures near their neutral values can hide overvaluation and undervaluation at different price levels.

Under direct exponentiation, most profiles lie below one over much of the price range. Smearing multiplies a model's price predictions by a common factor estimated from out-of-fold residuals, shifting its ratio profile without changing its relative shape. For Surrogate at $\rho\approx1$, smearing moves the held-out median ratio from 0.929 to 1.000, but raises the lowest-price bin's median ratio from 1.53 to 1.65 (Appendix~\ref{app:smearing}).

\subsection{Cross-Validation Paths and Penalty Selection}

\begin{figure}[!htbp]
\centering
\safeincludegraphics[width=4.6in]{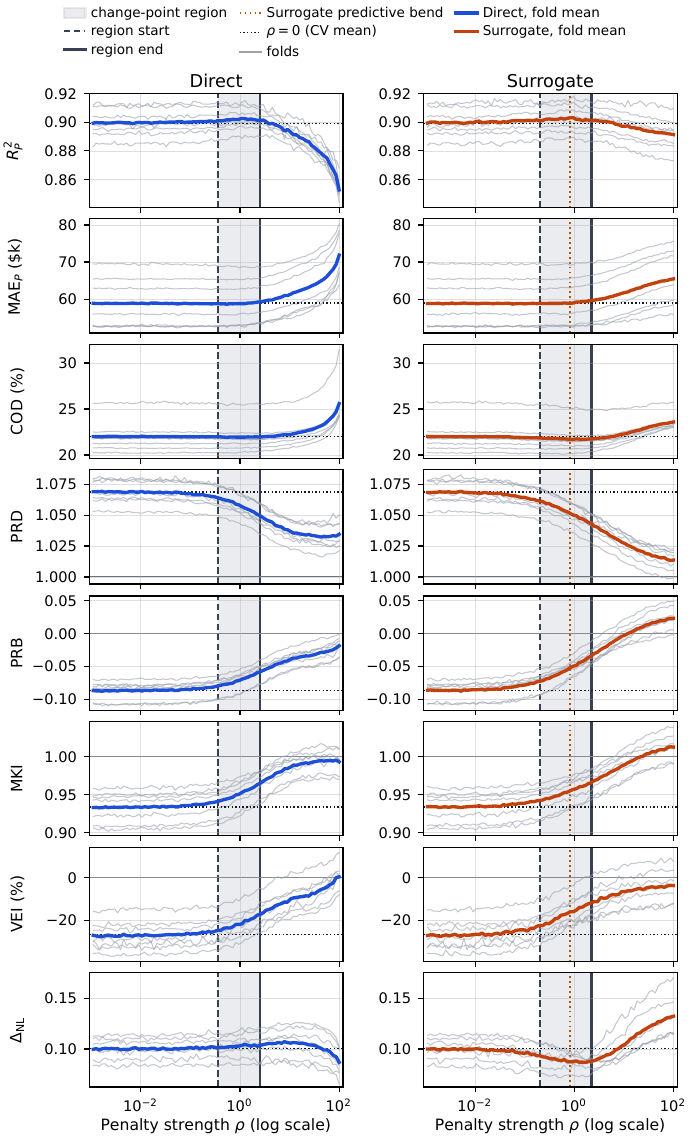}
\caption{Cross-validation results across penalty strengths for eight measures. Thin gray curves show individual folds, and thick colored curves show their equal-weight averages. The gray shaded area marks the change-point region, with dashed and solid vertical lines marking its start and end. The start also uses $\beta_{\log}$, shown in Appendix Figure~\ref{fig:cv_mechanism_metric_paths}. The dotted orange vertical line marks an earlier change in Surrogate's prediction trend. Its region instead ends at a later rise in the nonlinearity gap. The shaded region is distinct from the tolerance band used for selection in Table~\ref{tab:cv_selected_comparison}. Horizontal dotted lines show the $\rho=0$ averages, and gray horizontal lines mark neutrality where applicable.}
\label{fig:candidate_region_cv}
\end{figure}

Figure~\ref{fig:candidate_region_cv} shows the results across the seven cross-validation folds. Thin gray lines show individual folds, and the colored lines show the average for each method. The broad patterns are similar across folds and in the held-out and 2025-forward test sets, although the values and turning points vary. We use these results first to describe where the paths change and then to select exploratory penalties under explicit cost limits.

The shaded change-point regions in Figure~\ref{fig:candidate_region_cv} describe changes in the cross-validation paths. Each region starts where at least three of PRD, PRB, MKI, VEI, and $\beta_{\log}$ show an improving trend toward neutrality. For Direct, it ends where at least two prediction measures start a sustained deterioration. For Surrogate, it ends where the nonlinearity gap begins a stable rise after its minimum. The resulting regions are $[0.356,2.560]$ for Direct and $[0.202,2.223]$ for Surrogate. They describe the paths and are not used to select penalty strengths. Appendix~\ref{subsec:screening_interpretation} gives the full rule and results.

\paragraph{Nonlinear conditional-mean diagnostic.}
The nonlinearity gap $\Delta_{\mathrm{NL}}=\eta^2(e\mid Z)-\operatorname{Corr}(e,Z)^2$ measures conditional-mean structure beyond the best straight-line relationship (Section~\ref{subsec:metrics}). Its empirical spline estimate complements the first-order slope and ratio profiles; it does not measure every form of dependence. In CV, Direct's gap rises at intermediate penalties before falling at the largest values. Surrogate's gap initially falls, then rebounds as the penalty increases. Within its change-point region, a local ratio-profile dip deepens even while the global gap remains below its $\rho=0$ value (Appendix Figure~\ref{fig:candidate_region_ratio_profiles}). A smaller global gap therefore does not rule out an important local pattern.

The main costs differ between methods. Stronger Direct penalties eventually worsen prediction accuracy and dispersion, while stronger Surrogate penalties can deepen the ratio-profile trough. Surrogate's log-price RMSE also increases within its change-point region, even though its price-scale $R^2_P$ is higher under direct exponentiation (Appendix~\ref{app:smearing}).

\paragraph{Tolerance band and selection rules.}
\label{subsec:candidate_region_results}

To select penalties, we instead use a cross-validation tolerance band with explicit limits on prediction, dispersion, nonlinear structure, and valuation level, while the ratio profiles remain a separate check. For this illustration, the band starts at the first tested penalty that reduces $|\mathrm{PRB}|$ by at least 10\% and meets the cost limits, and it ends before the first subsequent tested penalty that exceeds a limit. All limits apply to equal-weight CV means. Relative to the $\rho=0$ fit, increases are limited to 2\% for $1-R^2_P$, MAE, MAPE, log-price RMSE, and COD, and to 10\% for $\Delta_{\mathrm{NL}}$. To limit changes in overall valuation level, the median ratio's distance from one may increase by at most 0.01.

The resulting bands are $[0.471,4.498]$ for Direct and $[0.115,0.954]$ for Surrogate. The next tested penalty exceeds the MAE limit for Direct and the log-price RMSE limit for Surrogate. Direct's band extends beyond its change-point region because prediction begins to worsen before a cost exceeds its limit. Surrogate's band ends earlier because log-price RMSE exceeds its limit before the nonlinearity gap reaches its minimum.

Unlike the change-point regions in Figure~\ref{fig:candidate_region_cv}, a tolerance band is found for both methods in Cook County and all eight external counties (Appendix~\ref{app:external_benchmark}). The upper endpoints vary by at most two tested penalty steps across the 5th--95th percentile bootstrap range and all leave-one-fold-out recalculations (Appendix~\ref{subsec:screening_interpretation}). Decision makers can choose different improvement targets and cost limits.

Within each band, we illustrate two ways to choose a tested penalty for each method using the equal-weight cross-validation results. The \emph{constrained} rule selects the smallest penalty that reduces $|\mathrm{PRB}|$ by at least 25\%, or makes no selection if none qualifies. The \emph{Nash} rule balances prediction, dispersion, regressivity, and nonlinear structure. It measures regressivity gains from $\rho=0$ and compares each cost with its chosen limit. Unlike the constrained rule, Nash does not require a separate PRB target, but it still depends on the measures, limits, and how they are combined. Table~\ref{tab:cv_selected_comparison} compares the four selected configurations (two for each method) against the $\rho=0$ fit. Appendix Table~\ref{tab:cv_selection_rules} reports the chosen penalties, PRB reductions, and stability checks for these and additional rules.

\begin{table}[!htbp]
\centering
\small
\caption{Comparison of the configurations selected by the constrained and Nash rules against the unpenalized custom-objective fit on the held-out (Panel A) and 2025-forward (Panel B) test sets.}
\label{tab:cv_selected_comparison}

\setlength{\tabcolsep}{3pt}
\renewcommand{\arraystretch}{1.13}

\resizebox{\textwidth}{!}{%
\begin{tabular}{@{}l*{9}{r}@{}}
\toprule
& & \multicolumn{4}{c}{\textbf{Direct}} & \multicolumn{4}{c}{\textbf{Surrogate}} \\
\cmidrule(lr){3-6}\cmidrule(l){7-10}
& & \multicolumn{2}{c}{Constrained} & \multicolumn{2}{c}{Nash} & \multicolumn{2}{c}{Constrained} & \multicolumn{2}{c}{Nash} \\
\cmidrule(lr){3-4}\cmidrule(lr){5-6}\cmidrule(lr){7-8}\cmidrule(l){9-10}
Metric & $\rho=0$ & $\rho=1.677$ & Improve. & $\rho=2.223$ & Improve. & $\rho=0.409$ & Improve. & $\rho=0.954$ & Improve. \\
\midrule
\multicolumn{10}{@{}l}{\textit{Panel A: Held-out evaluation}} \\
\midrule
$R^2_P$ & 0.8928 & 0.9006 & $(+7.3\%)$ & 0.9001 & $(+6.8\%)$ & 0.8948 & $(+1.8\%)$ & 0.8967 & $(+3.7\%)$ \\
MAE (\$) & 75,976 & 74,167 & $(+2.4\%)$ & 74,008 & $(+2.6\%)$ & 75,831 & $(+0.2\%)$ & 75,468 & $(+0.7\%)$ \\
$\operatorname{RMSE}_{\log P}$ & 0.2894 & 0.2900 & $(-0.2\%)$ & 0.2896 & $(-0.1\%)$ & 0.2948 & $(-1.9\%)$ & 0.2982 & $(-3.1\%)$ \\
\midrule
COD & 21.60\% & 21.57\% & $(+0.1\%)$ & 21.57\% & $(+0.2\%)$ & 21.36\% & $(+1.1\%)$ & 21.36\% & $(+1.1\%)$ \\
\midrule
PRD & 1.0690 & 1.0541 & $(\mathbf{+21.5\%})$ & 1.0512 & $(\mathbf{+25.7\%})$ & 1.0555 & $(\mathbf{+19.5\%})$ & 1.0486 & $(\mathbf{+29.6\%})$ \\
PRB & $-0.0884$ & $-0.0655$ & $(\mathbf{+26.0\%})$ & $-0.0620$ & $(\mathbf{+29.9\%})$ & $-0.0609$ & $(\mathbf{+31.2\%})$ & $-0.0454^{\ddagger}$ & $(\mathbf{+48.6\%})$ \\
VEI & $-26.17\%$ & $-18.59\%$ & $(\mathbf{+29.0\%})$ & $-17.85\%$ & $(\mathbf{+31.8\%})$ & $-16.62\%$ & $(\mathbf{+36.5\%})$ & $-12.60\%$ & $(\mathbf{+51.8\%})$ \\
MKI & 0.9230 & $0.9505^{\ddagger}$ & $(\mathbf{+35.6\%})$ & $0.9556^{\ddagger}$ & $(\mathbf{+42.3\%})$ & 0.9426 & $(\mathbf{+25.4\%})$ & $0.9527^{\ddagger}$ & $(\mathbf{+38.5\%})$ \\
\midrule
$\Delta_{\mathrm{NL}}$ & 0.1164 & 0.1231 & $(-5.8\%)$ & 0.1247 & $(-7.1\%)$ & 0.1035 & $(\mathbf{+11.2\%})$ & 0.0991 & $(\mathbf{+14.9\%})$ \\
\midrule[0.8pt]
\multicolumn{10}{@{}l}{\textit{Panel B: 2025-forward evaluation}} \\
\midrule
$R^2_P$ & 0.9039 & 0.9104 & $(+6.8\%)$ & 0.9105 & $(+6.9\%)$ & 0.9073 & $(+3.6\%)$ & 0.9083 & $(+4.6\%)$ \\
MAE (\$) & 79,139 & 76,831 & $(+2.9\%)$ & 76,755 & $(+3.0\%)$ & 78,121 & $(+1.3\%)$ & 78,196 & $(+1.2\%)$ \\
$\operatorname{RMSE}_{\log P}$ & 0.2791 & 0.2780 & $(+0.4\%)$ & 0.2781 & $(+0.4\%)$ & 0.2805 & $(-0.5\%)$ & 0.2837 & $(-1.7\%)$ \\
\midrule
COD & 21.23\% & 21.09\% & $(+0.7\%)$ & 21.05\% & $(+0.9\%)$ & 21.03\% & $(+1.0\%)$ & 20.94\% & $(+1.4\%)$ \\
\midrule
PRD & 1.0775 & 1.0618 & $(\mathbf{+20.2\%})$ & 1.0594 & $(\mathbf{+23.3\%})$ & 1.0674 & $(\mathbf{+13.0\%})$ & 1.0598 & $(\mathbf{+22.8\%})$ \\
PRB & $-0.1028$ & $-0.0785$ & $(\mathbf{+23.6\%})$ & $-0.0751$ & $(\mathbf{+27.0\%})$ & $-0.0803$ & $(\mathbf{+22.0\%})$ & $-0.0623$ & $(\mathbf{+39.4\%})$ \\
VEI & $-28.41\%$ & $-19.90\%$ & $(\mathbf{+30.0\%})$ & $-19.14\%$ & $(\mathbf{+32.6\%})$ & $-20.15\%$ & $(\mathbf{+29.1\%})$ & $-15.32\%$ & $(\mathbf{+46.1\%})$ \\
MKI & 0.9093 & 0.9362 & $(\mathbf{+29.6\%})$ & 0.9405 & $(\mathbf{+34.4\%})$ & 0.9226 & $(\mathbf{+14.6\%})$ & 0.9319 & $(\mathbf{+24.9\%})$ \\
\midrule
$\Delta_{\mathrm{NL}}$ & 0.1212 & 0.1254 & $(-3.5\%)$ & 0.1263 & $(-4.2\%)$ & 0.1061 & $(\mathbf{+12.4\%})$ & 0.0963 & $(\mathbf{+20.5\%})$ \\
\bottomrule
\end{tabular}%
}

\vspace{3pt}
\parbox{\textwidth}{\footnotesize
\emph{Notes.} Percentages compare each configuration with the $\rho=0$ fit. Positive values indicate improvement, while negative values indicate worsening. For $R^2_P$, the percentage is the relative reduction in price-scale mean squared error, calculated as $(R^2_{P,\rho}-R^2_{P,0})/(1-R^2_{P,0})$. For MAE, log-price RMSE, COD, and $\Delta_{\mathrm{NL}}$, percentages show relative reductions in each measure. For PRD, PRB, VEI, and MKI, they show relative reductions in distance from neutrality. $R^2_P$ and MAE use directly exponentiated predictions. Percentages can differ slightly from calculations using the rounded displayed values. Bold indicates improvement above 10\%, and \ddag\ indicates a point estimate within its reference range (Table~\ref{tab:assessment_metrics_summary}).
}
\end{table}

\begin{figure}[!htbp]
\centering
\safeincludegraphics[width=\textwidth]{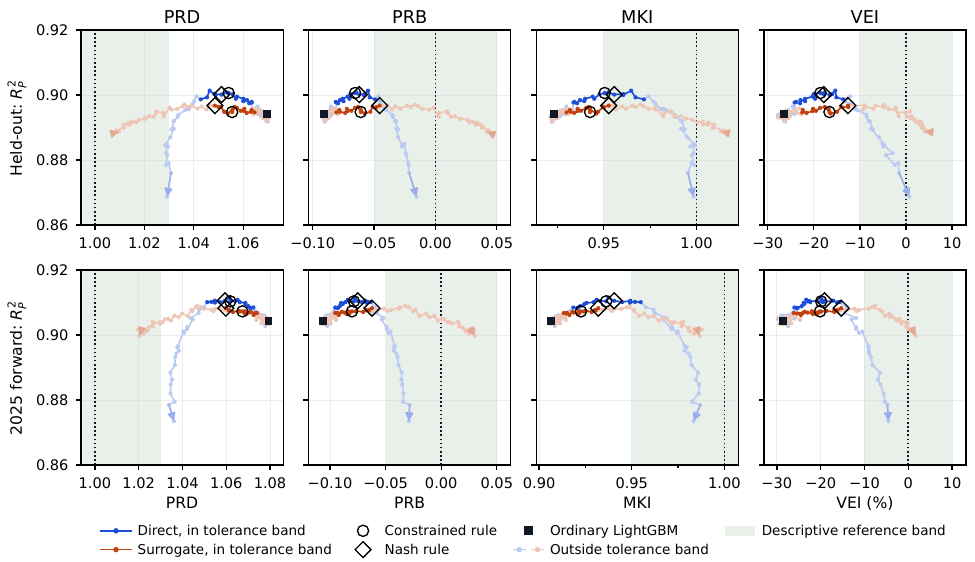}
\caption{Accuracy--equity paths on the held-out (top) and 2025-forward (bottom) test sets. The highlighted segments show the cross-validation tolerance bands: $\rho\in[0.471,4.498]$ for Direct and $[0.115,0.954]$ for Surrogate. Open circles and diamonds mark the constrained and Nash selections in Table~\ref{tab:cv_selected_comparison}. Black squares mark ordinary LightGBM; the linear benchmark is omitted. Shading and neutrality lines are as in Figure~\ref{fig:accuracy_equity_placeholder}. The restricted $R^2_P$ axis enlarges differences.}
\label{fig:accuracy_equity_cv_selected}
\end{figure}

\paragraph{Exploratory CV-selected configurations.}
On the held-out and 2025-forward test sets, all four selected configurations lower COD and move PRD, PRB, VEI, and MKI toward neutrality. The constrained selections reduce $|\mathrm{PRB}|$ by 22--31\% across the two methods and test sets. Under Nash, Surrogate reduces it by 49\% and 39\%, compared with 30\% and 27\% for Direct. The price-scale $R^2_P$ and MAE gains partly reflect overall valuation level. The $R^2_P$ gains shrink or reverse under level adjustments (Appendix Table~\ref{tab:level_cv_selected}). Figure~\ref{fig:accuracy_equity_cv_selected} shows the tolerance bands and the four selected configurations along the accuracy--equity paths.

\begin{figure}[!htbp]
\centering
\safeincludegraphics[width=0.75\textwidth]{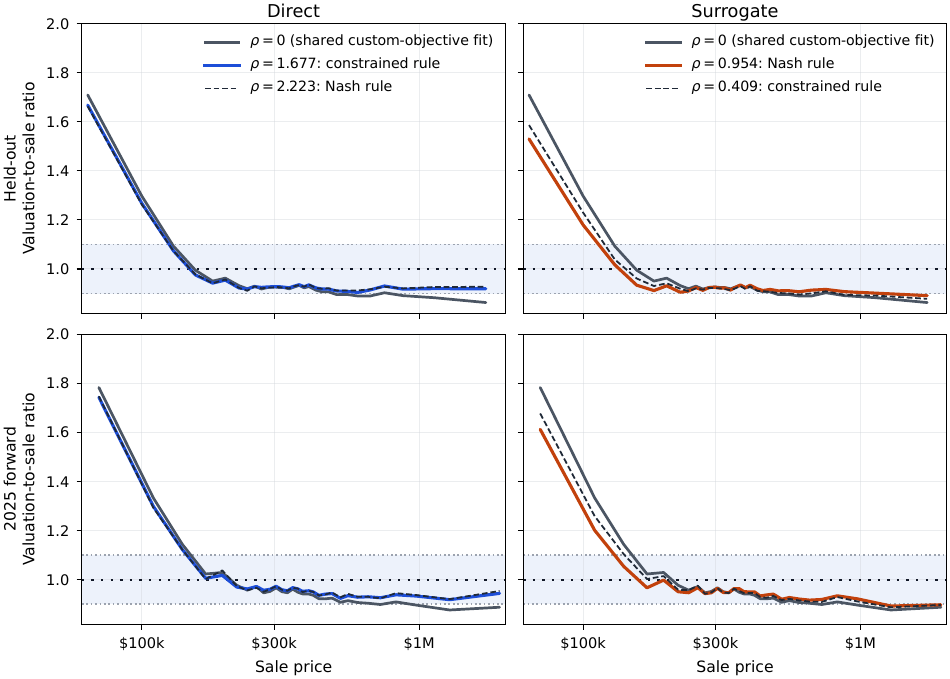}
\caption{Median valuation-to-sale ratios in 30 equal-count sale-price bins for the $\rho=0$ fit and the four configurations in Table~\ref{tab:cv_selected_comparison}. Within each test set, all configurations use the same bins, plotted at their median sale prices. The line at 1 indicates equal valuation and sale price; 0.9 and 1.1 are aggregate reference values.}
\label{fig:ratio_shape_path_placeholder}
\end{figure}

The costs differ: Direct increases the nonlinearity gap, while Surrogate decreases it but increases log-price RMSE. For the Surrogate Nash selection, held-out log-price RMSE rises by 3.1\%, exceeding the 2\% limit met in cross-validation. Figure~\ref{fig:ratio_shape_path_placeholder} also shows that Surrogate reduces overvaluation in the lowest-price bin more than Direct, although the bin's median ratio remains well above one.

\paragraph{Sampling uncertainty.} To assess uncertainty in these changes, we randomly resample sales with replacement 200 times within each test set and compare each selected configuration with the $\rho=0$ fit on the same resampled sales. The resulting 90\% bootstrap confidence intervals exclude zero for the PRD, PRB, VEI, and MKI improvements at all four configurations on both test sets. For Surrogate at $\rho=0.954$, the PRB reduction is 48.6\% (90\% CI: 46.8\%--50.8\%) on the held-out set and 39.4\% (37.7\%--41.4\%) on the 2025-forward set. For Direct at $\rho=1.677$, the corresponding reductions are 26.0\% (24.8\%--27.6\%) and 23.6\% (22.4\%--24.9\%). The intervals also exclude zero for the increase in log-price RMSE under both Surrogate selections and the increase in the nonlinearity gap under both Direct selections.

\subsection{Post-hoc Rescaling as an Add-on}
\label{subsec:posthoc_comparator}

The post-hoc rescaling defined in Section~\ref{subsec:comparators} changes the spread of fitted log-price predictions without refitting the model. Its scale is set from out-of-fold cross-validation predictions (Appendix~\ref{app:centered_spread_detail}). Used alone to bring the cross-validation slope of the $\rho=0$ fit to zero, it requires $b\approx1.160$. At that setting, held-out $R^2_P$ falls from $0.8928$ to $0.8063$, MAE rises from \$75,976 to \$87,182, and $\Delta_{\mathrm{NL}}$ rises from $0.1164$ to $0.1343$ (Appendix Table~\ref{tab:centered_spread}). This setting therefore has substantial costs relative to the unpenalized fit.

A smaller rescaling can instead be added to a trained objective. Combining Direct $\rho=1.677$ or Surrogate $\rho=0.954$ with $b=1.027$ moves PRB further toward zero than the corresponding trained model alone on both test sets (Table~\ref{tab:joint_posthoc_comparison}). The Surrogate combination retains a lower $\Delta_{\mathrm{NL}}$ than the unpenalized fit, although its log-price RMSE remains higher. Selecting $b$ jointly with $\rho$ under the tolerance rules generally changes the $|\mathrm{PRB}|$ reduction only modestly and makes the selections less stable. These combinations are exploratory.

\begin{table}[!htbp]
\centering
\footnotesize
\setlength{\tabcolsep}{3pt}
\renewcommand{\arraystretch}{1.10}
\newcommand{\jointgain}[1]{\textcolor{green!50!black}{(#1)}}
\newcommand{\jointloss}[1]{\textcolor{red!70!black}{(#1)}}
\newcommand{\jointbest}[1]{\textcolor{green!50!black}{\textbf{(#1)}}}
\newcommand{\jointchange}[1]{{\fontsize{6.5}{7.5}\selectfont #1}}
\caption{Comparison of the trained models and post-hoc rescaling with the unpenalized fit on the held-out and 2025-forward test sets.}
\label{tab:joint_posthoc_comparison}
\begin{tabular}{@{}l*{8}{c}@{}}
\toprule
Model & \shortstack[c]{$R^2_P$\\[-1pt]{\scriptsize $\Delta$ (pp)}} & \shortstack[c]{MAE (\$)\\[-1pt]{\scriptsize $\Delta$ (\%)}} & \shortstack[c]{COD\\[-1pt]{\scriptsize $\Delta$ (\%)}} & \shortstack[c]{PRD\\[-1pt]{\scriptsize $\Delta$ (\%)}} & \shortstack[c]{PRB\\[-1pt]{\scriptsize $\Delta$ (\%)}} & \shortstack[c]{VEI\\[-1pt]{\scriptsize $\Delta$ (\%)}} & \shortstack[c]{MKI\\[-1pt]{\scriptsize $\Delta$ (\%)}} & \shortstack[c]{$\Delta_{\mathrm{NL}}$\\[-1pt]{\scriptsize $\Delta$ (\%)}} \\
\midrule
\multicolumn{9}{@{}l}{\quad \textit{Panel A: Held-out evaluation}} \\
\midrule
Baseline & 0.8928 & 75,976 & 21.60 & 1.0690 & $-0.0884$ & $-26.17$ & 0.9230 & 0.1164 \\
Direct & \textbf{0.9006} & 74,167 & 21.57 & 1.0541 & $-0.0655$ & $-18.59$ & 0.9505 & 0.1231 \\[-2pt]
 & \jointchange{\jointbest{+0.78\,pp}} & \jointchange{\jointgain{+2.4\%}} & \jointchange{\jointgain{+0.1\%}} & \jointchange{\jointgain{+21.5\%}} & \jointchange{\jointgain{+26.0\%}} & \jointchange{\jointgain{+29.0\%}} & \jointchange{\jointgain{+35.6\%}} & \jointchange{\jointloss{-5.8\%}} \\
Surrogate & 0.8967 & 75,468 & \textbf{21.36} & 1.0486 & $-0.0454$ & $-12.60$ & 0.9527 & \textbf{0.0991} \\[-2pt]
 & \jointchange{\jointgain{+0.39\,pp}} & \jointchange{\jointgain{+0.7\%}} & \jointchange{\jointbest{+1.1\%}} & \jointchange{\jointgain{+29.6\%}} & \jointchange{\jointgain{+48.6\%}} & \jointchange{\jointgain{+51.8\%}} & \jointchange{\jointgain{+38.5\%}} & \jointchange{\jointbest{+14.9\%}} \\
Post-hoc & 0.9000 & 74,151 & 21.52 & 1.0548 & $-0.0663$ & $-19.09$ & 0.9486 & 0.1218 \\[-2pt]
 & \jointchange{\jointgain{+0.72\,pp}} & \jointchange{\jointgain{+2.4\%}} & \jointchange{\jointgain{+0.3\%}} & \jointchange{\jointgain{+20.6\%}} & \jointchange{\jointgain{+25.0\%}} & \jointchange{\jointgain{+27.1\%}} & \jointchange{\jointgain{+33.2\%}} & \jointchange{\jointloss{-4.6\%}} \\
Direct + post-hoc & 0.9004 & \textbf{73,903} & 21.60 & 1.0396 & $-0.0434$ & $-11.38$ & 0.9767 & 0.1277 \\[-2pt]
 & \jointchange{\jointgain{+0.76\,pp}} & \jointchange{\jointbest{+2.7\%}} & \jointchange{(+0.0\%)} & \jointchange{\jointgain{+42.5\%}} & \jointchange{\jointgain{+50.9\%}} & \jointchange{\jointgain{+56.5\%}} & \jointchange{\jointgain{+69.8\%}} & \jointchange{\jointloss{-9.6\%}} \\
Surrogate + post-hoc & \textbf{0.9006} & 74,411 & 21.43 & \textbf{1.0344} & $\mathbf{-0.0242}$ & $\mathbf{-5.43}$ & \textbf{0.9785} & 0.1019 \\[-2pt]
 & \jointchange{\jointbest{+0.78\,pp}} & \jointchange{\jointgain{+2.1\%}} & \jointchange{\jointgain{+0.8\%}} & \jointchange{\jointbest{+50.1\%}} & \jointchange{\jointbest{+72.6\%}} & \jointchange{\jointbest{+79.3\%}} & \jointchange{\jointbest{+72.0\%}} & \jointchange{\jointgain{+12.5\%}} \\
\addlinespace[3pt]
\midrule
\multicolumn{9}{@{}l}{\quad \textit{Panel B: 2025-forward evaluation}} \\
\midrule
Baseline & 0.9039 & 79,139 & 21.23 & 1.0775 & $-0.1028$ & $-28.41$ & 0.9093 & 0.1212 \\
Direct & 0.9104 & \textbf{76,831} & 21.09 & 1.0618 & $-0.0785$ & $-19.90$ & 0.9362 & 0.1254 \\[-2pt]
 & \jointchange{\jointgain{+0.65\,pp}} & \jointchange{\jointbest{+2.9\%}} & \jointchange{\jointgain{+0.7\%}} & \jointchange{\jointgain{+20.2\%}} & \jointchange{\jointgain{+23.6\%}} & \jointchange{\jointgain{+30.0\%}} & \jointchange{\jointgain{+29.6\%}} & \jointchange{\jointloss{-3.5\%}} \\
Surrogate & 0.9083 & 78,196 & 20.94 & 1.0598 & $-0.0623$ & $-15.32$ & 0.9319 & \textbf{0.0963} \\[-2pt]
 & \jointchange{\jointgain{+0.44\,pp}} & \jointchange{\jointgain{+1.2\%}} & \jointchange{\jointgain{+1.4\%}} & \jointchange{\jointgain{+22.8\%}} & \jointchange{\jointgain{+39.4\%}} & \jointchange{\jointgain{+46.1\%}} & \jointchange{\jointgain{+24.9\%}} & \jointchange{\jointbest{+20.5\%}} \\
Post-hoc & 0.9106 & 77,027 & 21.09 & 1.0630 & $-0.0807$ & $-21.42$ & 0.9348 & 0.1278 \\[-2pt]
 & \jointchange{\jointgain{+0.67\,pp}} & \jointchange{\jointgain{+2.7\%}} & \jointchange{\jointgain{+0.7\%}} & \jointchange{\jointgain{+18.8\%}} & \jointchange{\jointgain{+21.5\%}} & \jointchange{\jointgain{+24.6\%}} & \jointchange{\jointgain{+28.0\%}} & \jointchange{\jointloss{-5.4\%}} \\
Direct + post-hoc & 0.9078 & 77,408 & 21.07 & 1.0470 & $-0.0565$ & $-12.88$ & \textbf{0.9623} & 0.1310 \\[-2pt]
 & \jointchange{\jointgain{+0.40\,pp}} & \jointchange{\jointgain{+2.2\%}} & \jointchange{\jointgain{+0.8\%}} & \jointchange{\jointgain{+39.4\%}} & \jointchange{\jointgain{+45.1\%}} & \jointchange{\jointgain{+54.7\%}} & \jointchange{\jointbest{+58.4\%}} & \jointchange{\jointloss{-8.1\%}} \\
Surrogate + post-hoc & \textbf{0.9128} & 76,889 & \textbf{20.90} & \textbf{1.0454} & $\mathbf{-0.0410}$ & $\mathbf{-8.21}$ & 0.9574 & 0.0999 \\[-2pt]
 & \jointchange{\jointbest{+0.89\,pp}} & \jointchange{\jointgain{+2.8\%}} & \jointchange{\jointbest{+1.5\%}} & \jointchange{\jointbest{+41.5\%}} & \jointchange{\jointbest{+60.1\%}} & \jointchange{\jointbest{+71.1\%}} & \jointchange{\jointgain{+53.0\%}} & \jointchange{\jointgain{+17.6\%}} \\
\addlinespace[3pt]
\bottomrule
\end{tabular}
\vspace{1mm}
\begin{minipage}{\textwidth}
\scriptsize
\emph{Notes.} Values in parentheses show changes from the $\rho=0$ fit: percentage points for $R^2_P$ and relative percentages for the other measures. Positive values and green indicate improvement; negative values and red indicate worsening. For PRD and MKI, improvement means moving toward one; for PRB and VEI, toward zero. Bold marks the largest displayed improvement in each test set, including ties. COD and VEI are percentages. Direct uses $\rho=1.677$, Surrogate uses $\rho=0.954$, and the post-hoc rows use $b=1.027316$. Post-hoc alone rescales the $\rho=0$ fit to match Direct's cross-validation slope. Surrogate plus post-hoc was jointly selected over three penalties and four scales using cross-validation. Direct plus post-hoc was not jointly selected. Appendix~\ref{subsec:screening_interpretation} gives the selection details.
\end{minipage}
\end{table}

\subsection{Comparison at Matched First-Order Correction}
\label{subsec:matched_beta_results}

Because equal $\rho$ values do not represent equal correction, we compare the methods at six approximately matched cross-validation slopes within their shared range. Section~\ref{subsec:comparators} describes the construction, and Appendix Table~\ref{tab:matched_beta} reports the results. The slopes are matched in cross-validation and can differ on the held-out and 2025-forward test sets. At all five nonzero targets, Surrogate has the smallest held-out $\Delta_{\mathrm{NL}}$. At the first four, it also has the highest log-price RMSE.

At the strongest shared cross-validation target, Direct reaches the target only at $\rho\approx87$, far beyond its tolerance band. It has the lowest held-out $R^2_P$ and the highest MAE, log-price RMSE, and COD of the three methods. Within Direct's tolerance band, post-hoc rescaling achieves a similar first-order tradeoff, although its nonlinearity gap increases in the reported comparisons.

At the strongest shared target, Surrogate and post-hoc rescaling are nearly tied in held-out $R^2_P$ (0.8963 versus 0.8966), while post-hoc has lower MAE (\$74,236 versus \$76,128). Smearing reverses both orderings (Appendix~\ref{app:smearing}). At the stronger cross-validation target $\beta_{\log}=-0.03$, which Direct does not reach on the tested grid, Surrogate has higher held-out $R^2_P$ than post-hoc rescaling (0.8917 versus 0.8569), but also higher log-price RMSE (0.3443 versus 0.2999; Appendix Table~\ref{tab:matched_beta_ext}).

As a sensitivity check, we repeat the matching while giving each sale equal total weight. Ten of 315 sign comparisons change direction, all involving small differences. The nonlinearity-gap ordering and Direct's deterioration at the strongest shared target remain unchanged (Appendix~\ref{app:matched_beta_detail}).

\subsection{Chronological Stability}
\label{subsec:path_stability_results}

We also test the selection rules on later validation periods. For each of 54 cases (the CCAO application and the eight external counties, two methods, three fold horizons), we select a penalty using earlier folds and evaluate it on the next fold. Median $|\mathrm{PRB}|$ reductions are similar on the selection and next folds: 28\% and 27\% for the constrained rule, and 22\% and 21\% for Nash. All cost limits hold on the next fold in 79\% of cases where the constrained rule makes a selection and in 92\% of Nash cases. The constrained rule makes no selection in 28\% of cases. For comparison, the smallest positive penalty meets the limits in 91\% of cases (Appendix~\ref{subsec:screening_interpretation}).

The penalty paths have broadly similar shapes across cross-validation and the two test sets. Equity measures change first, while stronger penalties eventually introduce accuracy or ratio-shape costs (Figure~\ref{fig:candidate_region_cv} and Appendix Figures~\ref{fig:cv_predictive_metric_paths}--\ref{fig:cv_mechanism_metric_paths}). Within the change-point regions, PRB, VEI, and $\beta_{\log}$ move toward zero and price-scale $R^2_P$ increases at every tested point on both test sets. The values and turning points differ across samples. Along the full Surrogate path, the held-out slope rises slightly above zero, while the cross-validation average and 2025 slope remain negative.

Additional sensitivity checks examine whether the findings depend on the date boundary or on properties appearing in both training and evaluation data. The results are generally consistent with the main findings. Appendix~\ref{app:ccao_path_details} describes the checks and their limitations.

\subsection{Check Across Other Jurisdictions}
\label{subsec:external_results}
To examine whether the methods' results extend beyond Cook County, we evaluate them in eight other counties using ATTOM data (Appendix~\ref{app:external_benchmark}). For each county, we train a LightGBM baseline and use cross-validation from 2018--2024 to set penalties targeting a 25\% reduction in the magnitude of $\beta_{\log}$. We then evaluate the models on 2025 sales, comparing them with the county's unpenalized custom-objective fit. A sales-validation filter excludes 0.16--0.33\% of records with characteristics inconsistent with a single-family sale. The filter substantially raises some counties' 2025 baseline $R^2_P$, including Maricopa's from 0.297 to 0.858. The filter was added after the initial 2025 evaluation exposed a few implausible records; its thresholds are fixed constants or are computed from sales before 2025.

To make penalty scales more comparable across counties, we multiply $\rho$ by the variance of training log prices ($\widetilde\rho$; Appendix~\ref{app:implementation}). This provides a common scale but does not guarantee the same correction in every county. The penalties needed to reach the cross-validation target vary across counties. On 2025 sales, the resulting slope reductions range from about 19\% to 52\%. We also compare the methods with post-hoc rescaling of each county's unpenalized fit, calibrated to the same cross-validation target (Section~\ref{subsec:posthoc_comparator}).

Table~\ref{tab:external_summary} shows how the cross-validation choices perform on 2025 sales. In every county, both methods reduce $|\mathrm{PRB}|$ and $|\beta_{\log}|$, while log-price RMSE increases. Price-scale $R^2_P$ changes by at most 0.022. Paired bootstrap intervals that resample whole calendar months of 2025 sales (200 draws) exclude zero for the PRB and slope changes of both methods in all eight counties (Appendix~\ref{app:external_benchmark}). Direct increases $\Delta_{\mathrm{NL}}$ in every county. Surrogate increases it in three counties, but has a larger median increase in log-price RMSE, reaching 11.1\% in Middlesex. Across counties, both methods achieve larger median $|\mathrm{PRB}|$ reductions than post-hoc rescaling at the same cross-validation target.

\begin{table}[!htbp]
\centering
\footnotesize
\setlength{\tabcolsep}{5pt}
\caption{Median and range of 2025 results across the eight external counties. Changes are measured relative to each county's unpenalized custom-objective fit.}
\label{tab:external_summary}
\begin{tabular}{@{}lcccc@{}}
\toprule
Method & $|\mathrm{PRB}|$ red.\ (\%) & $\Delta R^2_P$ & $\operatorname{RMSE}_{\log P}$ change (\%) & $\Delta(\Delta_{\mathrm{NL}})$ \\
\midrule
Direct & 62.1 [41.2, 72.5] & $-$0.005 [$-$0.018, $+$0.022] & $+$1.2 [$+$0.2, $+$1.8] & $+$0.012 [$+$0.006, $+$0.021] \\
Surrogate & 63.5 [48.5, 91.7] & $+$0.006 [$-$0.004, $+$0.017] & $+$2.9 [$+$1.0, $+$11.1] & $-$0.002 [$-$0.025, $+$0.003] \\
Post-hoc (reference) & 53.1 [38.5, 73.7] & $-$0.002 [$-$0.015, $+$0.019] & $+$0.8 [$+$0.2, $+$1.1] & $+$0.013 [$+$0.007, $+$0.021] \\
\bottomrule
\end{tabular}
\vspace{1mm}
\begin{minipage}{\textwidth}
\scriptsize
\emph{Notes.} The penalties are set using cross-validation to reduce the magnitude of $\beta_{\log}$ by 25\%. For Direct, the reduction is measured from the unpenalized fit. For Surrogate, it is measured from the fit with its smallest tested penalty, $\widetilde\rho=0.001$. Post-hoc rescaling is calibrated to a 25\% cross-validation slope reduction from the unpenalized fit. All changes in the table compare 2025 results with the unpenalized fit. PRB reductions and log-price RMSE changes are relative percentages. $\Delta(\Delta_{\mathrm{NL}})$ is the change in the nonlinearity gap relative to the unpenalized fit; a positive value means the gap increased. Appendix Table~\ref{tab:external_full} reports the results for each county.
\end{minipage}
\end{table}

Appendix~\ref{app:external_benchmark} also reports the full 2025 paths and applies the Cook County tolerance and selection rules to each county.

The external results support the Cook County finding that both methods reduce measured regressivity, although the size of the reduction and its costs vary across counties. In Cook County, the exploratory selections reduce $|\mathrm{PRB}|$ by 22.0--39.4\% on 2025 sales (Table~\ref{tab:cv_selected_comparison}). The external benchmark uses a different selection criterion: a fixed 25\% cross-validation slope target. Its reductions are therefore not directly comparable with the Cook County selections.

\section{Discussion and Conclusions}
\label{sec:discussion}

Ordinary LightGBM predicts sale prices well, but its valuation ratios decline as sale prices rise. This motivates a targeted correction, although it does not show that machine learning itself causes regressivity. At some penalty strengths, both methods reduce the first-order trend with little change in price-scale accuracy relative to the custom-objective fit at $\rho=0$. Stronger penalties can worsen accuracy, valuation level, dispersion, or ratio patterns.

The matched first-order comparison shows what each method adds. At every nonzero matched cross-validation target, Surrogate has the smallest nonlinearity gap, but a higher held-out log-price RMSE than post-hoc rescaling. Direct has the weakest held-out accuracy and dispersion at the strongest target, which lies far beyond its tolerance band (Section~\ref{subsec:matched_beta_results}). Within that band, post-hoc rescaling achieves a similar first-order tradeoff to Direct, consistent with the fixed-space theory (Appendix~\ref{app:theory}). Rescaling can also complement either objective, although it increased $\Delta_{\mathrm{NL}}$ in the reported Cook County comparisons (Section~\ref{subsec:posthoc_comparator}). These comparisons apply at the specified targets, and the fixed-space theory does not predict the exact behavior of retrained trees.

Price-scale comparisons depend on how log predictions are returned to dollars. Under direct exponentiation, the unpenalized and selected fits have median valuation ratios below one on both test sets. The selected fits' small price-scale accuracy gains over $\rho=0$ partly reflect overall valuation level. The $R^2_P$ gains shrink or reverse under level adjustments estimated from cross-validation predictions (Appendix~\ref{app:smearing}). A common multiplicative adjustment changes valuation level but leaves the reported uniformity, vertical-equity, and residual-structure measures unchanged. The results therefore support reduced regressivity with essentially unchanged price-scale accuracy, rather than general predictive superiority for either method.

The selected configurations also have different costs. On both test sets, Direct increases the nonlinearity gap, while Surrogate increases log-price error. The ratio-profile trough deepens in cross-validation for Direct at $\rho=1.677$ and Surrogate at $\rho=0.954$ (Appendix Table~\ref{tab:cv_selection_ci}). The external results also show reductions in the first-order slope, with costs that vary across counties (Section~\ref{subsec:external_results}). Neither study establishes that a penalty value will work the same way in another period or assessor workflow.

\subsection{Limitations}
\label{subsec:external_results_boundary}

Sale price is both the prediction target and part of the valuation ratio. It
is an imperfect proxy for latent market value; unpredictable transaction
variation can produce a negative residual--sale-price association even under
a well-fitted model (Appendix~\ref{app:population_identity}). Moving that
association toward zero is therefore not, by itself, proof of improved equity
relative to unobserved market values. Neither objective directly optimizes
PRB, VEI, valuation level, nonlinear ratio structure, or local equity.

The evidence comes from verified sales in a Python research translation of
CCAO's published residential workflow. The LightGBM configuration is held
fixed at 994 trees to study the objectives, so the paths are not fully retuned
frontiers. Sales need not represent unsold properties, and countywide patterns
can hide differences across neighborhoods and property classes. Recent Cook
County evidence likewise finds that assessment regressivity differs across
property classes \cite{CaiWiley2026}. Appendix~\ref{app:townships} provides an exploratory township-level view: for the configurations examined, the township mean-ratio gradient moves toward zero on 2025 sales and becomes positive on held-out sales. These CCAO paths alone do not establish results for other jurisdictions.

The separate ATTOM benchmark adds a cross-county check, but uses commercially
harmonized sales and a limited history feature set rather than reproducing
each office's sales verification or appraisal workflow. An audit of two
commercial property-data brokers, including ATTOM, against Cook County
administrative records finds sale-price and coverage discrepancies large
enough to shift estimates of assessment regressivity
\cite{SmithHarveyGoldinHo2026Records}. Neither experiment
reports official assessments or estimates of final tax bills.

The reported differences along the penalty paths are mainly descriptive. The chronological cross-validation folds overlap, so their variation does not provide independent standard errors. The sale-bootstrap intervals measure uncertainty for the selected configurations, but do not capture variation from refitting the models or from one period to another. At the Nash selections, fold-based standard errors of the CV changes are 1.2--4.5 times the sale-bootstrap standard errors. The external intervals resample the twelve months of 2025, so they are coarse and do not capture variation across years. Price-scale comparisons also depend on the retransformation method. No penalty-selection rule has been validated prospectively, so we do not recommend a fixed operational setting.

\subsection{Future Work}

In ongoing work with CCAO, we plan to evaluate the Direct objective in the office's residential modeling workflow against an unpenalized fit on the same sales and time periods, before later assessment adjustments. Future work could compare the proposed objectives with spatial valuation models and test whether adding the objectives to those models improves results under the same chronological design. It could also examine predictive accuracy and equity measures across neighborhoods, property classes, price groups, and periods, with uncertainty estimates for each group. Training objectives could also account for differences across groups, rather than targeting only the overall price-related pattern.

In practice, assessors, subject-matter experts, and other decision makers would need to define the equity goals and acceptable tradeoffs for their setting, subject to applicable laws and regulations. The methods reduced measured regressivity in the Cook County application and in the eight-county benchmark, but their effects and costs may differ in other settings. A selection process could then identify a penalty that meets the chosen goals and limits, or make no selection if none qualifies. For example, cross-validation could be used to select the smallest eligible penalty. Stability checks, robust optimization, and evaluation across periods and subgroups could further strengthen the process. The present data do not establish appropriate goals or limits, or validate a selection rule for deployment.

\subsection{Conclusion}

In this CCAO application, both training objectives reduce measured regressivity at moderate penalties while maintaining price-scale accuracy, without changing the sales, predictors, or model class. At the four configurations selected by cross-validation, $|\mathrm{PRB}|$ falls by 22--49\% on the held-out and 2025-forward test sets, relative to the unpenalized fit, with narrow sale-bootstrap intervals. The correction is partial: valuation ratios remain elevated among the lowest-priced sales. The two methods have different costs: Direct increases the nonlinearity gap, whereas Surrogate lowers it but increases log-price error. Because their effects also vary across periods, the results do not identify a single preferred method. Within Direct's tolerance band, post-hoc rescaling of the unpenalized fit gives a similar first-order tradeoff, and it can also complement either objective. The external
benchmark finds the first-order correction in every county, but confirms that
log-price accuracy and nonlinear structure must be checked separately before
considering an operational setting.

\printbibliography

\appendix

\section{Metric Details}
\label{app:metrics}

\subsection{Accuracy and Supporting Ratio-Study Formulas}

For an evaluation sample $S$, let $n_S=|S|$ and $\bar P_S=n_S^{-1}\sum_{i\in S}P_i$. The predictive measures used in the main text are
\begin{align}
\label{eq:predictive_metrics}
\operatorname{MAE}_{P}(S)
&=
\frac{1}{n_S}\sum_{i\in S}\left|\widehat P_i-P_i\right|,
\\
{\operatorname{MAPE}_{P}(S)}
&{=100\times
\frac{1}{n_S}
\sum_{i\in S}
\left|
\frac{\widehat P_i-P_i}{P_i}
\right|
=100\times
\frac{1}{n_S}\sum_{i\in S}|r_i-1|},
\\
R^2_{P}(S)
&=
1-
\frac{\sum_{i\in S}(\widehat P_i-P_i)^2}
{\sum_{i\in S}(P_i-\bar P_S)^2},
\\
\operatorname{RMSE}_{\log P}(S)
&=
\left(
\frac{1}{n_S}\sum_{i\in S}e_i(f)^2
\right)^{1/2}.
\end{align}

The coefficient of variation (COV) is
\begin{equation}
\label{eq:cov}
\operatorname{COV}(S)
=
100\times
\frac{s_{r,S}}{\bar r_S},
\qquad
s_{r,S}
=
\left(
\frac{1}{n_S-1}
\sum_{i\in S}(r_i-\bar r_S)^2
\right)^{1/2}.
\end{equation}

For the Modified Kakwani Index (MKI), let $\pi(1),\allowbreak\ldots,\pi(n_S)$ index the observations in $S$ in nondecreasing order of sale price, with tied sale prices ordered by decreasing predicted value, and let
\[
q_j=\frac{j-\tfrac12}{n_S},
\qquad
\overline{\widehat P}_S
=
\frac{1}{n_S}\sum_{i\in S}\widehat P_i.
\]
The concentration index of predicted values ordered by sale price and the Gini coefficient of sale prices are
\begin{equation}
\label{eq:mki_components}
C_{\widehat P\mid P}(S)
=
\frac{2}{n_S\overline{\widehat P}_S}
\sum_{j=1}^{n_S}
\widehat P_{\pi(j)}q_j
-1,
\qquad
G_P(S)
=
\frac{2}{n_S\bar P_S}
\sum_{j=1}^{n_S}
P_{\pi(j)}q_j
-1.
\end{equation}
Provided there is sufficient sale-price variation, the index is
\begin{equation}
\label{eq:mki}
\operatorname{MKI}(S)
=
\frac{C_{\widehat P\mid P}(S)}{G_P(S)}.
\end{equation}
Many sales share a sale price, so the tie rule matters slightly. With tie-invariant average ranks instead, MKI is between $0.0005$ and $0.0010$ higher for the shared $\rho=0$ fit, ordinary LightGBM, and every configuration in the CV change-point regions, on every evaluation sample; the MKI improvements in Table~\ref{tab:cv_selected_comparison} change by at most $0.45$ percentage points, without changing their ordering. All reported values use the executed rule.

\subsection{Value-Proxy Convention}

The 2013 IAAO Standard defines the equal-weight market-value proxy as
$0.50(\mathrm{AV}/\mathrm{Median})+0.50\,\mathrm{SP}$.
The May 2026 exposure draft describes an equal-weight proxy in its VEI
appendix but displays the normalized valuation term without an explicit
$0.50$ multiplier. All PRB and VEI results here use the single equal-weight
convention in Eq.~\eqref{eq:market_value_proxy}, consistent with the 2013
Standard and the draft's stated interpretation; the executed metric code uses it for both.

\subsection{Mechanism and Residual-Structure Diagnostics}

This subsection defines the estimators of $\Delta_{\mathrm{NL}}$ and distance
correlation; $\beta_{\log}$ is defined in Eq.~\eqref{eq:log_ratio_slope}.

For the nonlinear diagnostic, let
\[
Z=\frac{y-\E[y]}{\sqrt{\Var(y)}},
\qquad
m(z)=\E[e\mid Z=z],
\]
and let $\Pi_{\mathrm{aff}}m$ denote the $L^2(P_Z)$ projection of $m$ onto the
affine span of $\{1,Z\}$. Pearson's correlation ratio, also called the
nonparametric coefficient of determination, is
\[
\eta^2(e\mid Z)
=
\frac{\Var\!\left(m(Z)\right)}{\Var(e)}
=
\frac{\Var\!\left(\E[e\mid Z]\right)}{\Var(e)}
\]
\cite{DoksumSamarov1995}. Because $Z$ is standardized, the affine component explains
the fraction $\Corr(e,Z)^2$ of the variance of $e$. Orthogonal projection therefore
gives
\begin{equation}
\label{eq:nonlinearity_gap_projection}
\Delta_{\mathrm{NL}}
=
\eta^2(e\mid Z)-\Corr(e,Z)^2
=
\frac{
\E\!\left[
\left\{
m(Z)-(\Pi_{\mathrm{aff}}m)(Z)
\right\}^2
\right]
}{
\Var(e)
}
\geq 0.
\end{equation}
Thus $\Delta_{\mathrm{NL}}$ isolates the conditional-mean explanatory power that is
not captured by the best affine residual--price relationship. The name \emph{correlation-ratio nonlinearity gap} and the symbol
$\Delta_{\mathrm{NL}}$ are ours; \textcite{DoksumSamarov1995} provide the
correlation-ratio foundation.

For the empirical CCAO analysis, the finite-sample implementation is fixed before
the penalized paths are compared. Each held-out or forward evaluation sample $S$ is
split into five fixed folds, the same for every model. For each fold, we
fit on the complementary observations (i) an affine regression of $e$ on $Z$ and
(ii) a cubic $B$-spline regression of $e$ on $Z$ with eight quantile knots, and we
score both fits on the omitted fold. Let
$\operatorname{MSE}_{\mathrm{aff}}(S)$ and
$\operatorname{MSE}_{\mathrm{spl}}(S)$ denote the resulting pooled out-of-fold mean
squared errors. The reported sample statistic is
\begin{equation}
\label{eq:nonlinearity_gap_estimator}
\widehat{\Delta}_{\mathrm{NL}}(S)
=
\max\left\{
0,\,
\frac{
\operatorname{MSE}_{\mathrm{aff}}(S)
-
\operatorname{MSE}_{\mathrm{spl}}(S)
}{
\widehat{\Var}_S(e)
}
\right\}.
\end{equation}
The nonnegative truncation respects the population inequality in
Eq.~\eqref{eq:nonlinearity_gap_projection}; finite-sample cross-fitting can otherwise
produce a small negative difference when the flexible fit does not improve
out-of-fold squared error. For readability, the empirical tables use
$\Delta_{\mathrm{NL}}$ as shorthand for
$\widehat{\Delta}_{\mathrm{NL}}(S)$.
The estimator is reported on the primary held-out and 2025-forward paths. The same cross-fitted estimator is applied, fold by fold, to the stored chronological validation predictions; those CV values were not used to define the $\rho$ grid, but they enter the change-point screen and the CV selections of Appendix~\ref{subsec:screening_interpretation}.

Finally, because $e=\log r$ and $y=\log P$,
\begin{equation}
\label{eq:dcor_diagnostic}
\operatorname{dCor}(e,y)
=
\operatorname{dCor}(\log r,\log P).
\end{equation}
Population distance correlation is zero if and only if the variables are independent, under
the usual moment conditions \cite{SzekelyRizzoBakirov2007}. The paper uses its
sample analogue only as an omnibus diagnostic of residual--price dependence; unlike
$\beta_{\log}$ it is unsigned, and unlike $\Delta_{\mathrm{NL}}$ it is not restricted
to conditional-mean structure. It has no assessor-standard reference range.

All reported values use the biased, double-centered \texttt{distance\_correlation}
estimator of the \texttt{dcor} Python package, with its default unit exponent, on the
full evaluation sample.

\section{Squared-Error Prediction and Residual--Value Dependence}
\label{app:population_identity}

Let $Y=\log P$, let $X$ be the observed feature vector, and let
\[
f^\star(X)=\E[Y\mid X]
\]
be the squared-error Bayes predictor. Define $e^\star=f^\star(X)-Y$. If $\E[Y^2]<\infty$, then
\begin{proposition}
\label{prop:bayes_covariance}
\[
\Cov(e^\star,Y)
=
-\E[\Var(Y\mid X)]
\le 0.
\]
\end{proposition}
\begin{proof}
Using the law of total variance,
\[
\Cov(e^\star,Y)
=
\Cov(\E[Y\mid X]-Y,Y)
=
\Var(\E[Y\mid X])-\Var(Y)
=
-\E[\Var(Y\mid X)].
\]
\end{proof}

This identity explains a source of negative log-residual dependence when important price variation remains unresolved after conditioning on $X$. It does not imply that every finite-sample model is regressive under every assessor metric. It also does not imply that adding features cannot improve both accuracy and equity.

\section{\texorpdfstring{Fixed-Space Regularization Paths: Direct and Surrogate}{Fixed-Space Regularization Paths: Direct and Surrogate}}
\label{app:theory}

This appendix isolates the data-fit geometry of the two penalties when the prediction space is held fixed, omitting model-specific complexity penalties. The Direct result provides an exact rank-one benchmark for the targeted covariance correction. The Surrogate result provides the corresponding log-price-distance-weighted projection benchmark and makes precise why its path need not remain one-dimensional. These finite-sample results explain the mechanisms; they do not describe LightGBM's boosting paths exactly.

\subsection{Direct Penalty: Rank-One Covariance Path}

Let $\Vcal\subseteq\R^n$ be a closed linear prediction space, define
\[
\langle u,v\rangle_n=\frac{1}{n}u^\top v,
\qquad
\|u\|_n^2=\langle u,u\rangle_n,
\qquad
c=y-\bar y\1,
\]
and consider
\begin{equation}
\label{eq:path_problem}
f_\rho\in\argminop_{f\in\Vcal}
\left\{
\|f-y\|_n^2
+\frac{\rho}{2}\langle f-y,c\rangle_n^2
\right\}.
\end{equation}
Let $P_{\Vcal}$ be the orthogonal projection under $\langle\cdot,\cdot\rangle_n$ and set
\[
f_0=P_{\Vcal}y,\qquad
C_0=\langle f_0-y,c\rangle_n,\qquad
g=P_{\Vcal}c,\qquad
A=\|g\|_n^2.
\]

\begin{proposition}[Covariance shrinkage path]
\label{prop:path}
If $A>0$, then
\[
f_\rho=f_0-\frac{\rho}{2}C_\rho g,
\qquad
C_\rho=\frac{C_0}{1+\rho A/2}.
\]
Consequently, the slope of residuals on $y$ is multiplied by
\[
q(\rho)=\frac{1}{1+\rho A/2}.
\]
\end{proposition}

\begin{proof}
For every $h\in\Vcal$, first-order optimality gives
\[
2\langle f_\rho-y,h\rangle_n
+\rho C_\rho\langle c,h\rangle_n=0.
\]
Since $f_0=P_{\Vcal}y$, $\langle f_0-y,h\rangle_n=0$. Writing
$d_\rho=f_\rho-f_0\in\Vcal$ and $g=P_{\Vcal}c$ gives
\[
2d_\rho+\rho C_\rho g=0.
\]
Taking the inner product with $c$ yields
\[
C_\rho=C_0-\frac{\rho}{2}AC_\rho,
\]
which proves the result.
\end{proof}

{\begin{corollary}[Centered-spread form of the fixed-space Direct path]
\label{cor:path_scaling}
If $A>0$ and $\1\in\Vcal$, then $g=f_0-\bar y\1$ and
\[
f_\rho
=
\bar y\1
+
b_\rho\bigl(f_0-\bar y\1\bigr),
\qquad
b_\rho
=
1-\frac{\rho}{2}C_\rho.
\]
Moreover, $C_0=-\|f_0-y\|_n^2$. Hence, whenever the unpenalized fit is non-perfect, $C_0<0$, $b_\rho>1$ for $\rho>0$, and $C_\rho$ approaches zero from below without changing sign at any finite $\rho$.
\end{corollary}

\begin{proof}
Because $\1\in\Vcal$,
\[
g=P_{\Vcal}(y-\bar y\1)=f_0-\bar y\1.
\]
Substituting this identity into Proposition~\ref{prop:path} gives the centered-spread representation. Also,
\[
c=(f_0-\bar y\1)+(y-f_0),
\]
and the projection residual $y-f_0$ is orthogonal to $\Vcal$, so
\[
C_0
=
\langle f_0-y,c\rangle_n
=
-\|f_0-y\|_n^2.
\]
The remaining statements follow from $C_\rho=C_0/(1+\rho A/2)$.
\end{proof}

\begin{remark}[Mechanical component of the in-sample residual--outcome covariance]
\label{rem:mechanical_covariance}
Corollary~\ref{cor:path_scaling} gives $C_0=-\|f_0-y\|_n^2\leq0$: part of the negative training covariance is mechanical, induced by fitting the same $y$ that defines the log ratio. The identity does not hold on a new evaluation sample and does not make out-of-sample regressivity spurious; it means that $C(f)$ is a controllable first-order mechanism, not an estimator of inequity relative to latent market value.
\end{remark}

This equivalence is specific to the fixed-prediction-space benchmark, in which Direct regularization is no richer than a one-parameter centered spread correction. Sections~\ref{subsec:posthoc_comparator} and~\ref{subsec:matched_beta_results} test this prediction empirically. It does not imply that the Direct-LightGBM path is one-dimensional: retraining can change tree splits and leaf structure, and the implementation uses only the diagonal of the covariance Hessian.

}

On a fitting block $T$, the centered rescaling that sets the first-order association to zero is
\[
b^{\star}_{T}=\widehat{\Var}_{T}(y)/\widehat{\Cov}_{T}(f_0,y),
\]
which reduces to $1/R^2$ only when
$f_0$ is a projection of $y$. Because the boosted learner is much less regressive in sample than
out of sample, $b^{\star}_{T}$ is an in-sample diagnostic; the empirical comparator is calibrated
on development validation folds (Appendix~\ref{app:centered_spread_detail}).

\begin{proposition}[Accuracy cost]
\label{prop:path_cost}
Under Proposition~\ref{prop:path},
\[
\|f_\rho-y\|_n^2-\|f_0-y\|_n^2
=
\frac{C_0^2}{A}\bigl(1-q(\rho)\bigr)^2.
\]
\end{proposition}

Within a fixed prediction space, the targeted covariance falls to first order near $\rho=0$, while the in-sample squared-error cost is second order. For boosted trees this is a local benchmark only. Retraining can change splits, leaves, and stopping iteration.

\subsection{Surrogate Penalty: Weighted-Projection Path}

{Let
\[
D=\diag(c_1^2,\ldots,c_n^2),
\qquad
W_\rho=I+\rho D,
\qquad
e_0=f_0-y,
\]
where $f_0=P_{\Vcal}y$ is the ordinary least-squares projection defined above. The exact fixed-space Surrogate problem corresponding to Eq.~\eqref{eq:surrogate_weighted_loss} is
\begin{equation}
\label{eq:surrogate_fixed_space_problem}
f_\rho^{\mathrm{surr}}
=
\argminop_{f\in\Vcal}
\left\{
\|f-y\|_n^2
+
\rho\langle f-y,D(f-y)\rangle_n
\right\}.
\end{equation}
Because $W_\rho$ is positive definite for every $\rho\geq0$, this problem has a unique solution. Define the weighted inner product
\[
\langle u,v\rangle_{n,W_\rho}
=
\langle u,W_\rho v\rangle_n,
\]
let $P_{\Vcal}^{W_\rho}$ denote projection onto $\Vcal$ under this inner product, and define the self-adjoint positive-semidefinite operator
\[
T=P_{\Vcal}D\big|_{\Vcal}
\]
together with the baseline tail-weighted residual direction
\[
g_{\mathrm{surr}}
=
P_{\Vcal}De_0.
\]}

{\begin{proposition}[Surrogate weighted-projection path]
\label{prop:surrogate_fixed_space}
For every $\rho\geq0$,
\begin{equation}
\label{eq:surrogate_weighted_projection}
f_\rho^{\mathrm{surr}}
=
P_{\Vcal}^{W_\rho}y,
\end{equation}
and its departure from the unpenalized fit is
\begin{equation}
\label{eq:surrogate_correction_operator}
f_\rho^{\mathrm{surr}}-f_0
=
-\rho
\left(I_{\Vcal}+\rho T\right)^{-1}
g_{\mathrm{surr}}.
\end{equation}
\end{proposition}

\begin{proof}
For every $h\in\Vcal$, first-order optimality for~\eqref{eq:surrogate_fixed_space_problem} gives
\[
\langle f_\rho^{\mathrm{surr}}-y,h\rangle_n
+
\rho\langle D(f_\rho^{\mathrm{surr}}-y),h\rangle_n
=
0,
\]
or equivalently
\[
\langle f_\rho^{\mathrm{surr}}-y,h\rangle_{n,W_\rho}=0.
\]
This is exactly the weighted-projection characterization in~\eqref{eq:surrogate_weighted_projection}. Now write
\[
\delta_\rho=f_\rho^{\mathrm{surr}}-f_0\in\Vcal.
\]
Since $f_0=P_{\Vcal}y$, the ordinary projection residual $e_0=f_0-y$ is orthogonal to $\Vcal$. Substituting
$f_\rho^{\mathrm{surr}}-y=e_0+\delta_\rho$ into the first-order condition and projecting the $D$ terms onto $\Vcal$ yields
\[
\delta_\rho
+
\rho T\delta_\rho
+
\rho g_{\mathrm{surr}}
=
0.
\]
Because $T$ is positive semidefinite, $I_{\Vcal}+\rho T$ is invertible for $\rho\geq0$, which proves~\eqref{eq:surrogate_correction_operator}.
\end{proof}
}

{\begin{corollary}[Local Surrogate direction and baseline-loss cost]
\label{cor:surrogate_local}
As $\rho\downarrow0$,
\begin{equation}
\label{eq:surrogate_local_direction}
f_\rho^{\mathrm{surr}}
=
f_0
-
\rho g_{\mathrm{surr}}
+
O(\rho^2)
=
f_0
-
\rho P_{\Vcal}
\left[
c^{\odot2}\odot e_0
\right]
+
O(\rho^2),
\end{equation}
where $\odot$ denotes elementwise multiplication and $c^{\odot2}=(c_1^2,\ldots,c_n^2)^\top$. Moreover,
\begin{equation}
\label{eq:surrogate_baseline_loss_cost}
\|f_\rho^{\mathrm{surr}}-y\|_n^2
-
\|f_0-y\|_n^2
=
\rho^2\|g_{\mathrm{surr}}\|_n^2
+
O(\rho^3).
\end{equation}
\end{corollary}

\begin{proof}
The expansion
\[
(I_{\Vcal}+\rho T)^{-1}
=
I_{\Vcal}-\rho T+O(\rho^2)
\]
in~\eqref{eq:surrogate_correction_operator} gives~\eqref{eq:surrogate_local_direction}. For the second claim, $\delta_\rho=f_\rho^{\mathrm{surr}}-f_0$ belongs to $\Vcal$ and $e_0=f_0-y$ is orthogonal to $\Vcal$, so
\[
\|e_0+\delta_\rho\|_n^2-\|e_0\|_n^2
=
\|\delta_\rho\|_n^2.
\]
Using $\delta_\rho=-\rho g_{\mathrm{surr}}+O(\rho^2)$ gives~\eqref{eq:surrogate_baseline_loss_cost}.
\end{proof}
}

\subsection{Direct--Surrogate Comparison in a Fixed Prediction Space}

{The preceding results expose a structural difference that is not visible from the upper-bound relation alone. For the Direct penalty, Proposition~\ref{prop:path} shows that the entire fixed-space path moves along the single direction $P_{\Vcal}c$; when $\1\in\Vcal$, Corollary~\ref{cor:path_scaling} identifies this direction with the centered fitted-value spread $f_0-\bar y\1$. The Direct correction is therefore rank one in this benchmark. For the Surrogate, the initial direction is instead
\[
-g_{\mathrm{surr}}
=
-P_{\Vcal}
\left[c^{\odot2}\odot e_0\right],
\]
the learnable component of the baseline residual pattern after weighting observations by squared log-price distance from the sample center. The complete Surrogate path generally changes direction as $\rho$ varies.}

{To see this explicitly, let $\{u_j\}$ be an orthonormal eigenbasis of $T$ on $\Vcal$, with
\[
Tu_j=\lambda_j u_j,
\qquad
\lambda_j\geq0,
\]
and write
\[
g_{\mathrm{surr}}=\sum_j a_j u_j.
\]
Then Proposition~\ref{prop:surrogate_fixed_space} gives
\begin{equation}
\label{eq:surrogate_spectral_path}
f_\rho^{\mathrm{surr}}-f_0
=
-\sum_j
\frac{\rho}{1+\rho\lambda_j}
a_j u_j.
\end{equation}
Thus, unless the nonzero components of $g_{\mathrm{surr}}$ lie in a single eigenspace of $T$, the relative contribution of the learnable correction modes changes with $\rho$. This contrasts with the one-direction Direct path. It also makes clear that the Surrogate does not inherit the Direct formula
$C_\rho=C_0/(1+\rho A/2)$: the Surrogate minimizes a log-price-distance-weighted prediction loss rather than the signed covariance itself, so monotone covariance shrinkage is not guaranteed by the fixed-space theory. Likewise, the spectral representation does not imply that stronger Surrogate regularization must generate any particular nonlinear ratio shape; the empirical shape and $\Delta_{\mathrm{NL}}$ results remain separate diagnostics.}

\section{Linear and Boosting Implementations}
\label{app:implementation}

\subsection{Linear Benchmark}

The linear benchmark in Table~\ref{tab:ccao_baseline_results} is an unpenalized
log-price ordinary least-squares fit. It uses the same sales
and temporal splits as LightGBM, with preprocessing fitted afresh on each
training block. The linear pipeline drops two nonnumeric location identifiers
used by LightGBM (\texttt{loc\_census\_tract\_geoid} and
\texttt{loc\_tax\_municipality\_name}) and adds target and one-hot encodings,
selected Box--Cox transforms and squared terms, imputation, winsorization,
and scaling. These representations are not nested, so their performance gap
cannot be attributed to model class alone. The research fit does not represent
the office's historical township-level linear models.

\subsection{LightGBM Custom-Objective Initialization and Scaling}

For the canonical custom objectives, let $b=\bar y$ denote the mean log sale price
in the current training sample. We fit the custom-objective model to centered labels
$\widetilde y_i=y_i-b$ with zero initial score and add $b$ back to predictions.
Equivalently, the fitted model predicts on the original log-price scale as
$f(x)=b+\widetilde f(x)$.
This construction is designed to match the native squared-error LightGBM starting prediction while leaving $e_i=f(x_i)-y_i$ and $c_i=y_i-\bar y$ unchanged.

The corresponding native/custom zero-penalty audit is reported in the next subsection.

The mathematical objectives in the main text are written as sample averages. In
the LightGBM custom-objective implementation we multiply the complete objective by
$n/2$, a common positive factor that does not change its minimizer. This yields a
native squared-error baseline gradient $g_i=e_i$ and Hessian $h_i=1$ at $\rho=0$.
For the direct objective, the supplied gradient and diagonal Hessian are
\[
g_i^{\mathrm{cov}}
=
e_i+\frac{\rho}{2}z c_i,
\qquad
h_i^{\mathrm{cov,diag}}
=
1+\frac{\rho}{2n}c_i^2,
\qquad
z=\frac{1}{n}\sum_{i=1}^n e_i c_i.
\]
For the surrogate, the supplied derivatives are
\[
g_i^{\mathrm{surr}}=e_i(1+\rho c_i^2),
\qquad
h_i^{\mathrm{surr}}=1+\rho c_i^2.
\]
Thus the direct implementation uses the exact covariance gradient and only the
diagonal of its dense covariance curvature, whereas the surrogate derivatives are
exactly observation-additive. The $\rho=0$ custom-objective path is retained as an
implementation control.

Implementation diagnostics do not show that the omitted cross-observation Direct
curvature is negligible at the evaluated penalties, nor that omitting it materially
changes the fitted regularization path; the main text therefore makes neither claim.

\paragraph{Penalty-scale normalization.}
For the squared-error data-fit terms, rescaling both $y$ and $f$ by a factor
$a$ multiplies squared error by $a^2$, while both $C(f)^2$ and
$\Psi^{\mathrm{surr}}(f)$ are multiplied by $a^4$. A corresponding
variance-normalized penalty coordinate is
\begin{equation}
\label{eq:normalized_rho}
\widetilde\rho=\rho\,\widehat{\Var}_{\mathrm{train}}(y).
\end{equation}
This compares the scales of the stated loss terms when the training-block
variance is computed without leakage. It does not equalize Direct and
Surrogate, account for all other LightGBM settings, or make a fitted penalty
strength portable to another sample or jurisdiction. The empirical CCAO
paths use the raw $\rho$ values reported in the main text.

\subsection{Zero-Penalty Implementation Audit}
\label{subsec:zero_control_results}

Before attributing within-family changes to positive regularization, we compare ordinary LightGBM with the Direct and Surrogate implementations at $\rho=0$. The two custom-objective controls coincide with each other in the reported metrics, but the stored fits are not numerically identical to ordinary LightGBM. Because the supplied zero-penalty gradient and Hessian match squared-error derivatives algebraically, this audit identifies a native-versus-custom execution-path difference rather than a positive-penalty objective effect.

{\begin{table}[!ht]
\centering
\caption{Implementation-control comparison at $\rho=0$. Prediction differences are log-price deviations from ordinary LightGBM, aligned on sale identifiers and computed independently on the held-out and 2025 samples.}
\label{tab:rho_zero_control}
\begin{tabular}{llrrrrr}
\toprule
Sample & Model & $R^2_P$ & $\operatorname{RMSE}_{\log P}$ & $\beta_{\log}$ & Mean $|\Delta\widehat y|$ & Max $|\Delta\widehat y|$ \\
\midrule
Held-out & Ordinary LightGBM & 0.894 & 0.289 & -0.150 & -- & -- \\
Held-out & Direct, $\rho=0$ & 0.893 & 0.289 & -0.147 & 3.24e-02 & 3.09e-01 \\
Held-out & Surrogate, $\rho=0$ & 0.893 & 0.289 & -0.147 & 3.24e-02 & 3.09e-01 \\
\addlinespace
2025 forward & Ordinary LightGBM & 0.904 & 0.278 & -0.164 & -- & -- \\
2025 forward & Direct, $\rho=0$ & 0.904 & 0.279 & -0.161 & 3.06e-02 & 2.83e-01 \\
2025 forward & Surrogate, $\rho=0$ & 0.904 & 0.279 & -0.161 & 3.06e-02 & 2.83e-01 \\
\bottomrule
\end{tabular}
\end{table}
}

\paragraph{Source of the difference.} A parity ladder at three learner capacities shows that label
centering does not cause the difference: ordinary LightGBM and a centered-label native L2 fit agree
to about $10^{-8}$ in log prediction at reduced capacity, whereas the centered-label native fit and
the custom-objective $\rho=0$ fit differ at every capacity (held-out maximum absolute log difference
0.297 at the canonical capacity). At canonical capacity, single-precision representation of the
centered labels also contributes to the ordinary-versus-centered difference. LightGBM's determinism controls leave this unchanged. Single-tree
traces isolate feature subsampling: with the tuned \texttt{colsample\_bytree} of 0.541 the first
trees already differ, and with \texttt{colsample\_bytree} set to $1$ the two learners are bitwise
identical. The difference is therefore a deterministic built-in-versus-custom execution path under
feature subsampling, not an error in the objective. Changing the tuned setting would define a
different learner, so Table~\ref{tab:rho_zero_control} is reported as measured.

\paragraph{Scaling audit.} In all nine fitting blocks, the executed derivatives match the $n/2$
scaling above ($g_i=e_i$ and $h_i=1$ at $\rho=0$; Direct diagonal Hessian $1+(\rho/2n)c_i^2$;
Surrogate Hessian $1+\rho c_i^2$), and the reported $\rho$ values are those the objective received.

\section{\texorpdfstring{
Primary CCAO Regularization-Path Details}{Primary CCAO Regularization-Path Details}}
\label{app:ccao_path_details}

This appendix gives the full path diagnostics behind Section~\ref{sec:results}; the five display points there are snapshots of these paths.

\subsection{Complete Regularization-Path Results}

The complete regularization path is held in the stored combined path table, which carries every
family--$\rho$ configuration together with the seven fold values, equal-weight CV mean and
standard deviation, held-out results, and 2025-forward results for the
Section~\ref{subsec:metrics} evaluation metrics. It is not redistributed.
Each family was first evaluated on 50 geometrically spaced positive values of $\rho$ from $0.1$ to $100$; a lower-tail extension then added 32 positive values between approximately $0.00110$ and $0.08685$ on the same logarithmic spacing, giving 82 positive values per family.
Table~\ref{tab:path_anchor_frozen} reports prediction, COD, PRD, PRB, VEI, and MKI at five display penalties $\{0.01,0.1,1,10,100\}$. In that order, the exact tested values are $\{0.0104811,0.1,0.954095,10.4811,100\}$. The table shows the linear benchmark, ordinary LightGBM, and the shared custom-objective zero-penalty origin separately. PRD and PRB have adopted guidance; the cited MKI and VEI bands are proposed. The zero-penalty implementation audit remains in Appendix~\ref{app:implementation}. No path row is omitted from the machine-readable results because of its observed performance.

\paragraph{Surrogate weight concentration.}
The Surrogate weights $w_i=1+\rho c_i^2$ can concentrate on sales far from
the center of the training log-price distribution. At $\rho=100$, their
effective sample size
$\bigl(\sum_i w_i\bigr)^2/\sum_i w_i^2$ on the development block is 86,609
of 344,607 rows, or 25.1\% of $n$; across the nine fitting blocks it ranges
from 25.0\% to 28.6\% of $n$. On the development block the median weight is
19.1, and the most extreme one percent of rows carry 10.7\% of total
weight. This audit describes the weights, not the causal effect of their
concentration on the fitted paths.

\begin{table}[!htbp]
\centering
\scriptsize
\setlength{\tabcolsep}{2.4pt}
\renewcommand{\arraystretch}{1.06}
\newcommand{\baselinecompmetric}[1]{\hspace*{0.6em}#1}
\caption{Valuation level, ratio dispersion, and residual-structure diagnostics for the unpenalized linear research benchmark and ordinary LightGBM.}
\label{tab:ccao_baseline_complementary}
\begin{tabularx}{\textwidth}{@{} >{\raggedright\arraybackslash}p{2.80cm} >{\centering\arraybackslash}X >{\centering\arraybackslash}X >{\centering\arraybackslash}X >{\centering\arraybackslash}X @{}}
\toprule
\textbf{Measure} & \multicolumn{2}{c}{\textbf{Held-out}} & \multicolumn{2}{c}{\textbf{2025-forward}} \\
\cmidrule(lr){2-3}\cmidrule(l){4-5}
& \textbf{Linear} & \textbf{LightGBM} & \textbf{Linear} & \textbf{LightGBM} \\
\midrule
\multicolumn{5}{@{}l}{\textbf{Valuation level}} \\
\cmidrule(r){1-1}
\baselinecompmetric{Median ratio $m_S$} & 0.969 & 0.929 & 1.020 & 0.950 \\
\baselinecompmetric{Mean ratio $\bar r_S$} & 1.020 & 0.989 & 1.081 & 1.015 \\
\baselinecompmetric{Weighted mean $\bar r_{W,S}$} & 0.979 & 0.924 & 1.027 & 0.941 \\
\addlinespace[5pt]
\multicolumn{5}{@{}l}{\textbf{Horizontal uniformity}} \\
\cmidrule(r){1-1}
\baselinecompmetric{COD} & 24.7\% & 21.6\% & 24.3\% & 21.3\% \\
\baselinecompmetric{COV} & 45.2\% & 39.7\% & 42.1\% & 37.2\% \\
\addlinespace[5pt]
\multicolumn{5}{@{}l}{\textbf{Mechanism and residual structure}} \\
\cmidrule(r){1-1}
\baselinecompmetric{$\beta_{\log}$} & -0.092 & -0.150 & -0.109 & -0.164 \\
\baselinecompmetric{$\Delta_{\mathrm{NL}}$} & 0.131 & 0.119 & 0.122 & 0.121 \\
\baselinecompmetric{$\operatorname{dCor}(e,y)$} & 0.250 & 0.387 & 0.269 & 0.422 \\
\bottomrule
\end{tabularx}
\vspace{1mm}
\begin{minipage}{\textwidth}
\scriptsize
\emph{Notes.}
Both models are research fits on the same sales; the linear fit is not the office's earlier production workflow. The definitions are in Section~\ref{subsec:metrics} and Appendix~\ref{app:metrics}. $\Delta_{\mathrm{NL}}$ and distance correlation describe residual structure rather than assessor-standard equity measures; a smaller global slope $|\beta_{\log}|$ need not imply a smaller nonlinearity gap. COD and COV are percentages. The pooled residential COD is descriptive and should not be treated as a property-class compliance test.
\end{minipage}
\end{table}

\subsection{Prediction, Valuation-Level, and Horizontal-Uniformity Paths}

The figures below show the paths over the tested grid; they mark no penalty interval.

\begin{figure}[!htbp]
\centering
\safeincludegraphics[width=0.66\textwidth,height=0.34\textheight,keepaspectratio]{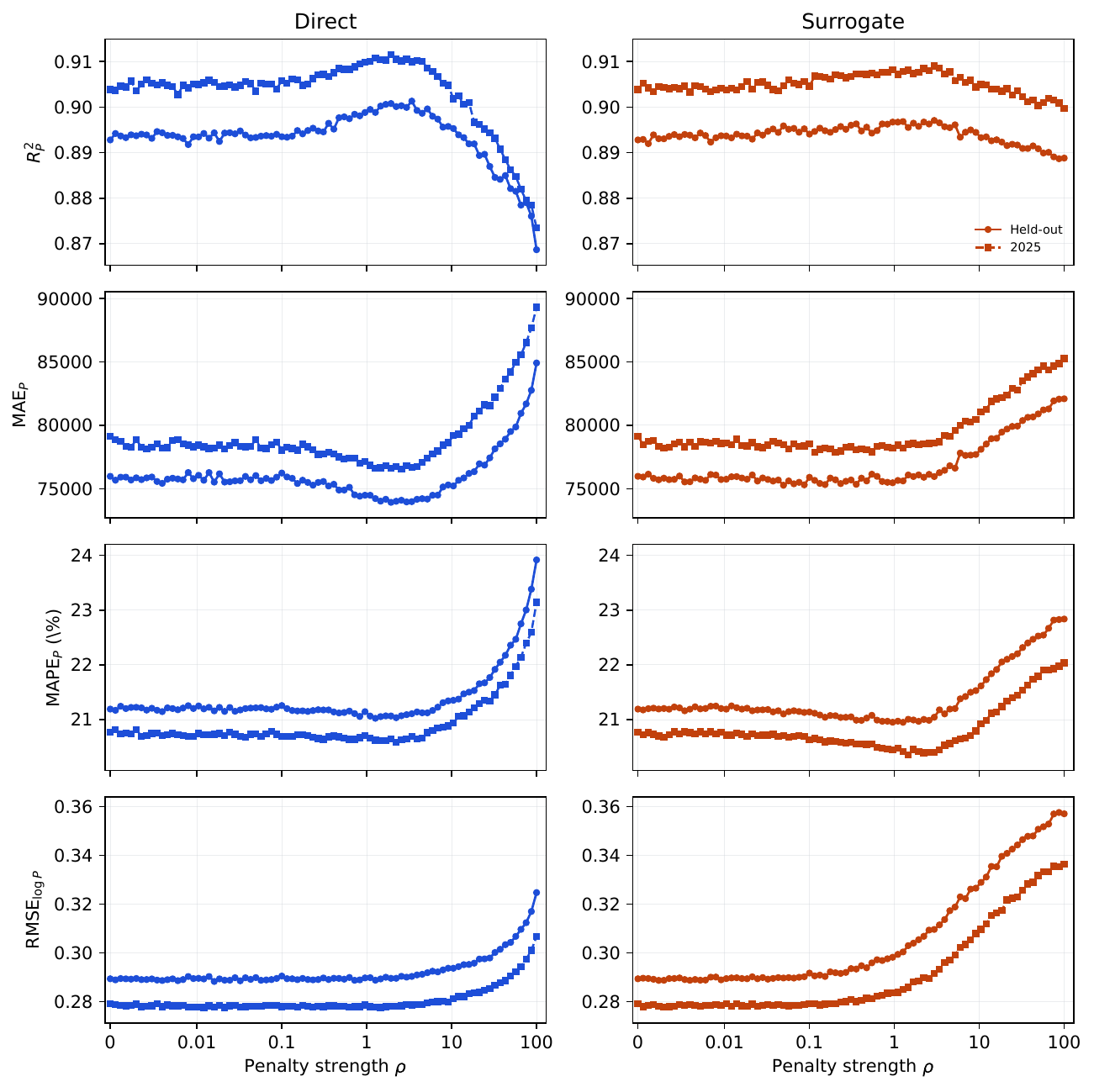}\\[1mm]
\safeincludegraphics[width=0.66\textwidth,height=0.34\textheight,keepaspectratio]{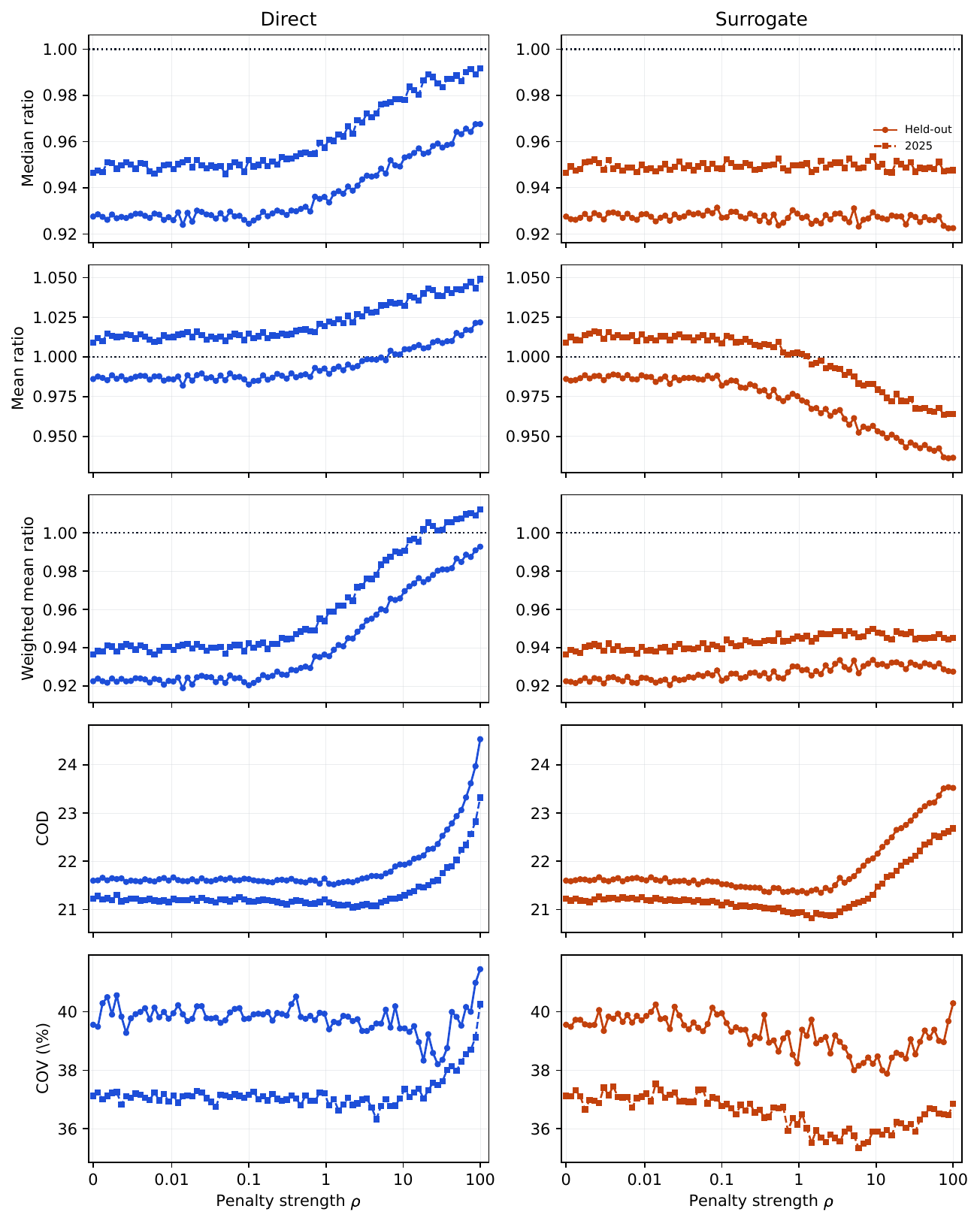}
\caption{Predictive-metric and valuation-level/uniformity paths along the Direct and Surrogate grids, with held-out and 2025 evaluations overlaid. Valuation-level panels include the ratio $=1$ reference. The leftmost tick, 0, marks the $\rho=0$ fit and is not on the logarithmic scale.}
\label{fig:other_metric_paths_placeholder}
\end{figure}

\subsection{Exact Rolling-Origin Fold Structure}

Table~\ref{tab:fold_structure} records the exact cumulative windows and sample sizes underlying the seven-fold chronological design described in Section~\ref{subsec:path_design}. In every fold, the chronologically newest 10\% of the cumulative window is used for validation.

\paragraph{Cross-validation summaries.} Besides D1, the equal-weight mean of each metric over the seven folds, we use two sensitivity summaries. D2 computes each metric on the pooled validation predictions; because some sales occur in two validation folds, D2 is retained only as a reproducibility sensitivity. D3 is a pooled one-sale-one-vote sensitivity: each appearance of sale $i$ receives weight $1/m_i$, where $m_i$ is the number of validation folds containing that sale, so each distinct sale has total weight one.

The last two validation blocks overlap: their origins are about four months apart, while each block contains the newest 10\% of its cumulative window. Across all folds, 151,153 validation appearances represent 130,165 distinct sales; 20,988 sales appear twice, and none appears more often. Every prediction is out of training sample for the model that produced it. Nevertheless, a repeated sale influences two fold metrics in D1 and enters the pooled D2 calculation twice. D1 remains the primary development coordinate; D3 checks sensitivity to giving each distinct sale one total vote. The folds are not independent replications, so their standard deviations describe variation across chronological windows, not standard errors or significance tests.

{\begin{table}[!ht]
\centering
\caption{Rolling-origin validation structure for the final CCAO development sample. In every fold, the newest 10\% of observations in the cumulative window form the validation block. The fold-6 and fold-7 validation blocks overlap, so the folds are not independent replications.}
\label{tab:fold_structure}
\begin{tabular}{clrr}
\toprule
Fold & Cumulative window & $n_{\mathrm{train}}$ & $n_{\mathrm{val}}$ \\
\midrule
1 & 2016-01-01--2017-04-01 & 46,888 & 5,209 \\
2 & 2016-01-01--2018-07-01 & 100,776 & 11,197 \\
3 & 2016-01-01--2019-10-01 & 151,187 & 16,798 \\
4 & 2016-01-01--2021-01-01 & 200,908 & 22,323 \\
5 & 2016-01-01--2022-04-01 & 252,486 & 28,053 \\
6 & 2016-01-01--2023-07-01 & 298,022 & 33,113 \\
7 & 2016-01-01--2023-11-09 & 310,147 & 34,460 \\
\bottomrule
\end{tabular}
\end{table}
}

\subsection{Chronological Fold Stability}

Three alternative temporal designs were compared with the primary design, whose fits are unchanged. A strict-date design moves every sale on a boundary date to the evaluation side (35--158 sales per boundary; the 2025 boundary is already strict, so the 2025 results are identical) and refits both families at 27 positive penalties (about every fourth tested value, including the display anchors) and $\rho=0$. An oracle diagnostic, which cannot be implemented prospectively, removes from each training block every parcel that reappears in its evaluation block (1.1--3.8\% of training sales) and is refitted at the same penalties. An unseen-parcel view refits nothing: it restricts the stored predictions to evaluation sales whose parcel is absent from the training block (66.6--90.5\% of each block).

Under the strict-date design, $R^2_P$ changes by at most 0.0041 and $\beta_{\log}$ by at most 0.006, and the slope path keeps its shape; its only sign difference is a held-out Surrogate maximum within 0.001 of zero. The Direct--Surrogate ordering of $\beta_{\log}$ changes only at small penalties, where the two paths nearly coincide. Targeted refits at 21 additional small penalties, and 9 large penalties for the Surrogate sign check (459 fits), found ordering reversals only where the Direct--Surrogate gap is below the matching tolerance $\tau=0.002$ under at least one of the two designs. The oracle diagnostic makes the uncorrected origin more regressive ($\beta_{\log}$ at $\rho=0$ lower by 0.012--0.029) and weakens the Surrogate's distance-correlation rebound, but leaves the other within-path conclusions unchanged. On never-seen parcels, the Surrogate still attains more first-order correction than Direct at 24--26 of the 27 penalties.

\begin{figure}[!htbp]
\centering
\safeincludegraphics[width=0.8\textwidth]{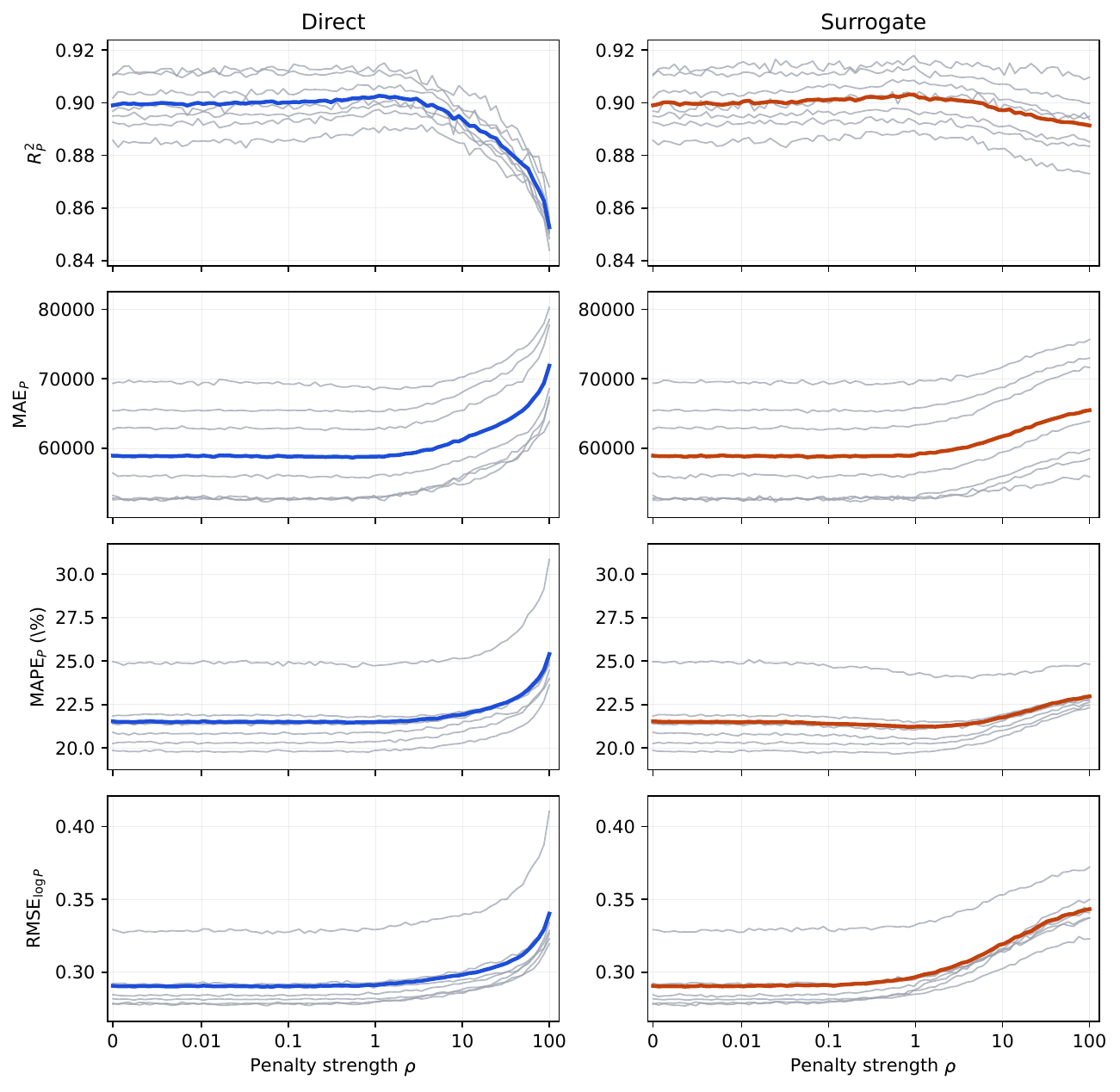}
\caption{Chronological-fold predictive-metric paths (thin gray) and equal-weight CV means (thick). The seven folds are overlapping cumulative-window constructions rather than independent replications, so their spread is descriptive and supports no standard error. The leftmost tick, 0, marks the $\rho=0$ fit and is not on the logarithmic scale.}
\label{fig:cv_predictive_metric_paths}
\end{figure}\begin{figure}[!htbp]
\centering
\safeincludegraphics[width=0.8\textwidth]{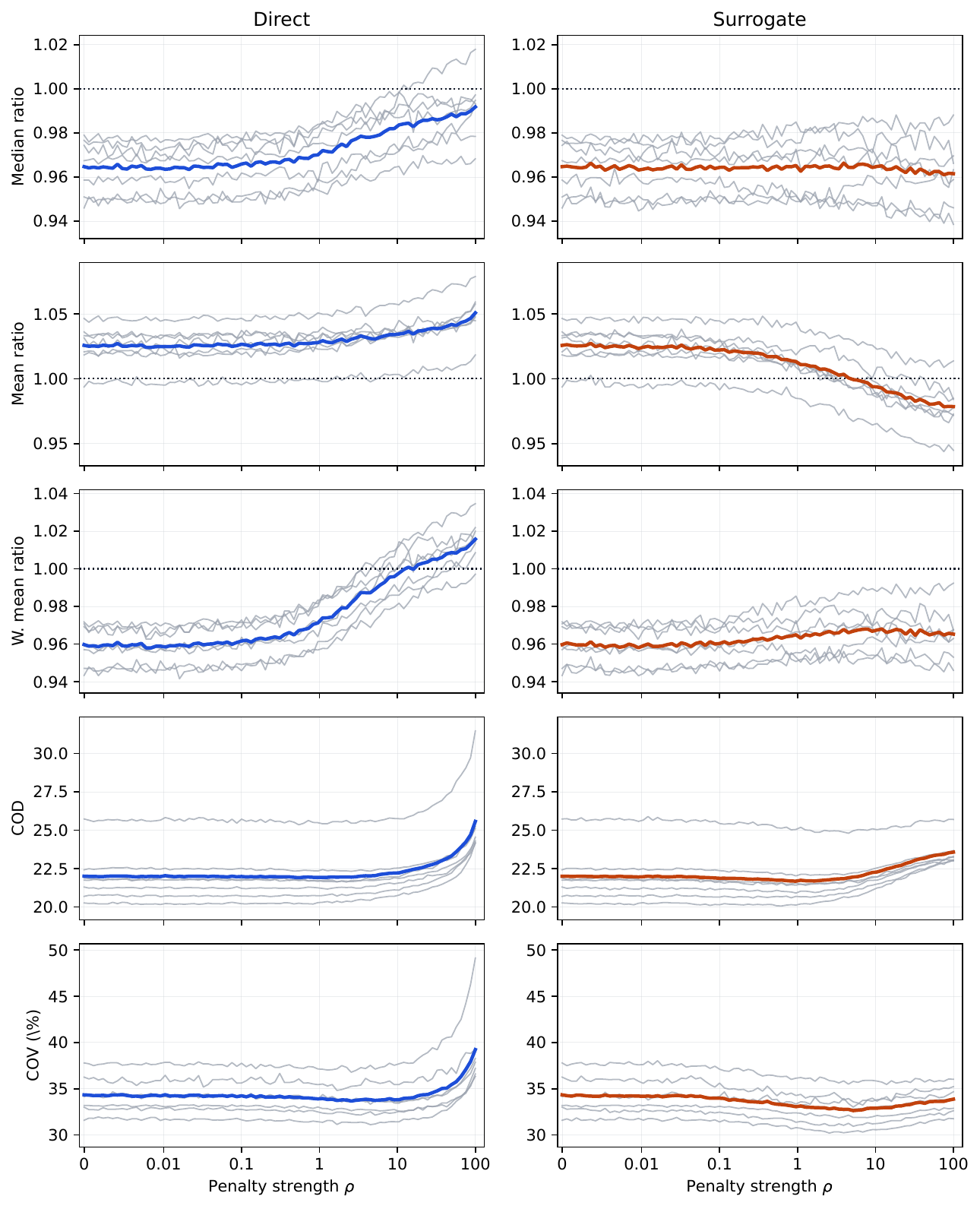}
\caption{Chronological-fold valuation-level and uniformity paths (thin gray) and equal-weight CV means (thick). Valuation-level panels include the ratio $=1$ reference. The leftmost tick, 0, marks the $\rho=0$ fit and is not on the logarithmic scale.}
\label{fig:cv_level_uniformity_paths}
\end{figure}\begin{figure}[!htbp]
\centering
\safeincludegraphics[width=0.8\textwidth]{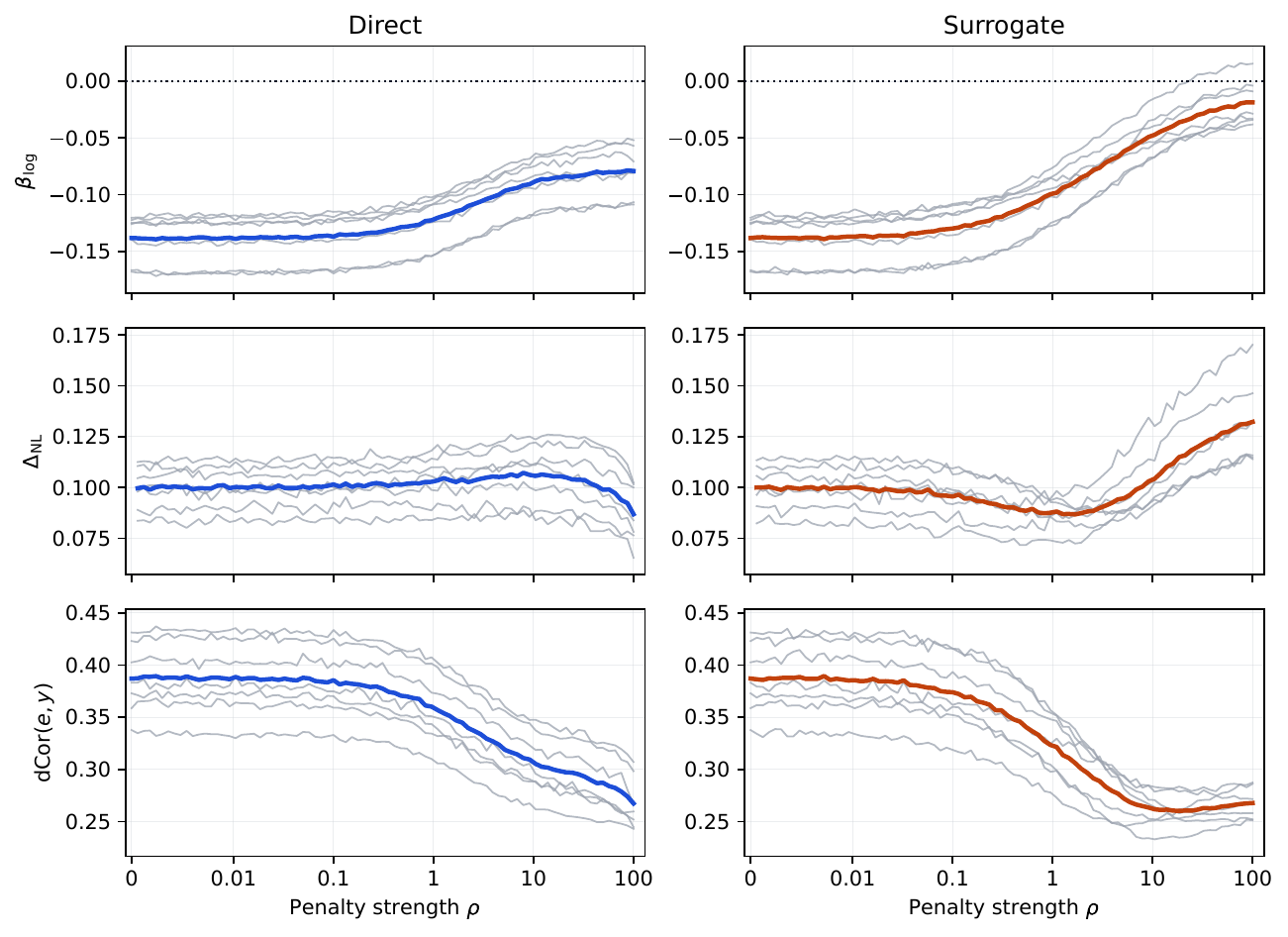}
\caption{Chronological-fold mechanism paths for $\beta_{\log}$, $\Delta_{\mathrm{NL}}$, and distance correlation (thin gray) and equal-weight CV means (thick). The $\beta_{\log}=0$ line is a first-order neutrality reference; $\Delta_{\mathrm{NL}}$ and dCor are not given forced zero lines. CV $\Delta_{\mathrm{NL}}$ uses the stored validation predictions and the same fixed estimator as the test sets; it did not define the $\rho$ grid. The leftmost tick, 0, marks the $\rho=0$ fit and is not on the logarithmic scale.}
\label{fig:cv_mechanism_metric_paths}
\end{figure}

\subsection{Change-Point Regions, Tolerance Bands, and Selection Rules}
\label{subsec:screening_interpretation}

For the forward-chaining check, each procedure, including its tolerance band, was recomputed on folds 1--$k$, and its choice was scored on fold $k+1$ against that fold's $\rho=0$ fit. CCAO used $k=3,4,5$ and the eight external counties used $k=4,5,6$, giving 54 county--method--horizon cases; the final CCAO folds overlap. Among non-abstaining cases, all cost limits held on the next fold in 79\% for the constrained rule (15 abstentions), 92\% for the Nash rule, 96\% for a 10\% $|\mathrm{PRB}|$ target, 74\% for the nine-diagnostic Nash rule of Table~\ref{tab:joint_posthoc_comparison}, and 57\% for maximizing the $|\mathrm{PRB}|$ reduction within the same limits. The smallest positive penalty met them in 91\%; all its failures are in $1-R^2_P$, the noisiest cost across folds. Fold $k+1$ was part of development, so this is not a prospective test.

For the tolerance band, relative increases from the $\rho=0$ fit are limited to $\tau=2\%$ for $1-R^2_P$, MAE, MAPE, log-price RMSE, and COD, and to 10\% for $\Delta_{\mathrm{NL}}$. The distance of the median ratio from one may increase by at most 0.01. The band starts at the first tested point meeting these limits and reducing $|\mathrm{PRB}|$ by at least 10\%, and ends before the first subsequent point that violates a limit. In the selection study, which also covers the eight external counties, a band exists for all 18 county--method units, including the two in which the change-point screen abstains (Allegheny Direct and Middlesex Surrogate; Appendix~\ref{app:external_benchmark}). Under the sale bootstrap, the 5--95\% range of its upper end lies within two grid steps of the full-sample end, and leave-one-fold-out recalculations move it by at most two. Log-price RMSE, MAPE, or COD bounds the Surrogate bands (jointly with $\Delta_{\mathrm{NL}}$ in St.~Louis County and with the median-ratio limit in Wayne); $\Delta_{\mathrm{NL}}$, COD, MAPE, or MAE bounds the Direct bands.

The observed curves have a useful ordering: small penalties change little,
intermediate penalties can improve first-order and assessor-facing measures
while retaining predictive fit, and larger penalties can worsen prediction
or leave a new nonlinear ratio pattern. The change-point regions in
Figure~\ref{fig:candidate_region_cv} summarize this ordering on the
chronological-CV paths. They are descriptive summaries, not safe intervals,
and raw penalty scales differ between Direct and Surrogate.

\paragraph{Change-point screen.} The screen uses only the seven-fold equal-weight CV paths
at the 82 tested positive penalties, on the coordinate $\log_{10}\rho$ and
without interpolation or refitting. Continuous piecewise-linear fits with up
to three segments, breakpoints at tested points, BIC selection, and a minimum
of five points per segment locate supported changes of regime; every event is
checked against the raw path. The \emph{activity onset} is the first tested
$\rho$ by which at least three of the five neutral-distance measures
$|\mathrm{PRD}-1|$, $|\mathrm{PRB}|$, $|\mathrm{MKI}-1|$, $|\mathrm{VEI}|$,
and $|\beta_{\log}|$ have entered a supported improving regime. For Direct,
the \emph{upper guardrail} is the first point of the earliest cluster, no
wider than two median log-grid steps, of sustained deterioration breakpoints
in at least two of $R^2_P$, MAE, MAPE, and $\operatorname{RMSE}_{\log P}$,
measured against the custom-objective $\rho=0$ fit. For the Surrogate, that
first predictive bend is recorded but does not bind. The guardrail is the
earlier of (i) a cluster of predictive-harm events, in which a deteriorating
metric has also lost its advantage over the $\rho=0$ fit, and (ii) a stable
rebound of CV $\Delta_{\mathrm{NL}}$ after its minimum. An endpoint is
leave-one-fold-out (LOFO) stable if at least five of seven recalculations
detect the same event within four median log-grid steps. The rule contains
no CCAO-specific penalty values: multiplying every $\rho$ by a constant
leaves the selected grid indices unchanged, and synthetic paths check that it
abstains when no guardrail is supported.

\paragraph{Screen outcomes.} Table~\ref{tab:candidate_region_rule}
reports the outcome. The Direct guardrail is the generic predictive cluster.
The Surrogate's predictive bend, $\rho=0.829$, does not bind, because at
the bend the Surrogate still has higher CV $R^2_P$ and lower MAPE than its
$\rho=0$ fit. MAE, however, is above its $\rho=0$ value at $\rho=0.625$ and
at every point from $\rho=0.954$. The Surrogate predictive-harm cluster ($R^2_P$ and MAPE) occurs
later, at $\rho=6.866$, so the $\Delta_{\mathrm{NL}}$ rebound binds. Its CV
minimum is at $\rho=1.677$.

\begin{table}[!htbp]
\centering
\small
\caption{CV change-point screen: reported outcome and leave-one-fold-out (LOFO) recalculations. Grid indices count the 82 positive penalties from the smallest (index 0).}
\label{tab:candidate_region_rule}
\begin{tabularx}{\textwidth}{@{}lXX@{}}
\toprule
 & Direct & Surrogate \\
\midrule
Reported region & $[0.356,\,2.560]$; indices 41--55 (15 points) & $[0.202,\,2.223]$; indices 37--54 (18 points) \\
Activity measures & PRD, PRB, MKI, VEI, $\beta_{\log}$ & PRD, PRB, VEI, $\beta_{\log}$ \\
Binding upper event & Predictive deterioration in $R^2_P$, MAE, MAPE, $\operatorname{RMSE}_{\log P}$ & Post-minimum $\Delta_{\mathrm{NL}}$ rebound; predictive bend at 0.829 nonbinding; harm cluster at 6.866 \\
LOFO activity onset & Indices 40--41 (6/7 at 41) & Indices 36--37 (6/7 at 37) \\
LOFO upper guardrail & Indices 53--56 (7/7 detected) & Indices 53--55 (rebound in 7/7) \\
\bottomrule
\end{tabularx}
\vspace{1mm}
\begin{minipage}{\textwidth}
\scriptsize
\emph{Notes.} LOFO recalculations reuse overlapping cumulative folds and are not independent tests. With fold~1 left out, the Surrogate has no qualifying predictive bend or harm cluster; its guardrail is still the $\Delta_{\mathrm{NL}}$ rebound. Penalty values are not comparable across families.
\end{minipage}
\end{table}

\paragraph{Every point inside the tolerance band.} Table~\ref{tab:candidate_region_points}
summarizes all 33 tested points inside the two tolerance bands of
Section~\ref{subsec:candidate_region_results} (17 Direct, 16 Surrogate)
against the shared custom-objective $\rho=0$ fit. Each value was recomputed
from the row-level predictions and matches the stored path table. On CV, PRB,
VEI, $\beta_{\log}$, and dCor improve at every point; $R^2_P$ is higher at
every point except Direct $\rho=3.907$, and COD is lower at 14 of 17 Direct
and all Surrogate points. MAE is higher than at the origin at 9 of 17 Direct
and 2 of 16 Surrogate points. CV $\operatorname{RMSE}_{\log P}$ is higher at
15 Direct and all 16 Surrogate points; it sets the upper end of the Surrogate
band. $\Delta_{\mathrm{NL}}$ moves in opposite directions: it is higher than
at the origin at every Direct point and lower at every Surrogate point. One
Surrogate point brings CV PRB inside $[-0.05,0.05]$, and no point brings VEI
inside $\pm10\%$ on CV, held-out, or 2025 sales. Direct raises the median
ratio toward one; the Surrogate leaves it near its origin value. Fold by
fold, PRB, VEI, and $\beta_{\log}$ improve at every band point in every fold;
MAE improves in about half of the point--fold pairs (46\% Direct, 58\%
Surrogate), and $\Delta_{\mathrm{NL}}$ in 16\% of Direct and all Surrogate
pairs.

{\begin{table}[!htbp]
\centering
\small
\setlength{\tabcolsep}{4pt}
\caption{Tested points inside the CV tolerance bands (Direct $[0.471,4.498]$, Surrogate $[0.115,0.954]$) relative to the custom-objective $\rho=0$ fit. Ranges are over the 17 Direct and 16 Surrogate band points on CV (equal-weight fold means). Counts give the band points that improve on the $\rho=0$ value in CV / held-out / 2025.}
\label{tab:candidate_region_points}
\begin{tabular}{@{}lrrrcc@{}}
\toprule
 & $\rho=0$ & Direct & Surrogate & \multicolumn{2}{c}{Points improving} \\
\cmidrule(lr){5-6}
Measure & CV & CV range & CV range & Direct (of 17) & Surrogate (of 16) \\
\midrule
$R^2_P$ & 0.8990 & 0.8988--0.9026 & 0.9008--0.9031 & 16/17/17 & 16/16/16 \\
MAE (\$) & 58,951 & 58,668--59,826 & 58,821--58,999 & 8/17/17 & 14/15/16 \\
$\operatorname{RMSE}_{\log P}$ & 0.2907 & 0.2906--0.2952 & 0.2907--0.2963 & 2/5/17 & 0/0/2 \\
PRB & $-0.087$ & $-0.078$ to $-0.050$ & $-0.078$ to $-0.050$ & 17/17/17 & 16/16/16 \\
VEI (\%) & $-26.8$ & $-23.7$ to $-14.3$ & $-25.0$ to $-16.0$ & 17/17/17 & 16/16/16 \\
$\beta_{\log}$ & $-0.138$ & $-0.129$ to $-0.100$ & $-0.129$ to $-0.100$ & 17/17/17 & 16/16/16 \\
Median ratio & 0.965 & 0.967--0.979 & 0.963--0.965 & 17/17/17 & 4/8/16 \\
COD & 22.01 & 21.91--22.04 & 21.69--21.88 & 14/10/17 & 16/16/16 \\
$\Delta_{\mathrm{NL}}$ & 0.1002$^{a}$ & 0.1013--0.1054 & 0.0873--0.0965 & 0/0/0 & 16/16/16 \\
dCor & 0.387 & 0.324--0.371 & 0.323--0.372 & 17/17/17 & 16/16/16 \\
\bottomrule
\end{tabular}
\vspace{1mm}
\begin{minipage}{\textwidth}
\scriptsize
\emph{Notes.} Improvement means higher $R^2_P$; lower MAE, $\operatorname{RMSE}_{\log P}$, COD, $\Delta_{\mathrm{NL}}$, and dCor; PRB, VEI, and $\beta_{\log}$ closer to zero; and a median ratio closer to one. $^{a}$The stored CV $\Delta_{\mathrm{NL}}$ tables cover positive penalties only; the $\rho=0$ CV value is the value recorded for the shared $\rho=0$ fit, which recomputation from the stored fold predictions with the same estimator reproduces exactly.
\end{minipage}
\end{table}
}

\begin{figure}[!htbp]
\centering
\safeincludegraphics[width=0.95\textwidth]{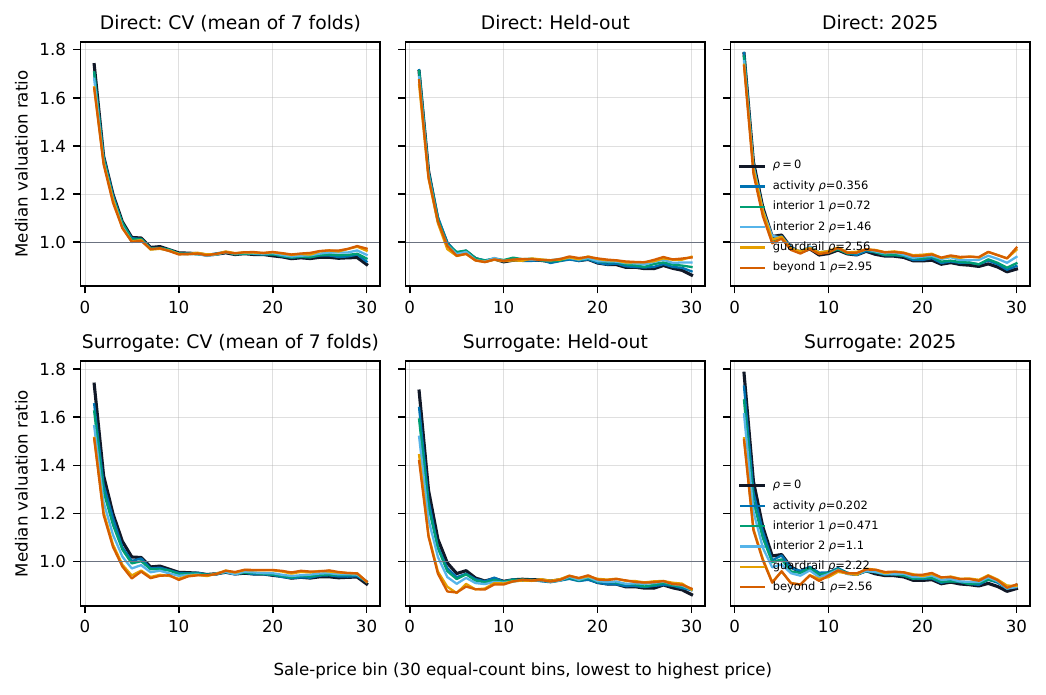}
\caption{Median valuation-to-sale ratios in 30 equal-count sale-price bins at the custom-objective $\rho=0$ fit and at points in and just beyond each descriptive CV change-point region: CV (mean of the seven fold profiles by bin rank), held-out, and 2025. Fold-specific bin cutoffs need not match in dollars. The line at 1 denotes equal valuation and sale price.}
\label{fig:candidate_region_ratio_profiles}
\end{figure}

\paragraph{Ratio shape.} For the descriptive change-point regions,
Figure~\ref{fig:candidate_region_ratio_profiles} shows 30-bin median-ratio
profiles at the origin, the activity onset, two interior points, the
guardrail, and the first point beyond it; trough depths at the
tolerance-band choices are in Table~\ref{tab:cv_selection_ci}. Inside the
Direct region the profiles flatten at high prices and develop only a shallow
dip (held-out depth at most $0.026$, measured as below).
Inside the Surrogate region a dip in the lower-price bins develops
through the upper half. Measured as the depth of the deepest local
depression in the 30-bin profile, it grows from $0.015$ at the origin to
$0.072$ at the guardrail on held-out sales and from $0.020$ to $0.066$ in
2025. The median across the seven CV folds grows from $0.027$ to $0.073$.
This descriptive depth is the largest gap between a bin's median ratio
and the smaller of the highest medians on either side of that bin; it depends
on the choice of 30 bins and has no assessor-facing threshold. The first
sale-price bin also stays well above a ratio of one in the displayed profiles.
The Surrogate guardrail marks the $\Delta_{\mathrm{NL}}$ rebound, although
the local dip has already grown inside the region. A near-zero PRB or lower
global $\Delta_{\mathrm{NL}}$ cannot replace a ratio-shape check.

\paragraph{Selection rules inside the tolerance band.} Table~\ref{tab:cv_selected_comparison} uses, for each family and rule, one tested grid point chosen inside the family's tolerance band from the equal-weight CV means. The criteria form four families: prediction ($1-R^2_P$, MAE, MAPE, $\operatorname{RMSE}_{\log P}$), dispersion (COD), regressivity (PRD, PRB, VEI, MKI), and nonlinear structure ($\Delta_{\mathrm{NL}}$); valuation level enters only through the band's median-ratio limit. $\beta_{\log}$ is reported as a mechanism diagnostic: removing it from the regressivity family changes none of the constrained or Nash choices in the 12 units of Cook and the five counties first studied (Philadelphia, Middlesex, Allegheny, St.~Louis County, and Miami-Dade). Each cost is scored as its relative deterioration from $\rho=0$ divided by its tolerance, and as zero if it improves; the prediction loss is the mean over its four metrics. The regressivity loss is the mean, over its four metrics, of the fraction of the $\rho=0$ distance from neutrality that remains. Every family loss $L_g$ therefore equals one at its disagreement point: $\rho=0$ for regressivity and the tolerance limit for costs. The \emph{constrained} rule selects the smallest penalty in the band that reduces $|\mathrm{PRB}|$ by at least 25\%, or abstains. The \emph{Nash} rule maximizes $\sum_g\log(1-L_g)$ over points where every $L_g<1$; minimizing $\sum_g L_g^2$ gives the same choices in all 12 of these units. Because the disagreement point does not depend on the candidates, restricting the candidates to a subset that still contains the Nash choice leaves it unchanged; this held in all 645 tested cases. The nine-diagnostic Nash and Euclidean utopia rules used for Table~\ref{tab:joint_posthoc_comparison} instead scale each diagnostic between its best and worst values over the candidates, so their choice can change when the candidate set changes.

Table~\ref{tab:cv_selection_rules} also reports a budget rule, not used for Table~\ref{tab:cv_selected_comparison}, that maximizes the $|\mathrm{PRB}|$ reduction while every cost, including $\Delta_{\mathrm{NL}}$, stays within 5\% or 10\%. It selects points at the limit: across the 18 units, all costs stayed within the budget out of time in only 67\% (5\%) and 60\% (10\%) of evaluations. It is not a recommended setting.

\paragraph{Bootstrap and stability checks.} Every band and choice was recomputed on paired sale-bootstrap replicates (499 for the bands and the constrained rule, 200 for the other rules) and on the seven leave-one-fold-out paths. The bootstrap resamples distinct sales, so a sale that appears in both overlapping folds is drawn jointly, and each configuration is compared with the $\rho=0$ fit on the same replicate. Because it reuses the stored predictions, the bootstrap conditions on the fitted models and does not capture uncertainty from refitting them. Table~\ref{tab:cv_selection_rules} reports how often these recalculations return the same point or one within one grid step, and Table~\ref{tab:cv_selection_ci} gives 90\% percentile intervals for every CV change at Direct $\rho=1.677$ and Surrogate $\rho=0.954$ (Table~\ref{tab:cv_selected_comparison}). At the Nash choices of these 12 units, the sale-bootstrap standard errors of the relative changes are 0.04--0.22\% for MAE, log-price RMSE, and COD, 0.3--1.4\% for $1-R^2_P$, and 0.2--0.9\% for $\Delta_{\mathrm{NL}}$, below their tolerances of 2\% and 10\%, so the tolerances are kept fixed rather than widened by standard errors. Resampling sale weeks instead of sales gives standard errors 0.8--1.5 times as large, and the PRB changes are 92--105 times their sale-bootstrap standard error in Cook County and 30--169 times in the five external counties. On the held-out and 2025 samples, a separate paired sale bootstrap within each sample (200 replicates) gives the intervals summarized in Section~\ref{subsec:candidate_region_results}. Requiring every limit to hold jointly in 90\% of replicates leaves both constrained choices unchanged. The ratio-profile trough is reported as a check rather than a criterion.

\begin{table}[!htbp]
\centering
\small
\caption{Development-CV selection rules for Cook County: chosen penalty, achieved $|\mathrm{PRB}|$ reduction, and stability.}
\label{tab:cv_selection_rules}

\setlength{\tabcolsep}{3.5pt}
\renewcommand{\arraystretch}{1.10}

\begin{tabular}{@{}llrrrrrc@{}}
\toprule
Rule & Family & $\rho$ & \makecell[r]{$|\mathrm{PRB}|$ red.\\CV / 2025 (\%)} & \makecell[r]{Largest CV\\cost (\%)} & \makecell[r]{LOFO\\same ($\pm1$)} & \makecell[r]{Boot\\$\pm1$} & \makecell[c]{Limits held\\held-out / 2025} \\
\midrule
Constrained, $|\mathrm{PRB}|$ $-25\%$ & Direct & 1.677 & 26 / 24 & 2.4 & 6/7 (7/7) & 1.00 & yes / yes \\
 & Surrogate & 0.409 & 26 / 22 & 1.0 & 4/7 (7/7) & 1.00 & yes / yes \\
\addlinespace[2pt]
Nash, tolerance disagreement & Direct & 2.223 & 31 / 27 & 2.4 & 6/7 (7/7) & 0.93 & yes / yes \\
 & Surrogate & 0.954 & 43 / 39 & 1.9 & 2/7 (7/7) & 0.98 & no / yes \\
\midrule
Budget, all costs $\leq5\%$ & Direct & 4.498 & 42 / 38 & 4.9 & 1/7 (1/7) & 0.46 & no / no \\
 & Surrogate & 2.947 & 67 / 69 & 4.9 & 3/7 (7/7) & 1.00 & no / yes \\
\addlinespace[2pt]
Budget, all costs $\leq10\%$ & Direct & 16.00 & 59 / 56 & 9.7 & 4/7 (6/7) & 0.48 & no / no \\
 & Surrogate & 10.48 & 97 / 100 & 9.9 & 2/7 (5/7) & 0.93 & no / no \\
\bottomrule
\end{tabular}

\vspace{1mm}
\begin{minipage}{\textwidth}
\footnotesize
\emph{Notes.}
The constrained and Nash rules choose among tested grid points inside the $\tau=2\%$ tolerance band (Direct $[0.471,4.498]$, Surrogate $[0.115,0.954]$); the budget rules choose among all tested penalties. The largest CV cost is the largest relative increase among $1-R^2_P$, MAE, MAPE, $\operatorname{RMSE}_{\log P}$, COD, and $\Delta_{\mathrm{NL}}$; for the first two rules $\Delta_{\mathrm{NL}}$ may rise by 10\%. LOFO counts the seven leave-one-fold-out recalculations returning the same point (within one grid step); the folds overlap and are not independent replications. Boot $\pm1$ is the share of paired sale-bootstrap replicates choosing a point within one grid step (499 replicates for the constrained rule, 200 for the others). ``Limits held'' asks whether every cost stays within the rule's own limits on the later sample. Penalty values are not comparable across families.
\end{minipage}
\end{table}

\begin{table}[!htbp]
\centering
\small
\caption{Development-CV changes at Direct $\rho=1.677$ and Surrogate $\rho=0.954$ (Table~\ref{tab:cv_selected_comparison}), relative to the custom-objective $\rho=0$ fit, with paired sale-bootstrap 90\% intervals.}
\label{tab:cv_selection_ci}

\setlength{\tabcolsep}{5pt}
\renewcommand{\arraystretch}{1.08}

\begin{tabular}{@{}lrrrr@{}}
\toprule
& \multicolumn{2}{c}{Direct $\rho=1.677$} & \multicolumn{2}{c}{Surrogate $\rho=0.954$} \\
\cmidrule(lr){2-3}\cmidrule(l){4-5}
Measure & Change (\%) & 90\% interval & Change (\%) & 90\% interval \\
\midrule
$1-R^2_P$ & $-3.2$ & $[-4.6,\,-1.6]$ & $-4.0$ & $[-6.2,\,-2.3]$ \\
MAE & $+0.1$ & $[-0.1,\,+0.3]$ & $+0.1$ & $[-0.2,\,+0.3]$ \\
MAPE & $0.0$ & $[-0.2,\,+0.2]$ & $-1.5$ & $[-1.7,\,-1.2]$ \\
$\operatorname{RMSE}_{\log P}$ & $+0.6$ & $[+0.4,\,+0.7]$ & $+1.9$ & $[+1.7,\,+2.2]$ \\
COD & $-0.3$ & $[-0.5,\,-0.1]$ & $-1.5$ & $[-1.7,\,-1.2]$ \\
\addlinespace[2pt]
PRD & $-22.0$ & $[-22.6,\,-21.4]$ & $-27.2$ & $[-28.0,\,-26.5]$ \\
PRB & $-26.0$ & $[-26.5,\,-25.5]$ & $-42.6$ & $[-43.4,\,-41.9]$ \\
VEI & $-27.6$ & $[-30.4,\,-25.0]$ & $-40.2$ & $[-44.1,\,-37.7]$ \\
MKI & $-36.5$ & $[-37.3,\,-35.9]$ & $-35.0$ & $[-36.0,\,-34.1]$ \\
$\beta_{\log}$ & $-17.0$ & $[-17.3,\,-16.7]$ & $-28.0$ & $[-28.5,\,-27.6]$ \\
\addlinespace[2pt]
$\Delta_{\mathrm{NL}}$ & $+2.4$ & $[+1.7,\,+3.1]$ & $-12.6$ & $[-13.7,\,-11.5]$ \\
Median ratio & $-24.9$ & $[-26.9,\,-22.9]$ & $-1.5$ & $[-4.1,\,+0.5]$ \\
Trough depth & $+35.1$ & $[+9.3,\,+55.4]$ & $+27.0$ & $[+14.3,\,+70.1]$ \\
\bottomrule
\end{tabular}

\vspace{1mm}
\begin{minipage}{\textwidth}
\footnotesize
\emph{Notes.}
Relative change in each cost (for $R^2_P$, in $1-R^2_P$) and in each other measure's distance from neutrality (from one for PRD, MKI, and the median ratio; from zero for PRB, VEI, and $\beta_{\log}$); positive values are worse. Trough depth is the 30-bin profile depth defined in the ratio-shape paragraph above. All values are equal-weight seven-fold CV means, so they differ from the held-out and 2025 values in Table~\ref{tab:cv_selected_comparison}. The 499 replicates resample distinct sales and are paired across configurations.
\end{minipage}
\end{table}

\subsection{Centered-Spread Post-Hoc Comparator Detail}
\label{app:centered_spread_detail}

\paragraph{Construction.} The comparator rescales stored predictions about the mean log price of
their fitting block, $f_b(x)=\bar y_T+b\bigl(f_0(x)-\bar y_T\bigr)$, where $\bar y_T$ is the fold-$k$
training mean, the development-pool mean for the held-out block, or the 2016--2024 mean for the 2025
block. This is the form of Corollary~\ref{cor:path_scaling}; at $b=1$ predictions are unchanged, and
no model is refit. Centering instead on the block's mean prediction adds a constant to all log
predictions in a block, so it cannot change $\beta_{\log}$, COD, COV, or dCor; at the evaluated base
values it changes MAE by at most \$1.54.

\paragraph{Roots.} On the D1 coordinate, $\beta_{\log}$ is linear in $b$, so the root that sets it to
zero is solved in closed form from the fold validation covariances: $b_0=1.160458$ for the
custom-objective $\rho=0$ fit and $1.160367$ for ordinary LightGBM. The pooled out-of-fold root,
$1.167027$, is a sensitivity because the overlapping validation blocks count some sales twice. The
in-sample root $b^{\star}_{T}$ (Appendix~\ref{app:theory}) ranges from 1.065397 to 1.092817 over the fitting
blocks and would under-correct by about half in $b-1$, because the learner's in-sample slope
($-0.061383$ to $-0.084934$) is much flatter than its out-of-sample slope (about $-0.14$ to $-0.16$). The
grid extends to $b_{\max}=1.209082$. All roots use development predictions only.

Table~\ref{tab:centered_spread} reports the comparator at these base values for the primary base fit
on both test sets and for ordinary LightGBM on held-out sales. Because the map is affine, it can
match a first-order target exactly, but for $b>0$ it only rescales, by $b$, the nonlinear part of
the conditional mean of the log error; it cannot remove it.

{\begin{table}[!htbp]
\centering
\scriptsize
\setlength{\tabcolsep}{2.6pt}
\renewcommand{\arraystretch}{1.06}
\caption{Centered-spread post-hoc comparator at four base values. The map
is $f_b(x)=\bar y_T+b(f_0(x)-\bar y_T)$ applied to stored predictions; no model was refit.
D1 is the primary development coordinate and D2 the pooled out-of-fold sensitivity, so ``D1 root''
is the base value that drives the primary development coordinate to first-order neutrality and
``D2 root'' the same for the sensitivity coordinate. VEI is a May-2026 Exposure Draft diagnostic and
is proposed rather than adopted guidance, so no VEI entry is a compliance determination.}
\label{tab:centered_spread}
\resizebox{\textwidth}{!}{%
\begin{tabular}{@{}lrrrrrrrrrrr@{}}
\toprule
Anchor & $b$ & $R^2_P$ & MAE & $\operatorname{RMSE}_{\log P}$ & COD & PRD & MKI
 & VEI & $\beta_{\log}$ & $\Delta_{\mathrm{NL}}$ & dCor \\
\midrule
\addlinespace[2pt]
\multicolumn{12}{l}{\textit{Panel A: \emph{Custom-objective $\rho=0$ origin} rescaled, held-out (primary)}} \\
\addlinespace[2pt]
\cmidrule(lr){1-12}
$b=1$ (no rescaling) & 1.000 & 0.8928 & \$75,976 & 0.2894 & 21.60 & 1.0690 & 0.9230 & -26.17 & -0.1474 & 0.1164 & 0.3822 \\
D1 root & 1.160 & 0.8063 & \$87,182 & 0.3071 & 22.59 & 0.9871 & 1.0726 & 14.77 & -0.0106 & 0.1343 & 0.2518 \\
D2 root & 1.167 & 0.7941 & \$88,689 & 0.3087 & 22.69 & 0.9839 & 1.0787 & 16.44 & -0.0050 & 0.1344 & 0.2526 \\
$b_{\max}$ & 1.209 & 0.6926 & \$99,978 & 0.3200 & 23.50 & 0.9630 & 1.1175 & 26.82 & 0.0309 & 0.1338 & 0.2681 \\
\addlinespace[2pt]
\multicolumn{12}{l}{\textit{Panel B: \emph{Custom-objective $\rho=0$ origin} rescaled, 2025 forward (primary)}} \\
\addlinespace[2pt]
\cmidrule(lr){1-12}
$b=1$ (no rescaling) & 1.000 & 0.9039 & \$79,139 & 0.2791 & 21.23 & 1.0775 & 0.9093 & -28.41 & -0.1612 & 0.1212 & 0.4168 \\
D1 root & 1.160 & 0.7991 & \$96,243 & 0.2939 & 21.91 & 0.9934 & 1.0581 & 11.37 & -0.0266 & 0.1454 & 0.2542 \\
D2 root & 1.167 & 0.7850 & \$98,248 & 0.2955 & 22.01 & 0.9900 & 1.0642 & 12.97 & -0.0211 & 0.1457 & 0.2536 \\
$b_{\max}$ & 1.209 & 0.6682 & \$112,802 & 0.3069 & 22.75 & 0.9685 & 1.1029 & 23.74 & 0.0142 & 0.1459 & 0.2619 \\
\addlinespace[2pt]
\multicolumn{12}{l}{\textit{Panel C: \emph{Ordinary LightGBM (standard raw-label native)} rescaled, held-out (secondary, practical)}} \\
\addlinespace[2pt]
\cmidrule(lr){1-12}
$b=1$ (no rescaling) & 1.000 & 0.8942 & \$75,655 & 0.2887 & 21.63 & 1.0695 & 0.9230 & -26.46 & -0.1496 & 0.1187 & 0.3871 \\
D1 root & 1.160 & 0.8024 & \$87,658 & 0.3060 & 22.61 & 0.9875 & 1.0728 & 14.34 & -0.0132 & 0.1375 & 0.2531 \\
D2 root & 1.167 & 0.7892 & \$89,277 & 0.3076 & 22.73 & 0.9840 & 1.0792 & 16.18 & -0.0074 & 0.1376 & 0.2537 \\
$b_{\max}$ & 1.209 & 0.6852 & \$100,604 & 0.3188 & 23.53 & 0.9633 & 1.1180 & 27.33 & 0.0282 & 0.1371 & 0.2678 \\
\bottomrule
\end{tabular}}
\vspace{1mm}
\begin{minipage}{\textwidth}
\scriptsize
\emph{Notes.}
Panels A and B rescale the \emph{Custom-objective $\rho=0$ origin}, the primary post-hoc
base because it is the common origin of both penalized families, so rescaling it holds the execution
path fixed. Panel~C rescales \emph{Ordinary LightGBM (standard raw-label native)} and is a
secondary variant that sets neither the comparison range nor the targets. \emph{Centered-label native
L2 (initialization-aligned)} has no post-hoc path. At $b=1$ the map returns the cached prediction
array unchanged, so the first row of each panel is the reference cell itself. VEI is reported in
percent and COD in percent; the pooled COD values do not establish property-class compliance. The alternative
centering about the fitting block's mean prediction is a named sensitivity evaluated at these same
four base values; its analytic difference from the map used here is constant in $x$, so it cannot
move $\beta_{\log}$, COD, COV or dCor at all. $\Delta_{\mathrm{NL}}$ and dCor carry no
assessor-facing reference range.
\end{minipage}
\end{table}
}

\subsection{Matched First-Order Targets and Attainment}
\label{app:matched_beta_detail}

\paragraph{Target construction.} Matching uses the D1 coordinate. The three families share the
development range $[-0.138271,-0.078651]$, whose upper end is set by Direct. Six targets are placed
by probing equally spaced points and taking the achieved development $\beta_{\log}$ of the nearest
fitted Direct configuration, so Direct matches exactly; the Surrogate uses its nearest fitted penalty
and the post-hoc map an exact solve. The matching tolerance is $\tau=0.002$, and the largest matching gap is $1.06\times10^{-3}$. The $j=0$ row
is an identity check (largest relative disagreement $4.4\times10^{-12}$). Because only development
slopes are matched, the held-out differences are between approximately matched configurations, not
effects at equal out-of-sample correction.

\paragraph{Distinct-sale sensitivity.} Because the last two validation blocks overlap, we repeat the
construction on the one-sale-one-vote D3 coordinate. The two coordinates differ by up to 0.0049 in
$\beta_{\log}$, and 13 of the 24 matched configurations change (11 post-hoc re-solves and two fitted
penalties at $j=2$). Of 315 headline sign comparisons, 259 agree, 41 fall below the materiality floor
on both coordinates, 5 are structural zeros, and 10 reverse, all in small differences (the largest is
held-out MAE at $j=2$, from $+$\$15 to $-$\$149). No $\Delta_{\mathrm{NL}}$ comparison changes sign,
and Direct's degradation at $j=5$ is unchanged; the Surrogate-versus-post-hoc $R^2_P$ near-tie at
$j=5$ reverses and is not interpreted. Fold spreads are descriptive, not standard errors.

{\begin{table}[!htbp]
\centering
\scriptsize
\setlength{\tabcolsep}{4pt}
\renewcommand{\arraystretch}{1.06}
\caption{Matched first-order comparison on the primary development coordinate D1.
Panel~A gives the six comparison targets and the configuration each family used; Panel~B gives the
held-out outcome at those targets for the three primary families. Every configuration is an actually
fitted model or an exactly solved rescaling, and no metric is interpolated between penalty values.
The match is on the \emph{development} coordinate and does not equalize held-out first-order
correction.}
\label{tab:matched_beta}
\resizebox{\textwidth}{!}{%
\begin{tabular}{@{}llrrrrrr@{}}
\toprule
\multicolumn{8}{l}{\textit{Panel A: comparison targets and the matched configurations}} \\
\cmidrule(lr){1-8}
$j$ & Target $\beta_{\log}$ (D1) & Direct $\rho$ & Surrogate $\rho$ & C-posthoc $b$
 & A-posthoc $b$ & & \\
\midrule
0 & -0.138 & 0 & 0 & 1.000000 & 0.999922 \\
1 & -0.126 & 0.720 & 0.133 & 1.013671 & 1.013592 \\
2 & -0.115 & 1.677 & 0.409 & 1.027316 & 1.027235 \\
3 & -0.102 & 3.907 & 0.829 & 1.042053 & 1.041972 \\
4 & -0.091 & 9.103 & 1.456 & 1.054971 & 1.054888 \\
5 & -0.079 & 86.851 & 2.560 & 1.069186 & 1.069103 \\
\midrule
\multicolumn{8}{l}{\textit{Panel B: held-out outcome at those targets}} \\
\cmidrule(lr){1-8}
$j$ & Family & $R^2_P$ & MAE & $\operatorname{RMSE}_{\log P}$ & COD
 & $\Delta_{\mathrm{NL}}$ & dCor \\
\midrule
\addlinespace[2pt]
0 & Direct & 0.8928 & \$75,976 & 0.2894 & 21.60 & 0.1164 & 0.3822 \\
 & Surrogate & 0.8928 & \$75,976 & 0.2894 & 21.60 & 0.1164 & 0.3822 \\
 & C-posthoc & 0.8928 & \$75,976 & 0.2894 & 21.60 & 0.1164 & 0.3822 \\
\addlinespace[2pt]
1 & Direct & 0.8984 & \$74,516 & 0.2888 & 21.60 & 0.1216 & 0.3637 \\
 & Surrogate & 0.8954 & \$75,416 & 0.2909 & 21.50 & 0.1132 & 0.3660 \\
 & C-posthoc & 0.8972 & \$74,897 & 0.2893 & 21.55 & 0.1192 & 0.3625 \\
\addlinespace[2pt]
2 & Direct & 0.9006 & \$74,167 & 0.2900 & 21.57 & 0.1231 & 0.3424 \\
 & Surrogate & 0.8948 & \$75,831 & 0.2948 & 21.36 & 0.1035 & 0.3358 \\
 & C-posthoc & 0.9000 & \$74,151 & 0.2895 & 21.52 & 0.1218 & 0.3438 \\
\addlinespace[2pt]
3 & Direct & 0.8993 & \$74,164 & 0.2910 & 21.70 & 0.1292 & 0.3227 \\
 & Surrogate & 0.8963 & \$75,505 & 0.2976 & 21.40 & 0.0983 & 0.3134 \\
 & C-posthoc & 0.9009 & \$73,725 & 0.2900 & 21.52 & 0.1243 & 0.3248 \\
\addlinespace[2pt]
4 & Direct & 0.8958 & \$75,296 & 0.2937 & 21.93 & 0.1298 & 0.3017 \\
 & Surrogate & 0.8955 & \$76,083 & 0.3030 & 21.39 & 0.0938 & 0.2938 \\
 & C-posthoc & 0.8999 & \$73,755 & 0.2908 & 21.55 & 0.1263 & 0.3096 \\
\addlinespace[2pt]
5 & Direct & 0.8760 & \$82,764 & 0.3170 & 23.97 & 0.1198 & 0.2730 \\
 & Surrogate & 0.8963 & \$76,128 & 0.3094 & 21.40 & 0.0914 & 0.2714 \\
 & C-posthoc & 0.8966 & \$74,236 & 0.2920 & 21.61 & 0.1282 & 0.2944 \\
\bottomrule
\end{tabular}}
\vspace{1mm}
\begin{minipage}{\textwidth}
\scriptsize
\emph{Notes.}
Targets are placed by a deterministic rule: probe equally spaced points in the
three-way development range, then take as the target the achieved development $\beta_{\log}$ of the
fitted Direct configuration nearest the probe. The target is therefore an achieved value rather than
the probe, Direct matches exactly by construction, and no selection discretion enters. Surrogate
rows use the nearest actually fitted penalty value and the post-hoc rows an exact solve, all within
the tolerance $\tau=0.002$. The $j=0$ row is an origin identity check: Direct at
$\rho=0$, Surrogate at $\rho=0$ and the post-hoc map at $b=1$ are the same object. A-posthoc is the
rescaling of \emph{Ordinary LightGBM (standard raw-label native)}, a secondary variant that sets neither the range nor the targets; its outcomes are not
reported. Direct and Surrogate penalty magnitudes are not comparable across families. COD is in percent
and lies outside the adopted $[5,15]$ range throughout. $\Delta_{\mathrm{NL}}$ and dCor are
residual-structure diagnostics with no assessor-facing reference range.
\end{minipage}
\end{table}
}

{\begin{table}[!htbp]
\centering
\scriptsize
\setlength{\tabcolsep}{3.4pt}
\renewcommand{\arraystretch}{1.06}
\caption{Extended development targets, which lie outside the three-way common range,
with an explicit attainment state for every family. Held-out evaluation. Rows recorded as not
attained carry blank outcome columns: they are neither dropped nor filled by interpolation. The
states describe the penalty grid this experiment evaluated, not the objectives and not penalty
values the experiment never examined.}
\label{tab:matched_beta_ext}
\begin{tabular}{@{}llllrrrrrrr@{}}
\toprule
Target $\beta_{\log}$ (D1) & Family & Configuration & State & Max attained $\beta_{\log}$
 & $R^2_P$ & MAE & $\operatorname{RMSE}_{\log P}$ & COD & $\Delta_{\mathrm{NL}}$ & dCor \\
\midrule
\addlinespace[2pt]
-0.06 & Direct & --- & \verb|NOT_ATTAINED| & -0.079 & --- & --- & --- & --- & --- & --- \\
 & Surrogate & 5.964 & attained & --- & 0.8934 & \$77,813 & 0.3230 & 21.81 & 0.0968 & 0.2518 \\
 & C-posthoc & 1.091 & attained & --- & 0.8868 & \$75,847 & 0.2944 & 21.75 & 0.1306 & 0.2752 \\
\addlinespace[2pt]
-0.03 & Direct & --- & \verb|NOT_ATTAINED| & -0.079 & --- & --- & --- & --- & --- & --- \\
 & Surrogate & 28.118 & attained & --- & 0.8917 & \$79,939 & 0.3443 & 22.84 & 0.1126 & 0.2553 \\
 & C-posthoc & 1.126 & attained & --- & 0.8569 & \$80,444 & 0.2999 & 22.10 & 0.1332 & 0.2559 \\
\addlinespace[2pt]
0.00 & Direct & --- & \verb|NOT_ATTAINED| & -0.079 & --- & --- & --- & --- & --- & --- \\
 & Surrogate & --- & \verb|NOT_ATTAINED| & -0.019 & --- & --- & --- & --- & --- & --- \\
 & C-posthoc & 1.160 & attained & --- & 0.8063 & \$87,182 & 0.3071 & 22.59 & 0.1343 & 0.2518 \\
\bottomrule
\end{tabular}
\vspace{1mm}
\begin{minipage}{\textwidth}
\scriptsize
\emph{Notes.}
Configuration is a fitted penalty value for the Surrogate rows and an exactly solved
base value for the post-hoc rows. The ``Max attained'' column is filled only for rows recorded as
not attained, and reports the largest development first-order correction that family reached over
the evaluated path. Over that path the Direct family did not attain any of the three extended
targets and the Surrogate family did not attain zero; both post-hoc families attain all three
exactly, because a one-parameter rescaling can be solved for any first-order target. Reaching a
target is not by itself an advantage: driving the post-hoc coordinate to neutrality costs the
price-scale accuracy recorded in Table~\ref{tab:centered_spread}. COD is in percent;
$\Delta_{\mathrm{NL}}$ and dCor carry no assessor-facing reference range.
\end{minipage}
\end{table}
}

\subsection{Retransformation Sensitivity}
\label{app:smearing}

Predictions are returned to dollars by direct exponentiation (Eq.~\eqref{eq:price_prediction}), a
convention rather than a conditional-mean correction. As a sensitivity, we apply a Duan smearing factor
\[
s=\frac{\sum_i w_i\exp(u_i)}{\sum_i w_i},
\qquad u_i=\log P_i-f(x_i),
\]
as $f(x_i)\mapsto f(x_i)+\log s$, estimated from one-sale-one-vote-weighted development out-of-fold
residuals and never re-estimated on an evaluation block. Note that $u_i=-e_i$. For the 2025 production refit, which has no out-of-fold analogue, the development-estimated factor is applied unchanged; this is an assumption of the sensitivity. Valuation-level ratios
scale by $s$; COD, COV, PRD, PRB, MKI, VEI, $\beta_{\log}$, $\Delta_{\mathrm{NL}}$, and dCor are
invariant to a common factor, so no uniformity, vertical-equity, or residual-structure result depends
on this choice. $R^2_P$, MAE, and MAPE change; $\operatorname{RMSE}_{\log P}$ is not recomputed.
Because the factor is common, it moves every 30-bin median ratio by the same proportion: for
Surrogate $\rho=0.954$, the held-out median ratio moves from 0.929 to 1.000 and the lowest-bin
ratio from 1.53 to 1.65; for Direct $\rho=1.677$, the corresponding values are 0.937 to 0.992 and
1.67 to 1.77.

Table~\ref{tab:smearing_matched_beta} shows the strongest matched target. The factor differs across
methods, largest for the Surrogate and smallest for post-hoc rescaling, and smearing reverses the
Surrogate-versus-post-hoc ordering in $R^2_P$ and MAE. That price-scale comparison is therefore
convention-dependent; the $\Delta_{\mathrm{NL}}$ and dCor comparisons are not.

\begin{table}[!ht]
\centering
\caption{Retransformation sensitivity at the matched first-order targets, development
coordinate D1 at the strongest common target, held-out evaluation. Baseline columns reproduce
Appendix Table~\ref{tab:matched_beta}; smeared columns apply the Duan factor $s$ to the
same stored predictions; nothing is refitted.}
\label{tab:smearing_matched_beta}
\begin{tabular}{@{}lrrrrr@{}}
\toprule
Family & $s$ & $R^2_P$ base & $R^2_P$ smeared & MAE base & MAE smeared \\
\midrule
Direct & 1.0636 & 0.8760 & 0.8498 & \$82,764 & \$87,979 \\
Surrogate & 1.0879 & 0.8963 & 0.9010 & \$76,128 & \$73,908 \\
C-posthoc & 1.0524 & 0.8966 & 0.8860 & \$74,236 & \$75,548 \\
\bottomrule
\end{tabular}
\end{table}

The same check applies to the price-scale gains of the exploratory CV-selected configurations in
Table~\ref{tab:cv_selected_comparison}. Table~\ref{tab:level_cv_selected} adds a second
development-estimated offset, $c_{\mathrm{med}}$, the one-sale-one-vote weighted median of the development
out-of-fold residuals $u_i$; it is a sensitivity and not part of either method. All configurations
in Table~\ref{tab:level_cv_selected} have median valuation ratios below one on both test sets: at $\rho=0$ the median valuation ratio is 0.928 on held-out and 0.946 on
2025 sales. Both corrections raise the valuation ratios of higher-priced sales, which dominate
price-scale squared error. For Direct $\rho=1.677$ and Surrogate $\rho=0.954$, with each configuration's own offset, the $R^2_P$ gains over $\rho=0$ shrink to
between $+0.0002$ and $+0.0021$ under the median-ratio offset and become changes between $-0.0041$ and
$-0.0001$ under smearing. The largest change is a loss, Direct on 2025 sales under smearing: $-0.0041$,
with paired PIN-cluster bootstrap 95\% interval $[-0.0068,-0.0015]$ (500 replicates; all intervals in this paragraph are 95\%). Their MAE reductions in
Table~\ref{tab:cv_selected_comparison} become changes between $-0.04\%$ and $+0.45\%$ under the
median-ratio offset and increases of 1.0--2.3\% under smearing. Two median-offset $R^2_P$ gains have
intervals that exclude zero, Direct on held-out sales ($+0.0021$, $[0.0004,0.0040]$) and Surrogate on 2025
sales ($+0.0021$, $[0.0001,0.0040]$); they reflect remaining under-prediction (calibrated median ratios 0.964
and 0.984) rather than shape. When each configuration receives its own least-squares common scale on the
evaluation sample, an infeasible benchmark used only for this decomposition, the $R^2_P$ differences from
$\rho=0$ lie between $-0.0017$ and $+0.0007$, and none is a gain distinguishable from zero. The PRB changes, between $+0.023$ and $+0.043$ with intervals no wider than $\pm0.002$, do not
depend on any common offset. For the other two picks, Direct $\rho=2.223$ and Surrogate $\rho=0.409$, the $R^2_P$ changes lie between $-0.0003$ and $+0.0018$ under the median-ratio offset and between $-0.0056$ and $+0.0004$ under smearing; the largest is again a loss, Direct on 2025 sales under smearing. No bootstrap intervals were computed for these two picks. The corrections therefore reduce measured regressivity at essentially
unchanged price-scale accuracy; their direct-exponentiation gains reflect valuation level, and the small
remaining changes have a sign that depends on the level convention.

\begin{table}[!ht]
\centering
\small
\setlength{\tabcolsep}{4.5pt}
\caption{Price-scale $R^2_P$ of the exploratory CV-selected configurations of Table~\ref{tab:cv_selected_comparison} under three level conventions. Each offset is estimated once from the development out-of-fold residuals $u_i=\log P_i-f(x_i)$ of the same configuration's historical development-fold models, with one-sale-one-vote weights, and is applied unchanged to the held-out refit and to the 2025 production refit. It shows how the $R^2_P$ comparison depends on the level convention.}
\label{tab:level_cv_selected}
\begin{tabular}{@{}lrrrrrrr@{}}
\toprule
& & \multicolumn{3}{c}{Held-out} & \multicolumn{3}{c}{2025 forward} \\
\cmidrule(lr){3-5}\cmidrule(l){6-8}
Configuration & $c_{\mathrm{med}}$ & Direct exp. & Median & Smeared & Direct exp. & Median & Smeared \\
\midrule
$\rho=0$ & 0.0374 & 0.8928 & 0.9015 & 0.9041 & 0.9039 & 0.9112 & 0.9131 \\
Direct $\rho=1.677$ & 0.0277 & 0.9006 & 0.9036 & 0.9031 & 0.9104 & 0.9115 & 0.9090 \\
Direct $\rho=2.223$ & 0.0273 & 0.9001 & 0.9019 & 0.9001 & 0.9105 & 0.9110 & 0.9075 \\
Surrogate $\rho=0.409$ & 0.0375 & 0.8948 & 0.9020 & 0.9037 & 0.9073 & 0.9130 & 0.9134 \\
Surrogate $\rho=0.954$ & 0.0351 & 0.8967 & 0.9025 & 0.9029 & 0.9083 & 0.9133 & 0.9129 \\
\bottomrule
\end{tabular}
\vspace{1mm}
\begin{minipage}{\textwidth}
\footnotesize
\emph{Notes.} Direct exp.\ reproduces Table~\ref{tab:cv_selected_comparison}. Median applies $\exp\{f(x_i)+c_{\mathrm{med}}\}$, where $c_{\mathrm{med}}$ is the weighted median of $u_i$, so the development median valuation ratio equals one. Smeared applies this appendix's Duan factor ($s=1.0624$, $1.0587$, $1.0597$, $1.0714$, $1.0765$). Ratio-study, vertical-equity, and residual-structure measures are unchanged by either offset. As elsewhere in this appendix, the 2025 production refit has no out-of-fold residuals of its own, so it receives the offset estimated from the historical development-fold models of the same configuration.
\end{minipage}
\end{table}

\section{Township-Level Changes in Valuation Ratios}
\label{app:townships}

This exploratory appendix describes where the corrections move valuations across Cook County's 38 political townships, a coarse but interpretable geographic unit; every township has at least 30 sales in each later sample. It uses the stored held-out and 2025 predictions of the shared $\rho=0$ fit, Direct $\rho=1.677$, Surrogate $\rho=0.954$, and Surrogate $\rho=0.954$ with post-hoc rescaling ($b=1.027$; Table~\ref{tab:joint_posthoc_comparison}). These illustrative configurations do not represent other penalties. For each configuration, a township's \emph{relative ratio} is its mean (or median) valuation ratio divided by the county mean (or median) ratio of the same configuration, minus one, in percent. It describes relative valuation level, not accuracy against sale price, and is neither PRD nor PRB. The \emph{township gradient} is the ordinary least-squares slope of the relative ratio on township mean log sale price, with townships weighted equally, in percentage points per log-price unit; a negative gradient indicates an overall tendency for lower-price townships to have higher valuation ratios than higher-price townships.

\begin{figure}[!htbp]
\centering
\safeincludegraphics[width=\textwidth]{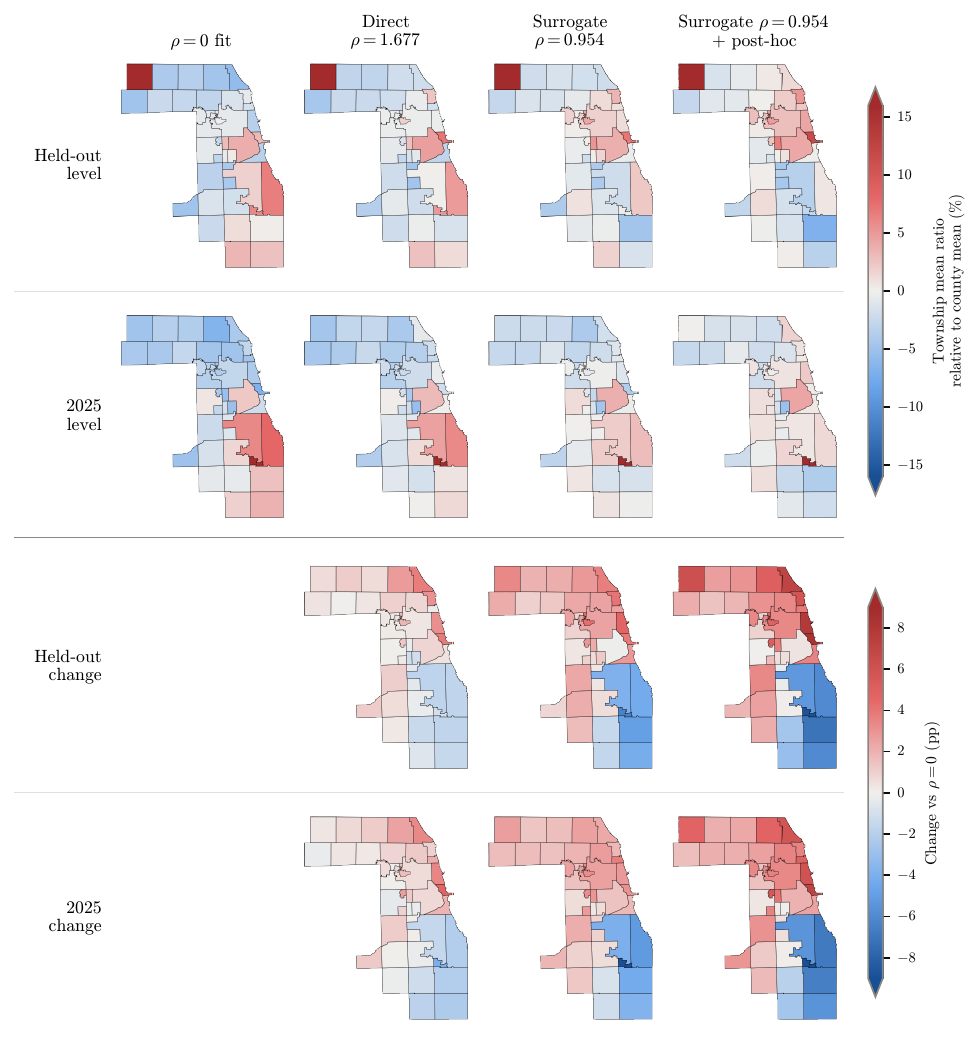}
\caption{Township mean valuation ratios relative to the county mean ratio of the same configuration (top two rows) and their change from the shared $\rho=0$ fit (bottom two rows), on held-out and 2025 sales. County mean ratios for $\rho=0$, Direct, Surrogate, and Surrogate with post-hoc rescaling are 0.986, 0.992, 0.975, and 0.982 (held-out) and 1.009, 1.021, 1.003, and 1.011 (2025). Color scales are shared across samples and clipped at $\pm16\%$ (levels) and $\pm9$ percentage points (changes), so a few townships (e.g., Barrington and Calumet, $+18\%$ to $+31\%$) saturate. Table~\ref{tab:townships} gives the township gradients and median-based summaries.}
\label{fig:township_errors}
\end{figure}

\begin{figure}[!htbp]
\centering
\safeincludegraphics[width=0.55\textwidth]{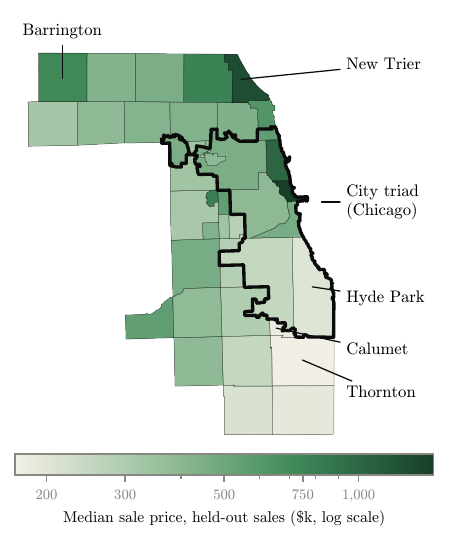}
\caption{Township median sale price on held-out sales (\$k, log color scale), for locating the townships in Figure~\ref{fig:township_errors}; 2025 township medians are about 7\% higher, with nearly identical ranking (Spearman 0.995). The bold outline marks the City triad (the eight Chicago townships). Labels identify townships mentioned in Figure~\ref{fig:township_errors} (Barrington, Calumet) and selected price extremes (Thornton, the cheapest; Hyde Park township, the cheapest in the City triad; New Trier, the most expensive outside it); they are not representative locations.}
\label{fig:township_reference}
\end{figure}

Figure~\ref{fig:township_errors} maps relative mean ratios and their changes, Figure~\ref{fig:township_reference} locates township prices, and Table~\ref{tab:townships} summarizes both means and medians. All three corrections move the negative 2025 mean gradient toward zero (from $-7.5$ to between $-4.6$ and $-0.8$), whereas on held-out sales they turn a smaller negative gradient ($-1.6$) positive ($+0.9$ to $+5.6$, largest for Surrogate with post-hoc). Median-based gradients are positive under every correction in both samples, and the average relative median ratio of the five cheapest townships is below zero under every correction ($-0.1$ to $-8.3\%$). The change maps are similar across the two samples: overall, the corrections raise higher-price townships and lower lower-price ones relative to the county, although not every township follows this pattern. The median across townships of within-township MAPE stays between 15.4\% and 16.6\%. These summaries are descriptive and based on sales rather than all parcels. No intervals are reported for them, a 30-sale minimum does not ensure precise township means, and township averages can hide differences within townships and across property classes.

\begin{table}[!htbp]
\centering
\footnotesize
\setlength{\tabcolsep}{3pt}
\caption{Township summaries of relative valuation ratios on held-out and 2025 sales.}
\label{tab:townships}
\begin{tabular}{@{}l rrrr rrrr r@{}}
\toprule
 & \multicolumn{4}{c}{Mean ratio vs.\ county mean} & \multicolumn{4}{c}{Median ratio vs.\ county median} & \\
\cmidrule(lr){2-5}\cmidrule(lr){6-9}
Configuration & Gradient & Cheapest 5 & Priciest 5 & SD & Gradient & Cheapest 5 & Priciest 5 & SD & MAPE \\
\midrule
\multicolumn{10}{@{}l}{\textit{Panel A: Held-out sales}} \\
$\rho=0$ fit & $-$1.6 & +3.8 & $-$2.7 & 4.7 & +2.2 & $-$2.5 & +0.8 & 2.2 & 16.3 \\
Direct, $\rho=1.677$ & +0.9 & +2.5 & +0.3 & 4.5 & +4.5 & $-$3.9 & +3.4 & 3.1 & 16.1 \\
Surrogate, $\rho=0.954$ & +2.7 & $-$0.5 & +0.9 & 4.7 & +5.7 & $-$6.7 & +3.0 & 4.0 & 16.4 \\
Surrogate, $\rho=0.954$ + post-hoc & +5.6 & $-$2.4 & +4.1 & 5.5 & +8.3 & $-$8.3 & +5.8 & 5.1 & 16.6 \\
\addlinespace[3pt]
\multicolumn{10}{@{}l}{\textit{Panel B: 2025 sales}} \\
$\rho=0$ fit & $-$7.5 & +9.5 & $-$5.1 & 6.4 & $-$1.6 & +1.7 & $-$1.4 & 2.7 & 15.9 \\
Direct, $\rho=1.677$ & $-$4.6 & +7.0 & $-$1.8 & 5.3 & +1.1 & $-$0.1 & +2.0 & 2.6 & 15.4 \\
Surrogate, $\rho=0.954$ & $-$3.5 & +4.6 & $-$2.3 & 4.0 & +1.3 & $-$2.5 & +0.0 & 2.3 & 15.8 \\
Surrogate, $\rho=0.954$ + post-hoc & $-$0.8 & +2.6 & +0.7 & 3.6 & +3.7 & $-$4.3 & +2.7 & 3.0 & 15.7 \\
\bottomrule
\end{tabular}

\vspace{1mm}
\begin{minipage}{\textwidth}
\scriptsize
\emph{Notes.} Relative ratios are township mean (median) valuation ratios divided by the county mean (median) ratio of the same configuration, minus one, in percent. Gradient: ordinary least-squares slope of the relative ratio on township mean log sale price across the 38 townships, weighted equally, in percentage points per log-price unit; negative values indicate the regressive direction. Cheapest 5 and Priciest 5: average relative ratio of the five townships with the lowest and highest mean log sale price. SD: standard deviation of relative ratios across townships, in percentage points. MAPE: median across townships of within-township MAPE (\%). Descriptive sale-based summaries without intervals.
\end{minipage}
\end{table}

\FloatBarrier

\section{External-Jurisdiction Benchmark}
\label{app:external_benchmark}

\paragraph{Design and protocol.} This is a separate benchmark of the same objective families, not a rerun of
the Cook County Assessor's Office's feature or sales-verification pipeline.
The primary counties are Wayne (MI), Philadelphia (PA), St.~Louis County
(MO), Allegheny (PA), Maricopa (AZ), King (WA), Miami--Dade (FL), and
Middlesex (MA). Wayne County is not Detroit, and St.~Louis County is not
St.~Louis City. Cook is a separate ATTOM reference; its sales and features
are not the restricted CCAO extract in the main experiment.

The benchmark's residential cohorts link ATTOM Assessor History features with
qualifying ATTOM Recorder transfers, accessed through Dewey Data under MIT
Libraries' institutional subscription
\cite{ATTOMAssessorHistory2025,ATTOMRecorder2025}; the primary pre-sale feature set includes
prior market history, but excludes assessment values, Tax Assessor and ACS
inputs, and future transaction fields. Seven expanding calendar-year folds
validate on 2018--2024 after training on earlier sales from 2016 onward;
models are refit on eligible 2016--2024 sales for evaluation on 2025 sales.
The county-specific LightGBM settings were selected on CV from a common
candidate grid before the penalty sweeps. This differs from the CCAO folds,
feature set, and fixed 994-tree configuration.
Both external families use $\widetilde\rho=\rho\widehat{\Var}_T(\log P)$,
where the variance is recomputed on each training block. This coordinate
does not make Direct and Surrogate strengths equivalent, and the main CCAO
results use unnormalized $\rho$.
For each unit and family, the 25\% anchor is the normalized penalty at which
the fold-mean $|\beta_{\log}|$ over the seven development folds falls 25\%
below its value at the family's reference point, interpolated linearly between
adjacent points of the stored CV path. The reference point is $\rho=0$ for
Direct and, because the Surrogate's CV path has no $\rho=0$ member, its smallest
grid point ($\widetilde\rho=0.001$, a custom-objective fit) for the Surrogate.
Each anchor was refitted on the eligible 2016--2024 sales and evaluated on 2025
sales. For Maricopa, the stored CV paths record a few penalty values with
last-digit differences across folds, so the original calculation averaged those
path points over only some folds. After merging them, the Direct anchor moves
from $\widetilde\rho=1.347$ to $1.576$ and was refitted, and the Maricopa
change-point regions become $[0.267,0.816]$ (Direct) and $[0.0873,0.562]$
(Surrogate); the overlaps reported below are unchanged.

The 2025-forward sample is distinct from CV; the CV fold means are not
independent pre-2025 test samples.

\paragraph{Record filter.} In several counties, a few
recorded transfers are inconsistent with a single-family sale; for example, a
\$118.5M Middlesex transfer of a 220,800 sqft building with no bedrooms is coded
as single-family. In all external counties, we exclude from training and
evaluation the sales with building area above 15,000 sqft, price per building
sqft outside the county's 2016--2024 0.1--99.9\% range, or a price more than ten
times above or below the as-of median of the ten nearest location-cell
comparables (sales before 1 January of the sale year). The filter removes 0.16--0.33\% of
sales: 333 of 182,042 in Wayne, 148 of 66,072 in Philadelphia, 337 of 145,040 in
St.~Louis County, 194 of 122,794 in Allegheny, 1,434 of 600,644 in Maricopa, 450 of 195,519 in King,
505 of 152,762 in Miami--Dade, and 210 of 83,343 in Middlesex, including 15, 4, 35,
41, 253, 51, 69, and 37 of their 2025 sales, respectively.
The filter was added after the initial 2025 evaluation exposed such records. Its
size and ratio constants were set then; its price-per-sqft ranges use only
2016--2024 sales, its comparable medians use only sales before 1 January of each
sale's year, and it applies equally to training and evaluation sales. The
filtered 2025 sample is therefore not an evaluation of a protocol fixed entirely
in advance.
For these counties, the development paths, change-point
regions, anchors, and 2025 evaluations were rerun unchanged on the
filtered cohorts; the Cook reference sample is unchanged. The 2025 baseline $R^2_P$ rises
from 0.833 to 0.838 (Wayne), 0.749 to 0.804 (Philadelphia), 0.812 to 0.858
(St.~Louis County), 0.717 to 0.844 (Allegheny), 0.297 to 0.858 (Maricopa), 0.750 to
0.783 (King), 0.725 to 0.842 (Miami--Dade), and 0.253 to 0.822 (Middlesex).

\paragraph{Reference fit.} All external changes are measured from the
custom-objective $\rho=0$ fit in the same unit. This fit is the Direct baseline
and shares its execution path with the Surrogate's penalized fits. A native
LightGBM $\rho=0$ fit starts from the same training mean but follows a
different execution path, and the two are not numerically identical: they
differ by a mean of 0.011--0.031 in log prediction per county, the same order
as refitting the custom objective at $\widetilde\rho=0.001$ (0.008--0.020).
Measuring the Surrogate's changes from the native fit instead leaves every
reported sign unchanged. The CCAO paths
have their own measured native/custom difference
(Appendix~\ref{subsec:zero_control_results}), and absolute prediction and
equity levels should not be compared directly with those in the CCAO experiment.

\paragraph{Baseline levels.} Table~\ref{tab:external_baselines} gives the
reference levels. All nine baselines have negative PRB and $\beta_{\log}$ on
their 2025 sales. The 2025 levels can shift substantially from the CV fold
means, and different cohort retention and available features limit direct
comparison between counties.

\begin{table}[!htbp]
\centering
\scriptsize
\setlength{\tabcolsep}{4pt}
\caption{Unpenalized custom-objective fits ($\rho=0$) on 2025 sales: the reference for every external change.}
\label{tab:external_baselines}
\begin{tabular}{@{}l rrrrrr@{}}
\toprule
Jurisdiction & 2025 sales & PRB & $\beta_{\log}$ & $R^2_P$ & $\operatorname{RMSE}_{\log P}$ & VEI (\%) \\
\midrule
Wayne, MI & 17,332 & $-0.112$ & $-0.206$ & 0.838 & 0.294 & $-35.2$ \\
Philadelphia, PA & 2,917 & $-0.203$ & $-0.308$ & 0.804 & 0.312 & $-42.6$ \\
St.~Louis County, MO & 13,684 & $-0.055$ & $-0.125$ & 0.858 & 0.300 & $-27.6$ \\
Allegheny, PA & 12,072 & $-0.118$ & $-0.217$ & 0.844 & 0.323 & $-40.0$ \\
Maricopa, AZ & 47,259 & $-0.058$ & $-0.140$ & 0.858 & 0.198 & $-13.5$ \\
King, WA & 17,761 & $-0.065$ & $-0.157$ & 0.783 & 0.219 & $-8.3$ \\
Miami--Dade, FL & 11,840 & $-0.048$ & $-0.147$ & 0.842 & 0.260 & $-13.2$ \\
Middlesex, MA & 9,185 & $-0.082$ & $-0.172$ & 0.822 & 0.237 & $-5.3$ \\
\midrule
Cook, IL (ATTOM reference) & 24,993 & $-0.109$ & $-0.184$ & 0.830 & 0.283 & $-27.8$ \\
\bottomrule
\end{tabular}
\vspace{1mm}
\begin{minipage}{\textwidth}
\scriptsize
\emph{Notes.} All external counties use the filtered cohort (\emph{Record filter} paragraph). Cook is an ATTOM reference sample, not an external county or the main restricted CCAO sample; its levels are not those of the CCAO experiment.
\end{minipage}
\end{table}

\paragraph{Results at the anchors.} Table~\ref{tab:external_full} gives the full
2025 results. On the stored development-only refits, the anchors reduce the
fold-mean slope magnitude by 18.6--28.0\% (Dev.), close to the 25\% target. On 2025
sales both families
reduce $|\beta_{\log}|$ and $|\mathrm{PRB}|$ in all eight external counties, but
the slope reduction ranges from 19.3\% to 36.1\% for Direct and from
18.9\% to 51.8\% for the Surrogate. PRD and MKI also move toward one in all nine units
for all three methods. Where the unpenalized VEI is smallest ($-5.3\%$ in Middlesex
and $-8.3\%$ in King), the corrections move VEI to near or above zero (from $-0.2\%$
to $+10.2\%$); only the Middlesex Surrogate ($+10.2\%$) leaves the proposed $\pm10\%$ band. The Middlesex
Surrogate anchor also moves PRB just past zero ($-0.082$ to $+0.007$).

\begin{table}[!htbp]
\centering
\scriptsize
\setlength{\tabcolsep}{3.2pt}
\caption{Full 2025 results at the 25\% development anchors and for the post-hoc rescaling, in the eight external counties and the Cook ATTOM reference sample. Every change is relative to the same unit's custom-objective $\rho=0$ fit (Table~\ref{tab:external_baselines}).}
\label{tab:external_full}
\begin{tabular}{@{}lllr rrrrrr@{}}
\toprule
Jurisdiction & Method & Setting & \makecell[r]{Dev.\\(\%)} & \makecell[r]{Slope\\red.\ (\%)} & \makecell[r]{$|\mathrm{PRB}|$\\red.\ (\%)} & $\Delta R^2_P$ & \makecell[r]{$\operatorname{RMSE}_{\log P}$\\change (\%)} & $\Delta\mathrm{VEI}$ & $\Delta(\Delta_{\mathrm{NL}})$ \\
\midrule
Wayne, MI & Direct & $\widetilde\rho=2.444$ & 26.6 & 32.7 & 59.1 & $+0.004$ & $+1.0$ & $+19.0$ & $+0.018$ \\
 & Surrogate & $\widetilde\rho=1.476$ & 24.7 & 39.5 & 72.5 & $+0.004$ & $+5.6$ & $+14.8$ & $+0.003$ \\
 & Post-hoc & $b=1.070$ & 25.0 & 27.0 & 48.5 & $+0.009$ & $+0.2$ & $+15.4$ & $+0.015$ \\
\addlinespace[2pt]
Philadelphia, PA & Direct & $\widetilde\rho=1.765$ & 26.2 & 26.5 & 49.0 & $-0.007$ & $+0.9$ & $+22.2$ & $+0.013$ \\
 & Surrogate & $\widetilde\rho=0.796$ & 24.1 & 26.1 & 49.6 & $+0.015$ & $+1.9$ & $+23.5$ & $-0.025$ \\
 & Post-hoc & $b=1.112$ & 25.0 & 25.1 & 46.1 & $-0.005$ & $+0.3$ & $+21.8$ & $+0.016$ \\
\addlinespace[2pt]
St.~Louis County, MO & Direct & $\widetilde\rho=2.012$ & 24.8 & 35.4 & 70.5 & $+0.022$ & $+0.2$ & $+14.8$ & $+0.008$ \\
 & Surrogate & $\widetilde\rho=0.631$ & 24.0 & 34.9 & 71.7 & $+0.008$ & $+3.2$ & $+10.8$ & $+0.003$ \\
 & Post-hoc & $b=1.040$ & 25.0 & 28.4 & 56.9 & $+0.019$ & $+0.3$ & $+11.5$ & $+0.007$ \\
\addlinespace[2pt]
Allegheny, PA & Direct & $\widetilde\rho=1.452$ & 24.9 & 29.2 & 55.3 & $-0.007$ & $+1.3$ & $+19.2$ & $+0.018$ \\
 & Surrogate & $\widetilde\rho=0.686$ & 24.2 & 34.1 & 66.4 & $-0.004$ & $+4.6$ & $+17.9$ & $+0.002$ \\
 & Post-hoc & $b=1.073$ & 25.0 & 26.1 & 49.5 & $-0.003$ & $+1.0$ & $+17.4$ & $+0.017$ \\
\addlinespace[2pt]
Maricopa, AZ & Direct & $\widetilde\rho=1.576$ & 25.1 & 19.3 & 41.2 & $-0.005$ & $+1.1$ & $+5.6$ & $+0.006$ \\
 & Surrogate & $\widetilde\rho=0.520$ & 24.4 & 23.6 & 48.5 & $+0.008$ & $+1.3$ & $+7.4$ & $-0.011$ \\
 & Post-hoc & $b=1.030$ & 25.0 & 18.4 & 38.5 & $-0.001$ & $+0.7$ & $+5.4$ & $+0.007$ \\
\addlinespace[2pt]
King, WA & Direct & $\widetilde\rho=1.894$ & 24.7 & 31.2 & 66.5 & $-0.005$ & $+1.8$ & $+10.2$ & $+0.008$ \\
 & Surrogate & $\widetilde\rho=0.514$ & 23.5 & 27.8 & 60.6 & $+0.017$ & $+2.6$ & $+8.0$ & $-0.003$ \\
 & Post-hoc & $b=1.050$ & 25.0 & 27.0 & 56.6 & $-0.002$ & $+0.9$ & $+9.1$ & $+0.008$ \\
\addlinespace[2pt]
Miami--Dade, FL & Direct & $\widetilde\rho=1.133$ & 26.1 & 22.1 & 65.1 & $-0.018$ & $+1.3$ & $+7.5$ & $+0.011$ \\
 & Surrogate & $\widetilde\rho=0.219$ & 26.1 & 18.9 & 56.0 & $+0.002$ & $+1.0$ & $+7.4$ & $-0.002$ \\
 & Post-hoc & $b=1.038$ & 25.0 & 22.1 & 63.8 & $-0.015$ & $+0.9$ & $+8.6$ & $+0.011$ \\
\addlinespace[2pt]
Middlesex, MA & Direct & $\widetilde\rho=2.354$ & 18.6 & 36.1 & 72.5 & $-0.004$ & $+1.7$ & $+14.0$ & $+0.021$ \\
 & Surrogate & $\widetilde\rho=1.201$ & 28.0 & 51.8 & 91.7 & $+0.002$ & $+11.1$ & $+15.4$ & $-0.007$ \\
 & Post-hoc & $b=1.078$ & 25.0 & 37.8 & 73.7 & $-0.002$ & $+1.1$ & $+14.3$ & $+0.021$ \\
\midrule
Cook, IL (ATTOM reference) & Direct & $\widetilde\rho=0.807$ & 24.6 & 19.9 & 33.2 & $+0.009$ & $-0.1$ & $+10.1$ & $+0.012$ \\
 & Surrogate & $\widetilde\rho=0.215$ & 24.9 & 22.5 & 38.6 & $+0.010$ & $+1.2$ & $+10.6$ & $-0.010$ \\
 & Post-hoc & $b=1.047$ & 25.0 & 20.8 & 34.3 & $+0.009$ & $-0.3$ & $+11.7$ & $+0.012$ \\
\bottomrule
\end{tabular}
\vspace{1mm}
\begin{minipage}{\textwidth}
\scriptsize
\emph{Notes.} Setting is the normalized penalty $\widetilde\rho$ or the post-hoc scale $b$. Dev.\ is the anchor-selection calculation: the achieved reduction in fold-mean $|\beta_{\log}|$ on the stored development-only refits, measured against each anchor's reference (for the Surrogate, its smallest grid point, a custom-objective fit); $b$ gives 25\% by construction. Slope and $|\mathrm{PRB}|$ reductions are percentage reductions in $|\beta_{\log}|$ and $|\mathrm{PRB}|$ on 2025 sales, and the $\operatorname{RMSE}_{\log P}$ change is a percentage change; VEI changes are in percentage points; a positive $\Delta(\Delta_{\mathrm{NL}})$ is a larger nonlinear gap. The post-hoc rescaling maps each unit's unpenalized log prediction $\hat f_0$ to $\bar y_T+b(\hat f_0-\bar y_T)$, where $\bar y_T$ is the training-block mean log price; $b$ was solved in closed form on the seven development folds and is reported as a reference. 
\end{minipage}
\end{table}

\paragraph{Post-hoc reference.} Table~\ref{tab:external_full} also reports a
reference rescaling of each unit's unpenalized fit as
$\bar y_T+b(\hat f_0-\bar y_T)$, with $b$ fixed in closed form on the
development folds to give exactly the 25\% fold-mean slope reduction that
defines the anchors. Matching $b$ instead to the anchors' achieved development
reductions, or to the original CV-path slopes, changes it by at most 0.021.
On 2025 sales the rescaling reduces $|\beta_{\log}|$ and $|\mathrm{PRB}|$ in all
nine units, but its $|\mathrm{PRB}|$ reduction is smaller than Direct's in seven
of the eight external counties. Its nonlinear-gap change stays within 0.005 of
Direct's, and its $\operatorname{RMSE}_{\log P}$ change is smaller than either
anchor's in every unit except St.~Louis County, where Direct's is smaller
($+0.2\%$ versus $+0.3\%$).

\paragraph{Inference.} The paired percentile intervals at the 25\% anchors
resample whole calendar months of 2025 sales (200 draws) and measure changes
from the custom-objective $\rho=0$ fit; the post-hoc intervals use the same
draws. Measuring the Surrogate from the native fit gives identical exclusions:
for each family, the
intervals for the PRB and slope changes exclude zero in all eight external
counties, and so do those for the change in log-price error, except Direct in
Philadelphia and St.~Louis County. For the post-hoc
rescaling, the PRB and slope intervals exclude zero in all nine units and the
$\operatorname{RMSE}_{\log P}$ interval in eight (all but Philadelphia). On the
same draws, the intervals for the $|\mathrm{PRB}|$ reduction and the
$\operatorname{RMSE}_{\log P}$ change exclude zero exactly when those for the
PRB and log-price-error changes do.
This is descriptive uncertainty for sales in one forward year, not evidence
about unsold parcels, official assessments, or final tax burdens. The activity anchor is the unit's activity onset in the
change-point screen below: the first tested normalized penalty on its CV
path by which at least three of the five neutral-distance measures (PRD, PRB,
MKI, VEI, and $\beta_{\log}$) have entered a supported improving regime. It is a
smaller penalty than the 25\% anchor in every unit. At this
anchor, both families again reduce $|\beta_{\log}|$ and
$|\mathrm{PRB}|$ in all eight external counties, by less than at the 25\% anchor,
and the costs remain: Direct's nonlinear gap rises in all eight, and the
Surrogate's $\operatorname{RMSE}_{\log P}$ in six (all but Philadelphia and St.~Louis County).

\paragraph{Tolerance bands and change-point screen.} The Cook tolerance band
(Section~\ref{subsec:candidate_region_results}), applied unchanged to each
county's normalized CV path, exists for both families in all eight external
counties (Table~\ref{tab:external_selected_full}). Its upper end is set by
$\Delta_{\mathrm{NL}}$ or COD for Direct (with MAPE in Middlesex) and by
log-price RMSE, COD, or MAPE for the Surrogate (jointly with the median-ratio limit in Wayne and with
$\Delta_{\mathrm{NL}}$ in St.~Louis County). The descriptive change-point
screen, by contrast, sometimes abstains: Allegheny has no stable Direct
region, and Middlesex has a Surrogate guardrail before its activity onset.
Eight of the nine study units have a protocol-valid Direct region (Allegheny
does not), and eight have a protocol-valid Surrogate region (Middlesex does
not). The normalized Direct regions overlap on $[0.387,0.562]$ among the eight
protocol-valid units, seven external counties plus the Cook ATTOM reference sample, so
this intersection is not a result for all eight external counties; the
Surrogate regions overlap only on $[0.127,0.184]$ across its eight protocol-valid
units. These overlaps are descriptions of the CV screen, not safe settings or
transferable penalty recommendations.

\begin{figure}[!htbp]
\centering
\safeincludegraphics[width=0.92\textwidth]{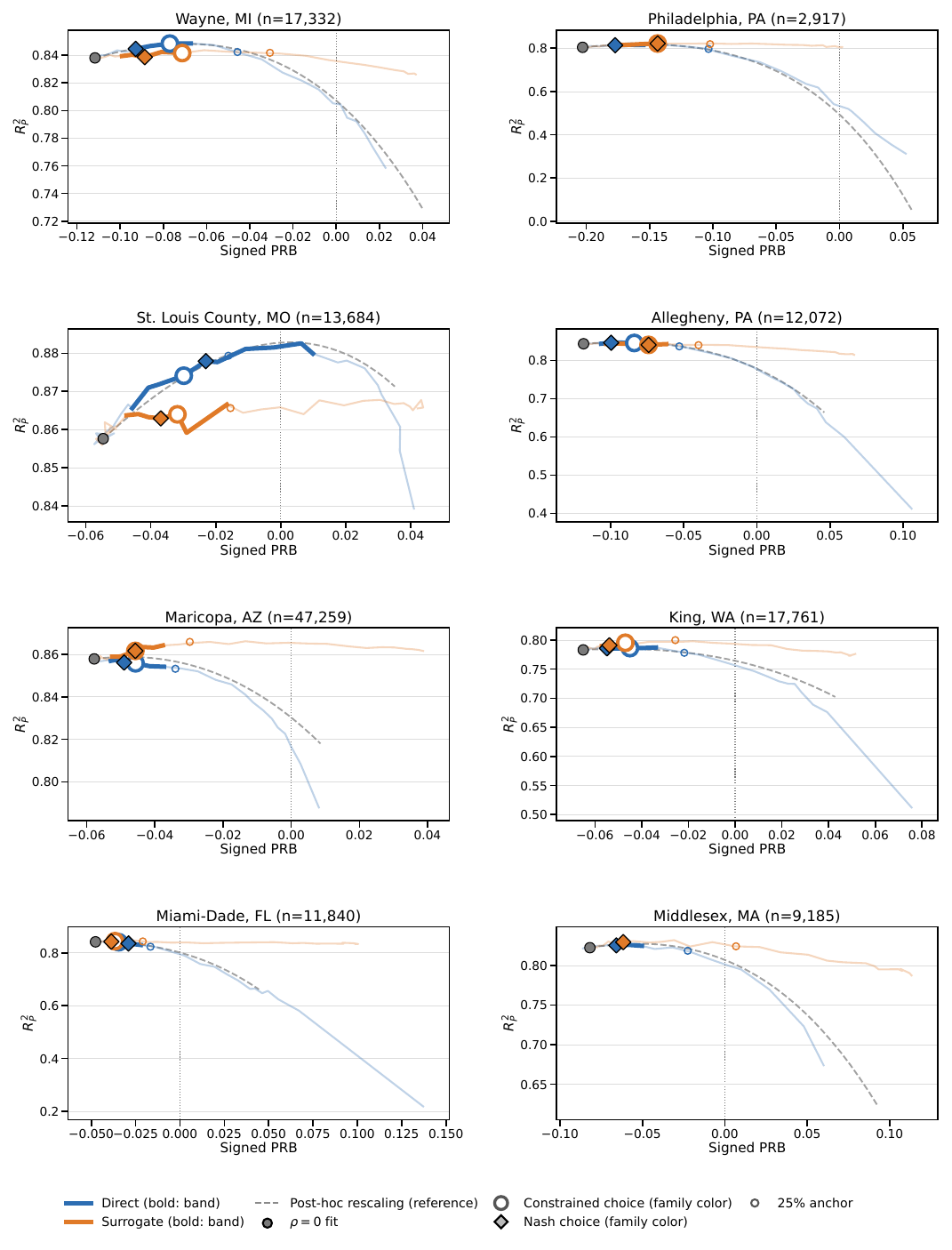}
\caption{2025 price-scale accuracy--regressivity paths in the eight external counties: signed PRB (horizontal) and $R^2_P$ (vertical) over every tested penalty. Bold segments mark each family's development-CV tolerance band; open circles and diamonds mark the constrained and Nash choices of Table~\ref{tab:external_selected_full}; small hollow circles mark the 25\% anchors of Table~\ref{tab:external_full}; gray circles mark the custom-objective $\rho=0$ fit. The dashed gray curve is the post-hoc rescaling of that fit, shown for reference.}
\label{fig:external_tradeoff_full}
\end{figure}

\paragraph{CV selections and full paths.} Table~\ref{tab:external_selected_full}
applies the Cook constrained and Nash rules inside each county's tolerance band.
The constrained rule abstains for Direct in Philadelphia and Middlesex and for the Surrogate in Middlesex, where no band point reduces CV $|\mathrm{PRB}|$ by 25\%. On 2025 sales, all 29 choices reduce $|\mathrm{PRB}|$ (by 12.6--57.8\%), with $R^2_P$ changes between $-0.006$ and $+0.020$. Direct raises $\Delta_{\mathrm{NL}}$ at all 14 of its choices and log-price RMSE at 9 (by at most 0.7\%); the Surrogate raises log-price RMSE at 13 of 15 (by at most 1.8\%) and $\Delta_{\mathrm{NL}}$ at 9. Figure~\ref{fig:external_tradeoff_full} shows the full 2025
paths with the bands and choices. Post-hoc rescaling appears only as a reference
path; selecting its scale jointly with the penalty is discussed in
Section~\ref{subsec:posthoc_comparator}.

\scriptsize
\begingroup\setlength{\tabcolsep}{3pt}
\begin{longtable}{lllllrrrrrr}
\caption{Development-CV selections in the eight external counties under the tolerance-band rules of Appendix~\ref{subsec:screening_interpretation}, and their 2025 outcomes. Changes are from the county's custom-objective $\widetilde\rho=0$ fit.}\label{tab:external_selected_full}\\
\toprule County & Method & Band ($\widetilde\rho$) & Rule & $\widetilde\rho$ & $R^2_P$ & PRB & $\Delta R^2_P$ & \makecell[r]{$|\mathrm{PRB}|$\\red.\ (\%)} & \makecell[r]{log-price\\RMSE (\%)} & $\Delta(\Delta_{\mathrm{NL}})$ \\ \midrule
\endfirsthead
\toprule County & Method & Band ($\widetilde\rho$) & Rule & $\widetilde\rho$ & $R^2_P$ & PRB & $\Delta R^2_P$ & \makecell[r]{$|\mathrm{PRB}|$\\red.\ (\%)} & \makecell[r]{log-price\\RMSE (\%)} & $\Delta(\Delta_{\mathrm{NL}})$ \\ \midrule
\endhead
Wayne, MI & Direct & $[0.387,1.18]$ & Constrained & 0.816 & 0.848 & $-0.077$ & $+0.010$ & 31.0 & $0.0$ & $+0.011$ \\
 &  &  & Nash & 0.387 & 0.845 & $-0.093$ & $+0.007$ & 16.9 & $-0.1$ & $+0.007$ \\
 & Surrogate & $[0.0873,0.387]$ & Constrained & 0.387 & 0.842 & $-0.071$ & $+0.003$ & 36.1 & $+1.7$ & $+0.002$ \\
 &  &  & Nash & 0.184 & 0.839 & $-0.089$ & $+0.001$ & 20.6 & $+1.0$ & $+0.001$ \\
\addlinespace
Philadelphia, PA & Direct & $[0.267,0.562]$ & Constrained & abstain & \multicolumn{6}{c}{no point reduces CV $|\mathrm{PRB}|$ by 25\% inside the band} \\
 &  &  & Nash & 0.267 & 0.814 & $-0.177$ & $+0.010$ & 12.6 & $0.0$ & $+0.004$ \\
 & Surrogate & $[0.0873,0.267]$ & Constrained & 0.267 & 0.822 & $-0.144$ & $+0.018$ & 29.2 & $-0.2$ & $-0.008$ \\
 &  &  & Nash & 0.267 & 0.822 & $-0.144$ & $+0.018$ & 29.2 & $-0.2$ & $-0.008$ \\
\addlinespace
St.~Louis County, MO & Direct & $[0.267,7.62]$ & Constrained & 0.816 & 0.874 & $-0.030$ & $+0.017$ & 45.4 & $-0.3$ & $+0.005$ \\
 &  &  & Nash & 1.18 & 0.878 & $-0.023$ & $+0.020$ & 57.8 & $-0.2$ & $+0.007$ \\
 & Surrogate & $[0.0602,0.562]$ & Constrained & 0.267 & 0.864 & $-0.032$ & $+0.006$ & 41.8 & $+1.0$ & $+0.001$ \\
 &  &  & Nash & 0.184 & 0.863 & $-0.037$ & $+0.005$ & 32.4 & $+0.8$ & $0.000$ \\
\addlinespace
Allegheny, PA & Direct & $[0.184,0.562]$ & Constrained & 0.562 & 0.846 & $-0.084$ & $+0.002$ & 29.3 & $+0.5$ & $+0.009$ \\
 &  &  & Nash & 0.267 & 0.846 & $-0.100$ & $+0.003$ & 15.9 & $-0.1$ & $+0.004$ \\
 & Surrogate & $[0.0602,0.387]$ & Constrained & 0.267 & 0.841 & $-0.074$ & $-0.003$ & 37.5 & $+1.8$ & $+0.001$ \\
 &  &  & Nash & 0.267 & 0.841 & $-0.074$ & $-0.003$ & 37.5 & $+1.8$ & $+0.001$ \\
\addlinespace
Maricopa, AZ & Direct & $[0.184,1.18]$ & Constrained & 0.562 & 0.856 & $-0.046$ & $-0.002$ & 21.0 & $+0.5$ & $+0.002$ \\
 &  &  & Nash & 0.387 & 0.856 & $-0.049$ & $-0.002$ & 15.2 & $+0.3$ & $+0.001$ \\
 & Surrogate & $[0.0286,0.267]$ & Constrained & 0.0873 & 0.862 & $-0.046$ & $+0.004$ & 20.9 & $+0.5$ & $-0.002$ \\
 &  &  & Nash & 0.0873 & 0.862 & $-0.046$ & $+0.004$ & 20.9 & $+0.5$ & $-0.002$ \\
\addlinespace
King, WA & Direct & $[0.267,1.18]$ & Constrained & 0.562 & 0.786 & $-0.045$ & $+0.003$ & 30.7 & $+0.4$ & $+0.004$ \\
 &  &  & Nash & 0.267 & 0.786 & $-0.055$ & $+0.003$ & 15.6 & $0.0$ & $+0.002$ \\
 & Surrogate & $[0.0415,0.127]$ & Constrained & 0.127 & 0.795 & $-0.047$ & $+0.012$ & 27.7 & $+0.4$ & $0.000$ \\
 &  &  & Nash & 0.0602 & 0.791 & $-0.054$ & $+0.008$ & 17.2 & $+0.1$ & $-0.002$ \\
\addlinespace
Miami--Dade, FL & Direct & $[0.184,0.816]$ & Constrained & 0.387 & 0.840 & $-0.034$ & $-0.001$ & 27.5 & $+0.3$ & $+0.004$ \\
 &  &  & Nash & 0.562 & 0.835 & $-0.029$ & $-0.006$ & 39.0 & $+0.7$ & $+0.006$ \\
 & Surrogate & $[0.0286,0.0873]$ & Constrained & 0.0602 & 0.844 & $-0.037$ & $+0.003$ & 23.0 & $+0.3$ & $+0.002$ \\
 &  &  & Nash & 0.0286 & 0.842 & $-0.039$ & $+0.001$ & 18.8 & $+0.3$ & $+0.004$ \\
\addlinespace
Middlesex, MA & Direct & $[0.387,0.816]$ & Constrained & abstain & \multicolumn{6}{c}{no point reduces CV $|\mathrm{PRB}|$ by 25\% inside the band} \\
 &  &  & Nash & 0.387 & 0.825 & $-0.066$ & $+0.003$ & 19.6 & $+0.1$ & $+0.008$ \\
 & Surrogate & $[0.0602,0.0873]$ & Constrained & abstain & \multicolumn{6}{c}{no point reduces CV $|\mathrm{PRB}|$ by 25\% inside the band} \\
 &  &  & Nash & 0.0873 & 0.829 & $-0.062$ & $+0.007$ & 24.7 & $+0.8$ & $+0.003$ \\
\bottomrule
\multicolumn{11}{p{0.97\textwidth}}{\scriptsize\emph{Notes.} Bands and rules are those of Cook County ($\tau=2\%$; Appendix~\ref{subsec:screening_interpretation}), applied to each county's normalized CV path; $\beta_{\log}$ is not a selection criterion. Constrained: smallest band point with CV $|\mathrm{PRB}|$ reduced by at least 25\%, otherwise abstain. Nash: tolerance-disagreement Nash product inside the band. Penalty values are not comparable across families. All counties use the filtered cohorts.}\\
\end{longtable}
\endgroup

\paragraph{Nonlinear and ratio-profile diagnostics.} Direct's nonlinear gap
rises in all eight external counties, and the Surrogate's falls in five
(Table~\ref{tab:external_full}). Under 20 alternative fold partitions, every external
$\Delta(\Delta_{\mathrm{NL}})$ sign and count reported here is unchanged.

\begin{figure}[!htbp]
\centering
\safeincludegraphics[width=\textwidth]{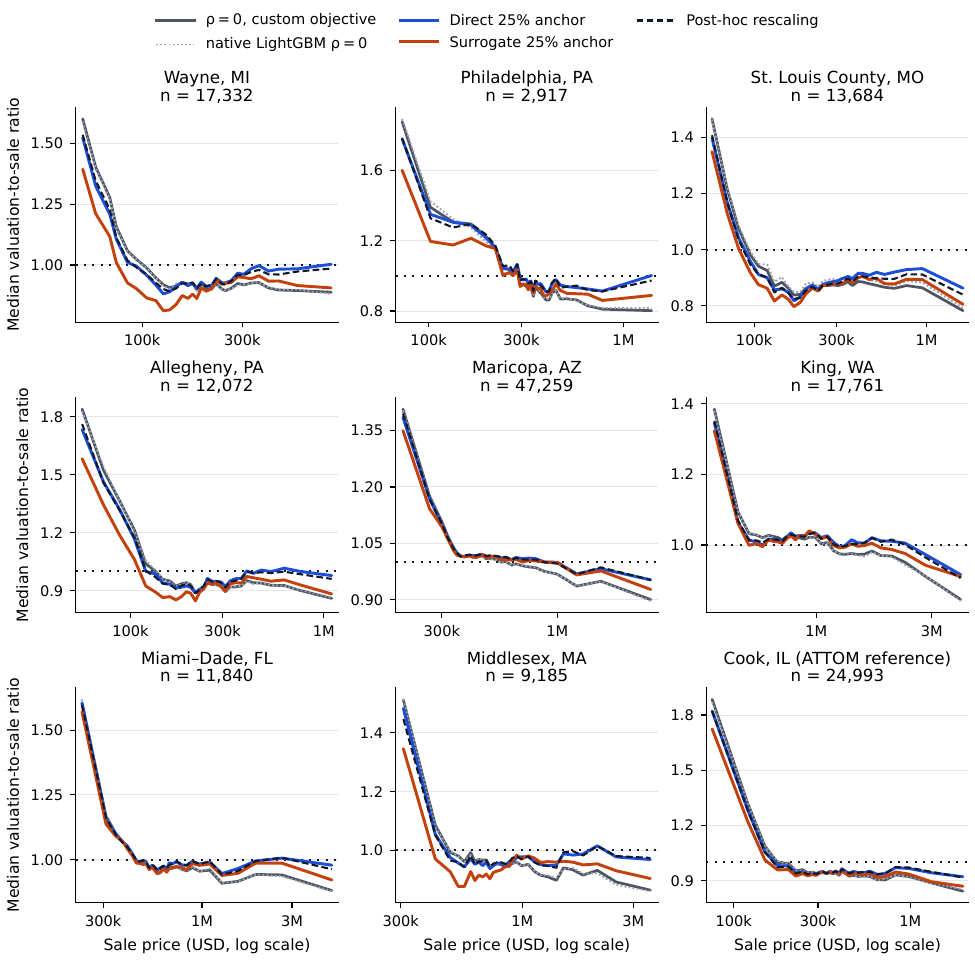}
\caption{Exploratory 2025 ratio profiles in the eight external counties and the Cook ATTOM reference sample: median valuation-to-sale ratios across 30 equal-count sale-price bins for the custom-objective $\rho=0$ fit (solid gray), a native LightGBM $\rho=0$ fit (dotted gray), the Direct and Surrogate 25\% development anchors, and the post-hoc rescaling (dashed). Within each unit, every curve uses the same bins and sales, plotted at the bins' median sale prices; vertical scales differ by panel. The dotted black line at 1 denotes equal valuation and sale price.}
\label{fig:external_ratio_profiles}
\end{figure}

Figure~\ref{fig:external_ratio_profiles} applies the construction of
Figure~\ref{fig:ratio_shape_anchor_profiles} to each unit's 2025 sales. The
Direct upper-price lift recurs: in all nine units Direct raises the median
ratios of the highest-priced third of bins relative to $\rho=0$ (by 0.02--0.08 on
average) and lowers those of the lowest-priced third, and the post-hoc rescaling
traces nearly the same profile.
We measure local depressions by the depth of
Appendix~\ref{subsec:screening_interpretation} and call a depression a
low-to-middle-price trough when its bottom lies in the lower two thirds of the
bins and it is at least 0.03 deeper than at $\rho=0$; the 0.03 threshold is
taken from the CCAO depths. Both
anchors have such a trough in four units (Wayne, St.~Louis County, Allegheny and Middlesex), and the
Surrogate also in King, but they deepen it differently: the Surrogate mainly lowers the
low-to-middle ratios, while much of Direct's added depth comes from raising the
upper-price ratios that bound the depression (more than half of it in all four
units). The Surrogate's trough is the deeper of the two in
Wayne and Middlesex. Because the bins are
formed from commercially recorded sale prices, discrepancies of the kind
documented for ATTOM in Cook County \cite{SmithHarveyGoldinHo2026Records} would
move sales between bins, and the lowest- and highest-priced bins are the most
exposed.

\FloatBarrier

\section*{Data, Code, and Institutional Statement}

The Cook County analysis uses a sales extract from the Cook County Assessor's Office (CCAO).
The external benchmark uses ATTOM data licensed through Dewey Data
\cite{ATTOMAssessorHistory2025,ATTOMRecorder2025}. We do not redistribute either dataset or the
record-level predictions. The analysis code and derived results are archived by the authors. All
results come from a research implementation, not from official CCAO assessments. The Direct
objective is available as an option in the CCAO's public residential modeling code
\cite{CCAOWagnerResCovPR2026}. The views expressed are those of the authors and do not
necessarily reflect those of the Cook County Assessor's Office.

\end{document}